\documentclass{article}

\usepackage[preprint]{neurips_2025}

\usepackage[utf8]{inputenc} % allow utf-8 input
\usepackage[T1]{fontenc}    % use 8-bit T1 fonts
\usepackage{hyperref}       % hyperlinks
\usepackage{url}            % simple URL typesetting
\usepackage{booktabs}       % professional-quality tables
\usepackage{amsfonts}       % blackboard math symbols
\usepackage{nicefrac}       % compact symbols for 1/2, etc.
\usepackage{microtype}      % microtypography
\usepackage{xcolor}         % colors

\usepackage{wrapfig}
\usepackage{hyperref}
 \usepackage{graphicx}
 \usepackage{subfig} 
\usepackage{url}
\usepackage{amsmath}
\usepackage{amsthm}
\usepackage{subcaption}
\usepackage{algorithmic}
\usepackage{algorithm}
\usepackage{bm}
\usepackage{array}
\usepackage{comment}
\usepackage{multirow}
\usepackage{amssymb} % For math symbols
\usepackage{amsfonts} % For math fonts
\usepackage{svg}

\newtheorem{theorem}{Theorem}[section]
\newtheorem{proposition}[theorem]{Proposition}
\newtheorem{lemma}[theorem]{Lemma}

\newtheorem{assumption}[theorem]{Assumption}

\newcommand\norm[1]{\lVert#1\rVert}
\newcommand{\Abs}[1]{\left|#1\right|}

\title{Entropy-Guided Sampling of Flat Modes in Discrete Spaces }

\author{
  Pinaki Mohanty\textsuperscript{1}, 
  Riddhiman Bhattacharya\textsuperscript{2}, 
  Ruqi Zhang\textsuperscript{1} \\
  \textsuperscript{1}Department of Computer Science, College of Science \& College of Engineering,\\
  Purdue University, West Lafayette, IN, USA \\
  \textsuperscript{2}Department of Biostatistics \& Bioinformatics, Duke University School of Medicine,\\
  Duke University, Durham, NC, USA \\
  \texttt{\{pmohanty, ruqiz\}@purdue.edu}, \texttt{riddhiman.bhattacharya@duke.edu }
}

\begin{document}

\maketitle

\begin{abstract}
Sampling from flat modes in discrete spaces is a crucial yet underexplored problem. Flat modes represent robust solutions and have broad applications in combinatorial optimization and discrete generative modeling. However, existing sampling algorithms often overlook the mode volume and struggle to capture flat modes effectively. To address this limitation, we propose \emph{Entropic Discrete Langevin Proposal} (EDLP), which incorporates local entropy into the sampling process through a continuous auxiliary variable under a joint distribution. The local entropy term guides the discrete sampler toward flat modes with a small overhead. We provide non-asymptotic convergence guarantees for EDLP in locally log-concave discrete distributions. Empirically, our method consistently outperforms traditional approaches across tasks that require sampling from flat basins, including Bernoulli distribution, restricted Boltzmann machines, combinatorial optimization, and binary neural networks.
\end{abstract}

\section{Introduction}
\begin{wrapfigure}{r}{0.40\textwidth}
\vspace{-0.8em}
    \centering
    \includegraphics[width=0.45\textwidth, height=0.35\textwidth]{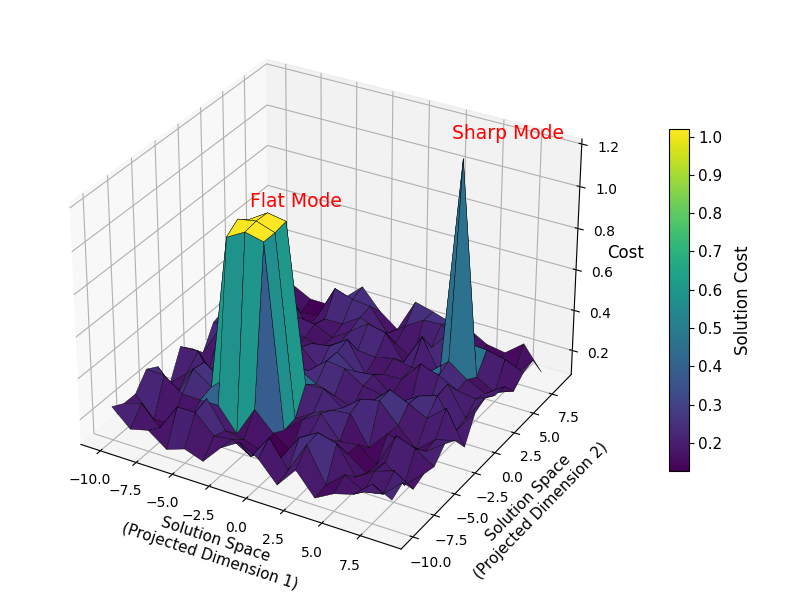}
    \caption{Cost landscape visualization on Traveling Salesman Problem (TSP). Flat modes imply robust solutions under budget, whereas sharp modes are highly sensitive to small changes, leading to abrupt cost increases.}
    \label{fig:motivation}
    \vspace{-1.em} 
\end{wrapfigure}

Discrete sampling is fundamental to many machine learning tasks, such as graphical models, energy-based models, and combinatorial optimization. Efficient sampling algorithms are crucial for navigating the complex probability landscapes of these tasks. Recent advancements in gradient-based methods have significantly enhanced the efficiency of discrete samplers by leveraging gradient information, setting new benchmarks for tasks such as probabilistic inference and combinatorial optimization~\citep{grathwohl2021oops, zhang2022langevin, rhodes2022enhanced, sun2021path, sun2023discrete,li2025reheated}. 

Sampling from flat modes in discrete spaces is a critical yet underexplored challenge. Flat modes, regions where neighboring states have similar probabilities, arise frequently in applications such as energy-based models and neural networks \citep{hochreiter1997flat,arbel2021generalized}. These regions not only represent mode parameter configurations with high generalization performance~\citep{hochreiter1997flat}, but they are also important in constrained combinatorial optimization tasks, where finding structurally similar solutions under a budget is required (see Figure \ref{fig:motivation} for illustration). While there has been growing interest in addressing flat regions in continuous spaces, particularly for tasks like neural network optimization and Bayesian deep learning~\citep{li2024entropymcmc, izmailov2021bayesian,chaudhari2019entropy}, the discrete counterpart remains largely unexplored, highlighting a significant gap.

In this paper, we propose \emph{Entropic Discrete Langevin Proposal} (EDLP), that incorporates the concept of flatness-aware local entropy ~\citep{baldassi2016unreasonable} into Discrete Langevin Proposal (DLP) \citep{zhang2022langevin}. By coupling discrete and flat-mode-guided variables, we obtain a broader, entropy-informed joint target distribution that biases sampling towards flat modes. Specifically, while updating the primary discrete variable using DLP, we simultaneously perform continuous Langevin updates on the auxiliary variable. Through the interaction between discrete and auxiliary variables, the discrete sampler will be steered toward flat regions. We summarize our contributions as follows:
\begin{itemize}
    \item We propose Entropic DLP (EDLP), an entropy-guided, gradient-based proposal for sampling discrete flat modes. EDLP efficiently incorporates local entropy guidance by coupling discrete and continuous variables within a joint distribution.
    
    \item  We provide non-asymptotic convergence guarantees for EDLP in locally log-concave distributions, offering the first such bound for unadjusted gradient-based discrete sampling.
    
    \item Through extensive experiments, we demonstrate that EDLP outperforms existing discrete samplers in capturing flat-mode configurations across various tasks, including Ising models, restricted Boltzmann machines, combinatorial optimization, and binary Bayesian neural networks. We release the code at \url{https://github.com/pmohanty98/EDLP}. 
\end{itemize}

\section{Related Works}
\textbf{Gradient-Based Discrete Sampling.}
Gradient-based methods have significantly improved sampling efficiency in discrete spaces. Locally informed proposals method by \citet{zanella2020informed} leverages probability ratios to explore discrete spaces more effectively. Building on this, \citet{grathwohl2021oops} introduced a gradient-based approach to approximate the probability ratio, further improving sampling efficiency. Discrete Langevin Proposal (DLP), introduced by \citet{zhang2022langevin}, adapts the principles of the Langevin algorithm \citep{grenander1994representations, roberts1996exponential, roberts2002langevin}, originally designed for continuous spaces, to discrete settings. This algorithm enables parallel updates of multiple coordinates using a single gradient computation, boosting both computational efficiency and scalability.

\textbf{Flatness-aware Optimization.}
In early neural network optimization, flatness in energy landscapes emerged as crucial for improving generalization. \citet{hochreiter1994simplifying} linked flat minima to better generalization due to their robustness to parameter perturbations. \citet{ritter1988flat} further emphasized the stability advantages of flat regions. Further, \citet{lecun1990optimal} linked learning algorithm stability to flatness, suggesting optimization methods to exploit this. Later, \citet{gardner1989training} analyzed training algorithms using a statistical mechanics framework, highlighting energy landscape topology’s role. In Bayesian deep learning, \citet{li2024entropymcmc} introduced Entropy MCMC (EMCMC) to bias posterior sampling towards flat regions, achieving better generalization of Bayesian neural networks. 

Our EDLP differs from existing works by targeting flat modes in discrete distributions. A key algorithmic innovation lies in bridging discrete and continuous spaces. 
% Enabling gradient-based updates by calculating gradients as the difference between discrete and continuous states provides finer granularity for exploring discrete states. 
This allows the sampler to explore intermediate regions between discrete states and gain a richer understanding of the discrete landscape, enhancing its ability to sample effectively from flat modes. Further, to our knowledge, we are the first to provide non-asymptotic results for DLP-type algorithms without the MH step, as established in Theorem \ref{thm:edula}, addressing a critical gap in the literature.

\section{Preliminaries \label{sec:prelim}}
\textbf{Target Distribution.}
We define a target distribution over a discrete space using an energy function. The target distribution is given by $
\pi(\bm{\theta}) = \frac{1}{Z} \exp(U(\bm{\theta})),$
where $\bm{\theta}$ is a \(d\)-dimensional discrete variable within domain $\bm{\Theta}$, \(U(\bm{\theta})\) represents the energy function, and \(Z\) is the normalizing constant ensuring \(\pi(\bm{\theta})\) is a proper probability distribution. We make the following assumptions consistent with the literature on gradient-based discrete sampling \citep{grathwohl2021oops, sun2021path, zhang2022langevin}:
1. The domain \(\bm{\Theta}\) is factorized coordinatewisely  i.e. \(\bm{\Theta} = \Pi_{i=1}^d \bm{\Theta}_i\). 
2. The energy function \(U\) can be extended to a differentiable function in \(\mathbb{R}^d\). This extension is crucial for applying gradient-based sampling methods, as it allows the use of gradient information.

\textbf{Langevin Algorithm.}
In continuous spaces, the Langevin algorithm is a powerful sampling method that follows a Langevin diffusion to update variables: $ \bm{\theta'}_{k+1} = \bm{\theta}_k + \frac{\alpha}{2}\nabla U(\bm{\theta}_k) + \sqrt{\alpha}\bm{\epsilon}_k,$ where $\bm{\epsilon}_k \sim \mathcal{N}\left(\textbf{0}, \textbf{I}_{d \times d}\right)$. The gradient assists the sampler in efficiently exploring high-probability regions.

\textbf{Discrete Langevin Proposal.}
The Discrete Langevin Proposal (DLP) is an extension of the Langevin algorithm tailored for discrete spaces, introduced by \citet{zhang2022langevin}. At a given position $\bm{\theta}$, the proposal distribution $q(\cdot|\bm{\theta})$ determines the next position. The proposal distribution in DLP is formulated as:
\begin{align}
    q(\bm{\theta}'|\bm{\theta}) &= \frac{\exp\left(-\frac{1}{2\alpha}\|\bm{\theta}' - \bm{\theta} - \frac{\alpha}{2}\nabla U(\bm{\theta})\|^2\right)}{Z_{\bm{\Theta}}(\bm{\theta})},
    \label{eq:lang-proposal}
\end{align}
where $Z_{\bm{\Theta}}(\bm{\theta})$ is the normalizing constant. DLP can be employed without or with a Metropolis-Hastings (MH) step, resulting in the discrete unadjusted Langevin algorithm (DULA) and the discrete Metropolis-adjusted Langevin algorithm (DMALA), respectively.

\textbf{Local Entropy.}
Local entropy is a critical concept in flatness-aware optimization techniques, which is used to understand the geometric characteristics of energy landscapes \citep{baldassi2016unreasonable,chaudhari2019entropy,baldassi2019shaping}. It is defined as:
\begin{equation}\label{eq:local_entropy}
    \mathcal{F}(\bm{\theta}_a; \eta) = \log\left(\sum_{\bm{\theta} \in \bm{\Theta}} \exp\left\{U(\bm{\theta}) - \frac{1}{2\eta}\|\bm{\theta} - \bm{\theta}_a \|^2\right\}\right),
\end{equation}
where $\eta$ is a scalar parameter controlling the sensitivity to flatness in the landscape. Local entropy provides a measure of the density of configurations around a point, thus identifying regions with high configuration density and flat energy landscapes.

\section{Entropic Discrete Langevin Proposal \label{sec:method}}

\subsection{Target Joint Distribution: Coupling Mechanism}\label{sec:coupling}

We propose leveraging local entropy (Eq.\ref{eq:local_entropy}) to construct an auxiliary distribution that emphasizes flat regions of the target distribution. This auxiliary distribution smoothens the energy landscape, acting as an external force, driving the exploration of flat basins. Figure~\ref{fig:eta_effect} in the Appendix \ref{sec:app_eta} illustrates the motivation behind our approach and the impact of the parameter $\eta$ on the smoothened target distribution.

We start with the original target distribution \( p(\bm{\theta}) \propto \exp(U(\bm{\theta})) \). By incorporating local entropy, we derive a smoothed target distribution in terms of a new variable \(\bm{\theta}_a\):
{%\footnotesize
\begin{equation}
    p(\bm{\theta}_a) \propto \exp{\mathcal{F}(\bm{\theta}_a; \eta)} = \sum_{\bm{\theta} \in \bm{\Theta}} \exp{\left\{ U(\bm{\theta}) - \frac{1}{2\eta} \|\bm{\theta} - \bm{\theta}_a\|^2 \right\}}
    \label{eq:theta_a_posterior}
\end{equation}
}

Inspired by the coupling method introduced by \citet{li2024entropymcmc} in their Section 4.1, we couple $\bm{\theta}$ and $\bm{\theta}_a$ as follows:

\begin{lemma}\label{lemma:joint_posterior}
Given \(\widetilde{\bm{\theta}} = [\bm{\theta}^T, \bm{\theta}_a^T]^T \in \bm{\Theta}\times \mathbb{R}^{d}\), the joint distribution p(\(\widetilde{\bm{\theta}}\)) is:
\begingroup
%\footnotesize
\begin{equation}
    p(\widetilde{\bm{\theta}}) = p(\bm{\theta}, \bm{\theta}_a) \propto \exp\left\{ U(\bm{\theta}) - \frac{1}{2\eta} \|\bm{\theta} - \bm{\theta}_a\|^2 \right\}
    \label{eq:joint_posterior}
\end{equation}
\endgroup
By construction, the marginal distributions of \(\bm{\theta}\) and \(\bm{\theta}_a\) are the original distribution \( p(\bm{\theta}) \) and the smoothed distribution \( p(\bm{\theta}_a) \) (Eq.~\ref{eq:theta_a_posterior}). 
\end{lemma}
This result directly follows from Lemma 1 under Section 4.1 in \citet{li2024entropymcmc}. The joint hybrid-variable, \(\widetilde{\bm{\theta}}\) lies in a product space where first d coordinates are discrete-valued and the remaining d coordinates lie in $\mathbb{R}^d$. Consequently, the energy function of  \(\widetilde{\bm{\theta}}\) becomes \( U(\widetilde{\bm{\theta}}) = U(\bm{\theta}) - \frac{1}{2\eta} \|\bm{\theta} - \bm{\theta}_a\|^2 \), and its gradient is given by:
{%\footnotesize
\begingroup
\begin{equation}\label{eq:grad}
    \nabla_{\widetilde{\bm{\theta}}} U_{\eta}(\widetilde{\bm{\theta}}) = \left[
        \begin{array}{c}
        \nabla_{\bm{\theta}} U_{\eta}(\widetilde{\bm{\theta}}) \\
        \nabla_{\bm{\theta}_a} U_{\eta}(\widetilde{\bm{\theta}})
        \end{array}
    \right] = \left[
        \begin{array}{c}
        \nabla_{\bm{\theta}} U(\bm{\theta}) - \frac{1}{\eta} (\bm{\theta} - \bm{\theta}_a) \\
        \frac{1}{\eta} (\bm{\theta} - \bm{\theta}_a)
        \end{array}
    \right].
\end{equation}
\endgroup
}
\subsection{Sampling Algorithm: Local Entropy Guidance in Discrete Langevin Proposals}
We propose EDLP, an extension of DLP designed to enhance sampling efficiency from flat modes. In our framework (Algorithm \ref{alg:edlp}), the Langevin update for \(\bm{\theta}_a\) follows the distribution $
q_{\alpha_a}(\bm{\theta'}_a|\widetilde{\bm{\theta}})$:
\
\begin{equation}\label{eq:prop1}
q_{\alpha_a}(\bm{\theta'}_a|\widetilde{\bm{\theta}}) =\frac{1}{\sqrt{2\pi\alpha_a}^d} \exp\left( -\frac{1}{2\alpha_a} \| \bm{\theta'}_a - \bm{\theta}_a - \frac{\alpha_a}{2} \nabla_{\bm{\theta}_a} U_{\eta}(\widetilde{\bm{\theta}}) \|^2 \right).
\end{equation}

Unlike the standard DLP, where transitions are purely between discrete states, EDLP leverages the current joint variables $\widetilde{\bm{\theta}} = [\bm{\theta}^T, \bm{\theta}_a^T]^T$ to propose the next discrete state. By incorporating the coupling between the variables, we refine the DLP proposal by replacing \(\nabla U(\bm{\theta})\) with \(\nabla_{\bm{\theta}} U_{\eta}(\widetilde{\bm{\theta}})\). This adjustment results in the modified proposal:
\begin{equation}\label{eq:prop2}
    q_{\alpha}(\bm{\theta'}|\widetilde{\bm{\theta}}) \propto \exp\left(-\frac{1}{2\alpha} \| \bm{\theta'} - \bm{\theta} - \frac{\alpha}{2} \nabla_{\bm{\theta}} U_{\eta}(\widetilde{\bm{\theta}}) \|^2\right).
\end{equation}
To further simplify, we use coordinate-wise factorization from DLP to obtain $
 q_{\alpha}(\bm{\theta'}|\widetilde{\bm{\theta}}) = \prod_{i=1}^{d} q_{\alpha_i}(\theta_i'|\widetilde{\bm{\theta}})$, where \( q_{\alpha_i}(\theta_i'|\widetilde{\bm{\theta}}) \) is a categorical distribution:
\begingroup
\begin{align}
   \text{Cat}\left(\text{Softmax}\left(\frac{1}{2} \nabla_{\theta} U_{\eta}(\widetilde{\bm{\theta}})_i (\theta_i' - \theta_i) - \frac{(\theta_i' - \theta_i)^2}{2\alpha}\right)\right).
   \label{eq:catprop2}
\end{align}
\endgroup
By synthesizing Equations \eqref{eq:prop1} and \eqref{eq:catprop2}, we derive the full proposal distribution:
\begingroup
\begin{align}{\label{eq:proposal}}
q_{\gamma}(\widetilde{\bm{\theta'}}|\widetilde{\bm{\theta}}) \propto & q_{\alpha}(\bm{\theta'}|\widetilde{\bm{\theta}}) q_{\alpha_a}(\bm{\theta'}_a|\widetilde{\bm{\theta}}) 
\end{align}
\endgroup
where $\gamma=(\alpha, \alpha_a)$. 

This factorized proposal in Eq.~\eqref{eq:proposal} is purely a design choice to simplify sampling. The proposal distribution is called the \emph{Entropic Discrete Langevin Proposal} (EDLP). At the current joint position $\widetilde{\bm{\theta}}$, EDLP generates the next joint position. EDLP can be paired with or without a Metropolis-Hastings step \citep{metropolis1953equation, hastings1970monte} to ensure the Markov chain's reversibility. These algorithms are referred to as EDULA (Entropic Discrete Unadjusted Langevin Algorithm) and EDMALA (Entropic Discrete Metropolis-Adjusted Langevin Algorithm), respectively. We will collect samples of \(\bm{\theta}\), as the marginal distribution of \(p(\widetilde{\bm{\theta}})\) over \(\bm{\theta}\) yields our desired discrete target distribution. 

Alongside the vanilla EDLP, we introduce a computationally efficient \emph{Gibbs-like-update} (GLU) version, in the Appendix \ref{sec:glu}, which involves alternating updates instead of simultaneous updates of our variables. We provide a sensitivity analysis of the hyperparameters in Appendix \ref{sec:app_eta}.
\begin{algorithm}[t]
% \footnotesize
  \caption{Entropic Discrete Langevin Proposal: EDULA and EDMALA}
  \begin{algorithmic}
    \label{alg:edlp}
    \STATE \textbf{Inputs:} Main variable $\bm{\theta} \in \bm{\Theta}$ , Auxiliary variable $\bm{\theta}_a \in \mathbb{R}^d$, Main stepsize $\alpha$, Auxiliary stepsize $\alpha_{a}$, Flatness parameter $\eta$
    \STATE \textbf{Initialize:} $\bm{\theta}_a \gets \bm{\theta}, \mathcal{S} \gets \emptyset$
    
    \LOOP
      \STATE \textbf{Construct} $\nabla_{\widetilde{\bm{\theta}}}U_{\eta}(\widetilde{\bm{\theta}})$ as in Equation \eqref{eq:grad}
      \FOR{$i = 1$ \textbf{to} $d$}
        \STATE \textbf{Construct} ${q_{i}}_{\alpha}(\cdot|\widetilde{\bm{\theta}})$ as in Equation~\eqref{eq:catprop2}
        \STATE \textbf{Sample} ${\bm{\theta}_i}' \sim {q_{i}}_{\alpha}(\cdot|\widetilde{\bm{\theta}})$
      \ENDFOR
      \STATE \textbf{Compute}  $\bm{\theta'_a} \gets \bm{\theta}_a + \frac{\alpha_{a}}{2}\nabla_{\bm{\theta}_a}U_{\eta}(\widetilde{\bm{\theta}}) + \sqrt{\alpha_{a}}\bm{\epsilon}$ 
      \quad \text{where }{$\bm{\epsilon} \sim \mathcal{N}(\bm{0}, \bm{I})$}
      
      \vspace{0.5em}
      \STATE $\triangleright$ Optionally, do the MH step
      \STATE \textbf{Compute} $q_{\alpha}(\widetilde{\bm{\theta'}}|\widetilde{\bm{\theta}}) = \prod_i {q_i}_{\alpha}(\widetilde{\bm{\theta'_i}}|\widetilde{\bm{\theta}})$
      \STATE \hspace{4em} and $q_{\alpha}(\widetilde{\bm{\theta}}|\widetilde{\bm{\theta'}}) = \prod_i {q_i}_{\alpha}(\widetilde{\bm{\theta_i}}|\widetilde{\bm{\theta'}})$
      \STATE \textbf{Set} $\bm{\theta} \leftarrow \bm{\theta'}$ and $\bm{\theta}_a \gets \bm{\theta'_a}$ with probability 
      
      \[
      \min\left(1,  \frac{q_{\alpha}({\bm{\theta}}|\widetilde{\bm{\theta'}})}{q_{\alpha}({\bm{\theta'}}|\widetilde{\bm{\theta}})}\frac{q_{\alpha_a}({\bm{\theta_a}}|\widetilde{\bm{\theta'}})}{q_{\alpha_a}({\bm{\theta_a'}}|\widetilde{\bm{\theta}})}\frac{\pi(\widetilde{\bm{\theta'}})}{\pi(\widetilde{\bm{\theta}})}\right)
      \].
    \IF{\textit{after burn-in}}
      \STATE \textbf{Update} 
        $\mathcal{S} \gets \mathcal{S}\cup\{\bm{\theta}\}$
    \ENDIF
    \ENDLOOP
    \STATE \textbf{Output:} $\mathcal{S}$
  \end{algorithmic}
\end{algorithm}

\section{Theoretical Analysis \label{sec:theory}}
In this section, we provide a theoretical analysis of the convergence rate of EDLP i.e. EDULA and EDMALA. We make similar assumptions as \citet{pynadath2024gradientbased}. Those are as follows,

\begin{assumption} \label{assumption:5.1}
The function $U(\cdot) \in C^2(\mathbb{R}^d)$ has $M$-Lipschitz gradient.
\end{assumption}

\begin{assumption} \label{assumption:5.2}
For each $\bm{\theta} \in \mathbb{R}^d$, there exists an open ball containing $\bm{\theta}$ of some radius $r_{\bm{\theta}}$, denoted by $B(\bm{\theta}, r_{\bm{\theta}})$, such that the function $U(\cdot)$ is $m_{\bm{\theta}}$-strongly concave in $B(\bm{\theta}, r_{\bm{\theta}})$ for some $m_{\bm{\theta}} > 0$. 
\end{assumption}

\begin{assumption} \label{assumption:5.3}
$\bm{\theta}_a$ is restricted to a compact subset of $\mathbb{R}^d$ labeled $\bm{\Theta}_a$.
\end{assumption}
We define $\text{diam}(\bm{\Theta}) = \sup_{\bm{\theta}, \bm{\theta}' \in \bm{\Theta}} \|\bm{\theta} - \bm{\theta}'\|$, and $\text{diam}(\bm{\Theta_a}) = \sup_{\bm{\theta}_a, \bm{\theta}_a' \in \bm{\Theta}_a} \|\bm{\theta}_a - \bm{\theta}_a'\|$. Let $\vartheta{(\bm{\Theta}, \bm{\Theta_a})}=\inf_{\substack{\bm{\theta}, \bm{\theta}' \in \bm{\Theta}; \bm{\theta}_a,\bm{\theta}_a' \in \bm{\Theta_a} }} (\bm{\theta} - \bm{\theta}_a)^{\top}(\bm{\theta}' - \bm{\theta}_a')$  and $\Delta(\bm{\Theta},\bm{\Theta}_a)=\sup_{\bm{\theta} \in \bm{\Theta},\,  \bm{\theta_a} \in \bm{\Theta_a}} \|\bm{\theta_a}-\bm{\theta}\| $.
Let the joint valid bounded space be $\widetilde{\bm{\Theta}}$ and finally define $ a \in \arg\min_{\bm{\theta} \in \bm{\Theta}} \|\nabla U(\bm{\theta})\|$ as the set of values which minimizes the energy function in $\bm{\Theta}$.

Assumptions \ref{assumption:5.1} ,\ref{assumption:5.2}, and \ref{assumption:5.3} are standard in optimization and sampling literature \cite{bottou2018optimization,dalalyan2017further, durmus2017nonasymptotic}. Under Assumption \ref{assumption:5.2}, $U(\cdot)$ is $m$-strongly concave on $\text{conv}(\bm{\Theta})$, following Lemma C.3 from \citet{pynadath2024gradientbased}. The total variation distance between two probability measures $\mu$ and $\nu$, defined on some space $\bm{\theta} \subset \mathbb{R}^d$ is$\|\mu - \nu\|_{TV} = \sup_{A \subseteq B(\bm{\theta})} |\mu(A) - \nu(A)|$ where $B(\bm{\theta})$ is the set of all measurable sets in $\bm{\theta}$.
\subsection{Convergence Analysis for EDULA}
Since EDULA does not have the target as the stationary distribution, we establish mixing bounds for it in two steps. 
%To our knowledge, we are the first to provide non-asymptotic results for DLP-type algorithms without the MH step, as established in Theorem \ref{thm:edula}.
We first prove that when both the stepsizes ($\alpha$ , $\alpha_a$) tend to zero, the asymptotic bias of EDULA is zero for target distribution $\tilde{\pi}{(\widetilde{\bm{\theta}})} \propto e^{(U(\bm{\theta})-\frac{1}{2\eta}\norm{\bm{\theta}-\bm{\theta}_a}^2)}$.

\begin{proposition}\label{prop:1}
    Under Assumptions~\ref{assumption:5.1}, and~\ref{assumption:5.3}, the Markov chain as defined in ~\eqref{eq:proposal} 
is reversible with respect to some distribution $\pi_{\gamma}$ and $\pi_{\gamma}$ converges weakly to $\pi$ as $\alpha\rightarrow 0$ and $\alpha_a \rightarrow 0$. Further, for any $\alpha>0, \alpha_a>0$,
\[
% \resizebox{0.5\hsize}{!}{$
\norm{\pi_{\gamma} -\tilde{\pi}}_1\le 
     Z\exp\left(\frac{M}{4} -\frac{1}{2\alpha}+\frac{\Delta(\bm{\Theta}, \bm{\Theta_a})^2-\vartheta{(\bm{\Theta}, \bm{\Theta_a})}}{2\eta}\right),
% $}
\]
where $Z$ is the normalizing constant of $\pi(\bm{\theta})$.
\end{proposition}
The parameter $\alpha_a$ is consumed during the computation of the stationary distribution $\pi_\gamma$, explicitly not appearing in the bound. However, $\alpha_a$ indirectly influences the geometric terms $\Delta(\bm{\Theta}, \bm{\Theta}_a)$ and $\vartheta(\bm{\Theta}, \bm{\Theta}_a)$. Larger $\alpha_a$ increases $\Delta^2(\bm{\Theta}, \bm{\Theta}_a)$ due to a greater diameter and reduces $\vartheta(\bm{\Theta}, \bm{\Theta}_a)$ due to weaker alignment, thereby loosening the bound. In contrast, smaller $\alpha_a$ tightens convergence guarantees. This parallels the observable role of $\alpha$ in the bound i.e. bias vanishes to 0 as $\alpha \rightarrow 0$.
Next we establish our main result for EDULA which levarages Proposition~\ref{prop:1} and the ergodicity of the EDULA chain, as a consequence of Lemma~\ref{lemma:minorization} in the Appendix.
\begin{theorem}\label{thm:edula}
Under Assumptions \ref{assumption:5.1}, and \ref{assumption:5.3} ,  in Algorithm \ref{alg:edlp}, Markov chain P exhibits,
\[
% \resizebox{0.7\hsize}{!}{$
\|P^k(x, \cdot) - \tilde{\pi}\|_{TV} \leq (1 - \bar{\eta}^*)^k + 
    Z \exp\left(\frac{M}{4} - \frac{1}{2\alpha} + 
    \frac{\Delta(\bm{\Theta}, \bm{\Theta_a})^2 - \vartheta(\bm{\Theta}, \bm{\Theta_a})}{2\eta}\right)
% $}
\]
\end{theorem}
where $\bar{\eta}^*$ is a constant that can be explicitly computed (see \eqref{eta:minor:def} in the Appendix). In essence, $\bar{\eta}^*= f(\alpha,\alpha_a,\text{diam}(\bm{\Theta}),\text{diam}(\bm{\Theta}_a),\Delta(\bm{\Theta}_a,\bm{\Theta}))$, where $f$ is increasing exponentially in the first two arguments and decreasing exponentially in the last three arguments. Theorem~\ref{thm:edula} shows that sufficiently small learning rates bring the samples generated by Algorithm~\ref{alg:edlp} closer to the target distribution. However, excessively small rates hinder convergence by limiting exploration, while large rates cause the sampler to overshoot the target. Thus, choosing an appropriate learning rate is critical for balancing exploration and convergence. 

\subsection{Convergence Analysis for EDMALA}
We establish a non-asymptotic convergence guarantee for EDMALA using a uniform minorization argument.
\begin{theorem}\label{thm:edmala}
Under Assumptions \ref{assumption:5.1} ,\ref{assumption:5.2}, and \ref{assumption:5.3} , and $\alpha < \frac{2}{M}$ in Algorithm \ref{alg:edlp}, Markov chain P is uniformly ergodic under,
\[
% \resizebox{0.3\hsize}{!}{$
\norm{P^k (x,\cdot) -\tilde{\pi}}_{TV}\le (1-\epsilon_{\gamma})^k
% $}
\]
where, 
\resizebox{0.92\hsize}{!}{$
\epsilon_{\gamma} = 
\exp \left\{
    \begin{aligned}
        &-\left(\frac{M}{2} + \frac{1}{\alpha} - \frac{m}{4}\right) \text{diam}(\bm{\Theta})^2 - \frac{1}{2} \|\nabla U(a)\| \, \text{diam}(\bm{\Theta})- \left(\frac{3\alpha_a}{8\eta^2} + \frac{2}{\eta}\right) \Delta(\bm{\Theta}, \bm{\Theta}_a)^2 + \frac{\vartheta(\bm{\Theta}, \bm{\Theta}_a)}{\eta}
    \end{aligned}
\right\}$}
\end{theorem}

One notices, $\epsilon_{\gamma}$ is exponentially decreasing in the size of the set, $\bm{\Theta}$, its distance from $\bm{\Theta}_a$. Further, as $\alpha \rightarrow 0$, $\epsilon_{\gamma} \rightarrow 0$, causing the convergence factor $1 - \epsilon_{\gamma}$ to approach 1. This slows the convergence rate, as the chain takes longer to approach the stationary distribution.

One notices, for $\eta \rightarrow \infty$ (weaker coupling), the bounds in Proposition \ref{prop:1} and Theorem \ref{thm:edmala} align with those of DULA \cite{zhang2022langevin} and DMALA \citep{pynadath2024gradientbased}, respectively. Note that the convergence of the chains for both EDULA and EDMALA imply convergence of the marginals as the projection maps are continuous. In fact, deriving a rate of convergence for them is also possible, but we omit it here as that is not the goal of this paper.

\section{Experiments \label{sec:exp}}
We conducted an empirical evaluation of the Entropic Discrete Langevin Proposal (EDLP) to demonstrate its effectiveness in sampling from flat regions compared to existing discrete samplers. Our experimental setups mainly follow \citet{zhang2022langevin}. EDLP is benchmarked against a range of popular baselines, including Gibbs sampling, Gibbs with Gradient (GWG) \citep{grathwohl2021oops}, Hamming Ball (HB) \citep{titsias2017hamming}, Discrete Unadjusted Langevin Algorithm (DULA), and Discrete Metropolis-Adjusted Langevin Algorithm (DMALA)~\citep{zhang2022langevin}. For consistency in comparing DLP samplers with their entropic counterparts, we maintain $\alpha$ values across most instances. We retain \citet{zhang2022langevin}’s notation for consistency: Gibbs-$X$ for Gibbs sampling, GWG-$X$ for Gibbs with Gradient, and HB-$X$-$Y$ for Hamming Ball. To the best of our knowledge, fBP \citep{baldassi2016unreasonable} is the only algorithm that targets flat regions in discrete spaces. However, it is not directly comparable to EDLP and the other samplers in our study due to methodological and practical reasons (see Appendix \ref{sec:app_fbp} for details).
\subsection{Motivational Synthetic Example}
\begin{wrapfigure}{r}{0.50\textwidth}
  \vspace{-3.5em}  % pull figure closer to preceding text
  \centering
  \includegraphics[width=0.5\textwidth]{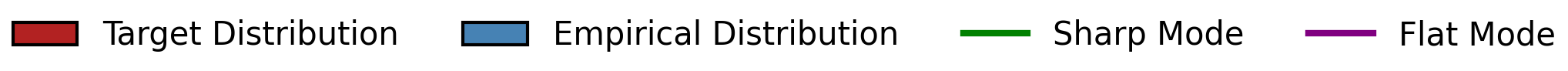}
  \vspace{2pt}
  \begin{tabular}{cc}
    \includegraphics[width=0.22\textwidth, height=3.4cm]{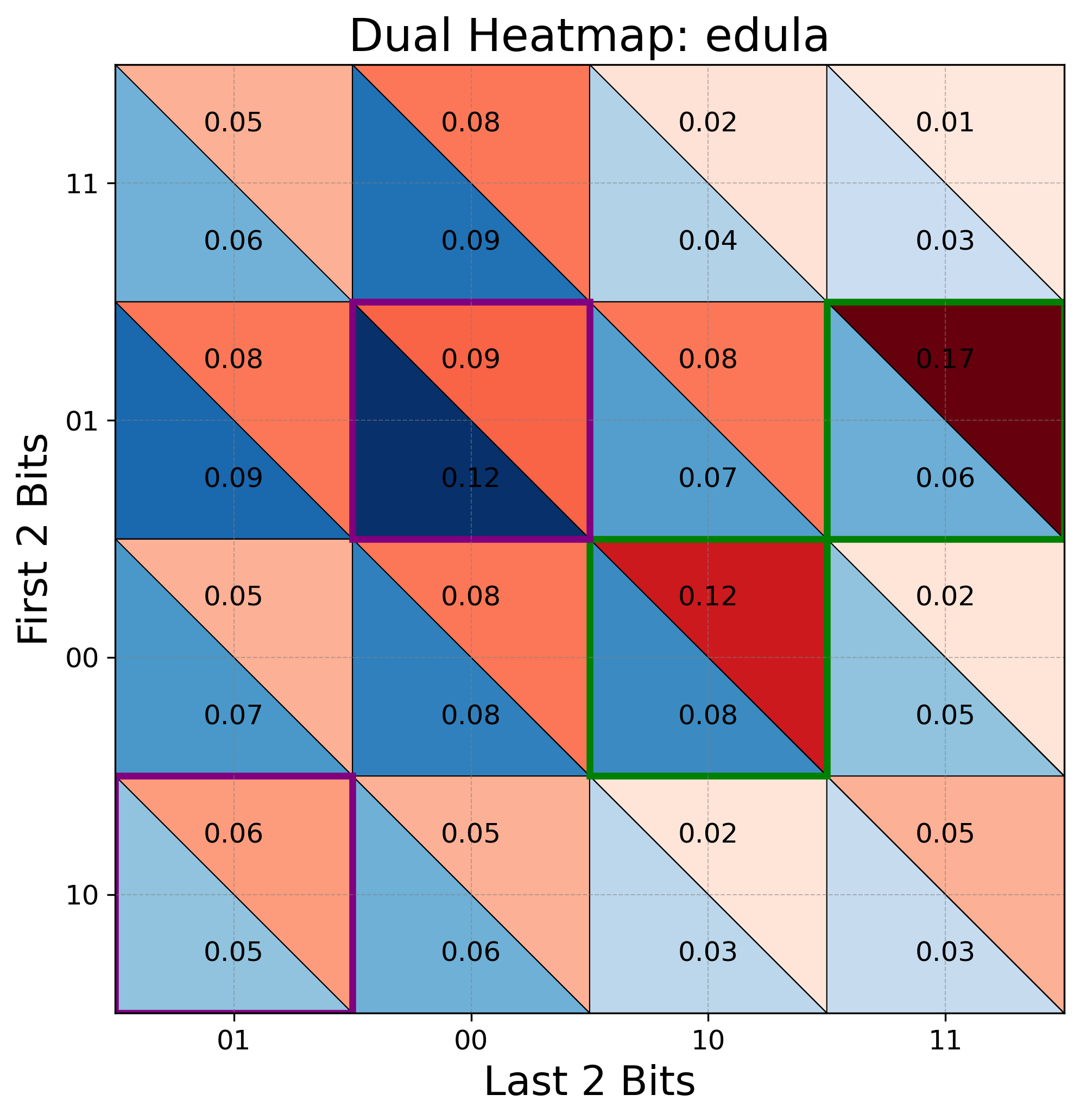} &
    \includegraphics[width=0.22\textwidth, height=3.4cm]{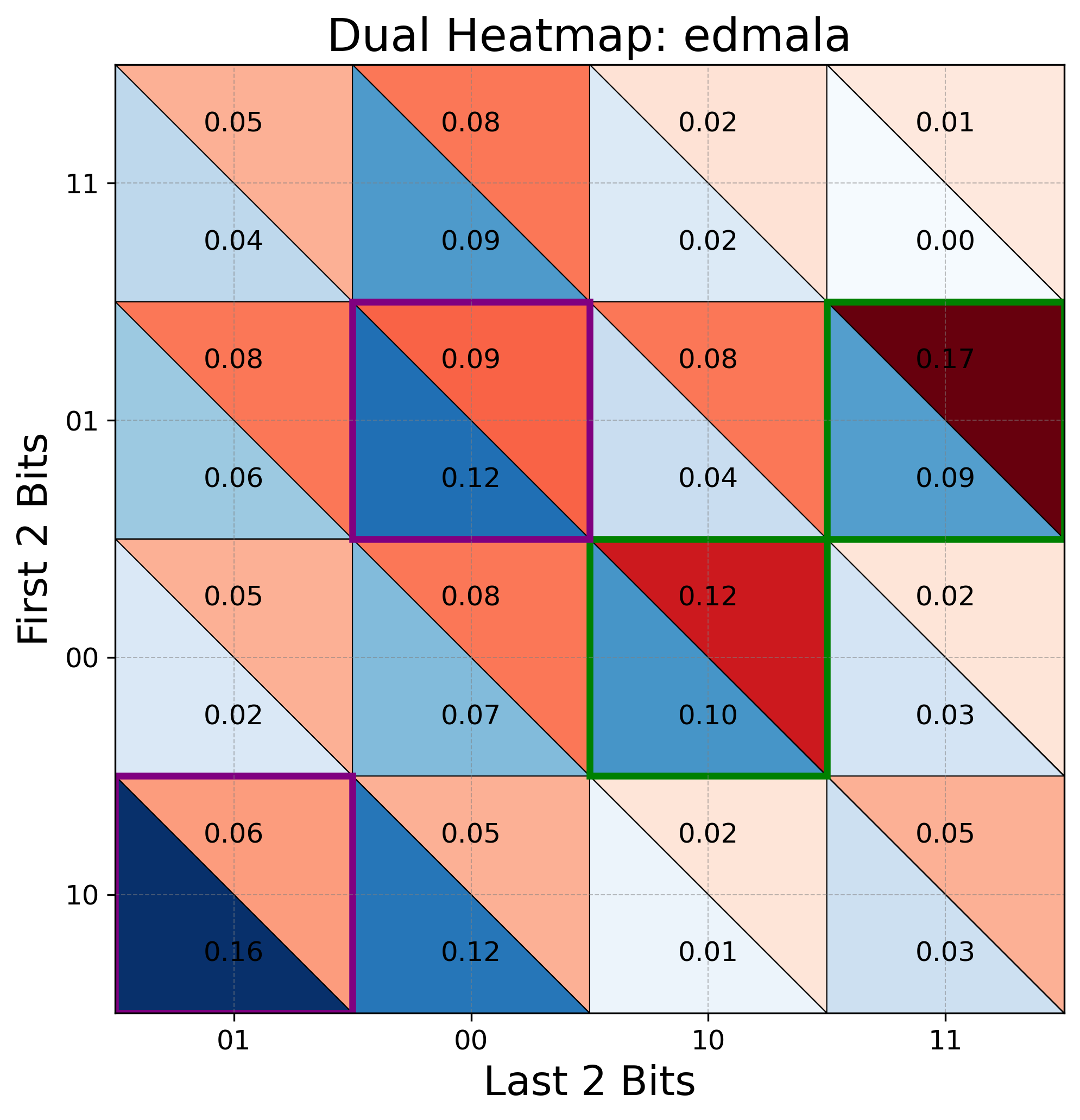} \\
    \includegraphics[width=0.22\textwidth, height=3.4cm]{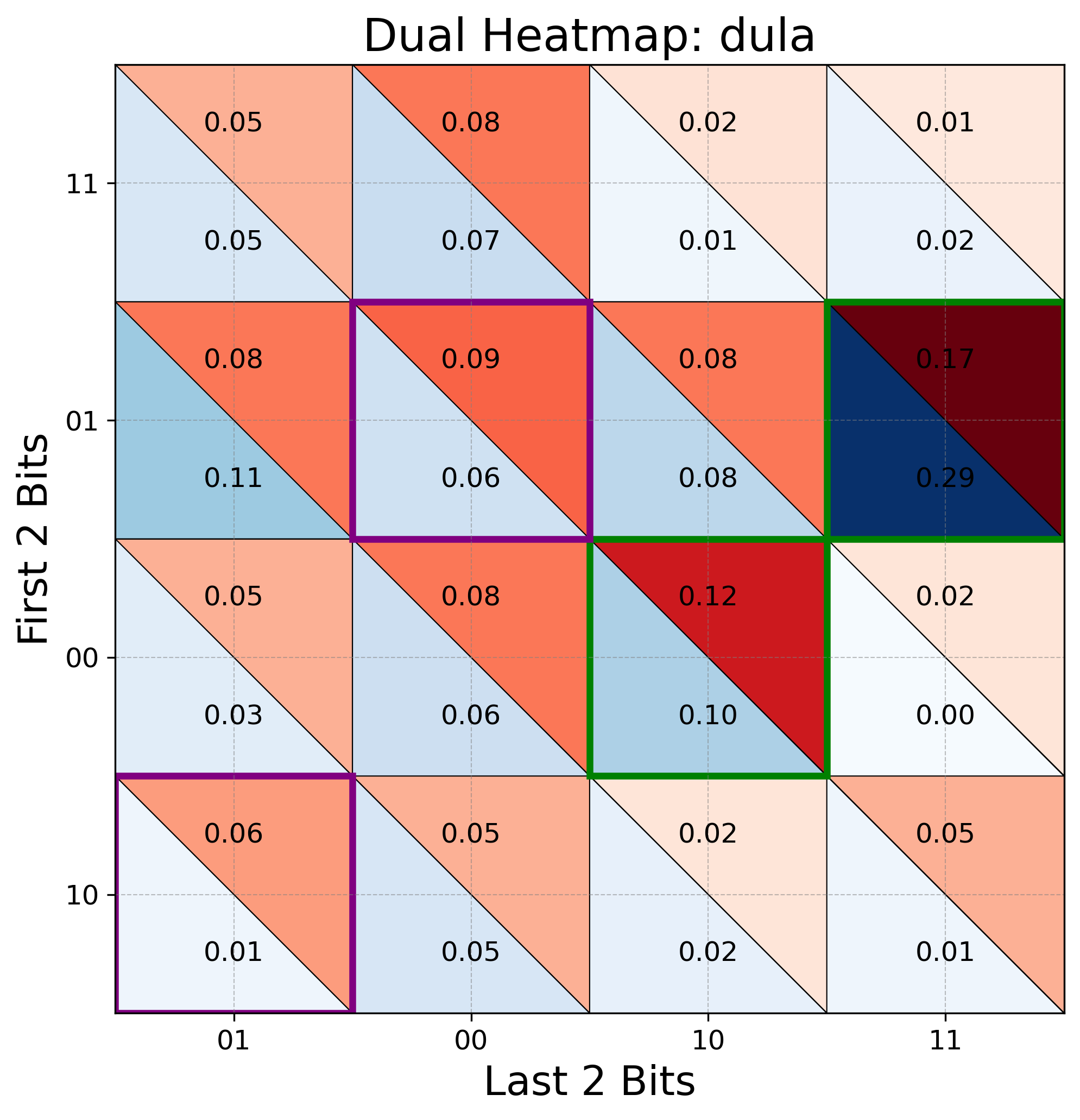} &
    \includegraphics[width=0.22\textwidth, height=3.4cm]{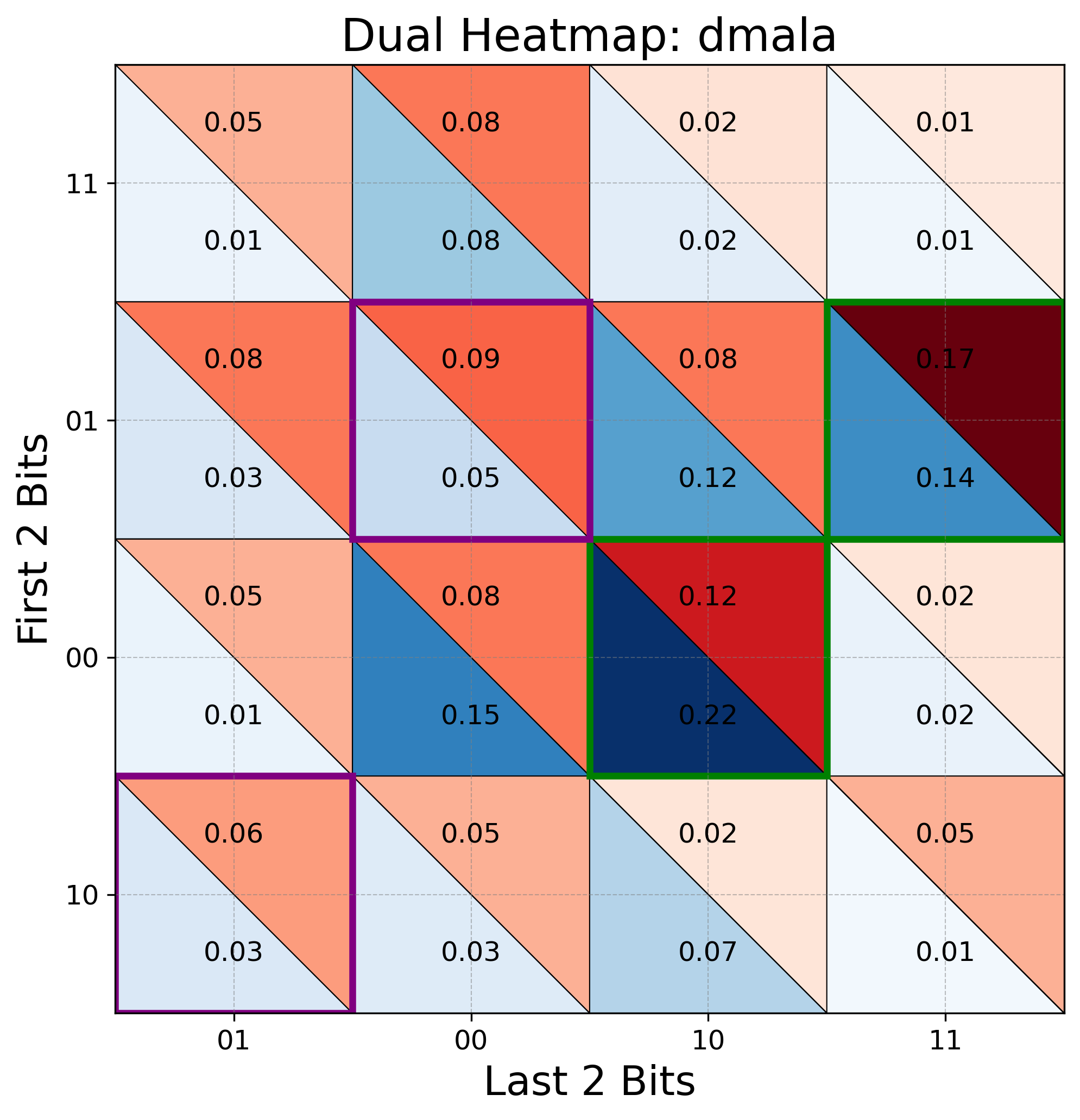}
  \end{tabular}
  \vspace{-1em}
  \caption{\small Overlay Heatmaps for EDULA, EDMALA, DULA, and DMALA.}
  \label{fig:bernoulli_eval}
  \vspace{-2.5em}
\end{wrapfigure}
We consider sampling from a joint quadrivariate Bernoulli distribution. Let \(\bm{\theta} = (\theta_1, \theta_2, \theta_3, \theta_4)\) be a 4-dimensional binary random vector, where each \(\theta_i \in \{0, 1\}\). The joint probability distribution is specified by \(p_{\bm{\theta}}\), which represents the probability of the vector (\( \theta_1, \theta_2, \theta_3, \theta_4 \)). For a given state \(\bm{\theta}\)
then energy function is given by :
\[
U(\bm{\theta}) = \sum_{{a} \in \{0,1\}^4} \left( \prod_{n=1}^4 \theta_n^{a_n} (1 - \theta_n)^{1 - a_n} \right) \ln p_{{a}},
\]

The target distribution over the 4D Joint Bernoulli space contains both sharp and flat modes, each analyzed over their 1-Hamming distance neighborhoods. Sharp modes, such as 0010 and 0111, have high probability mass but are surrounded by neighbors with significantly lower probabilities, indicating steep local gradients. In contrast, flat modes like 0100 and 1001 are characterized by relatively uniform probabilities among their immediate neighbors, reflecting smoother local geometry. For the true target distribution's visualization refer to Figure \ref{fig:bern_target} in Appendix \ref{subsec:app-bern}.
We ran 4 chains of DULA, EDULA, DMALA, and EDMALA in parallel for 1000 iterations, with an initial burn of 200. From Figure~\ref{fig:bernoulli_eval}, EDMALA and EDULA demonstrate a strong preference to visit flat modes,  without becoming stuck in the high-probability sharp modes. In contrast, DULA and DMALA show a bias toward the sharp modes, showing to be less adept at exploring the flat areas where the probability mass is more evenly distributed. Despite showing flatness bias, entropic samplers still achieve well-matching samples to the target distribution.

\subsection{Sampling for Traveling Salesman Problems}

In TSP, the objective is to find the shortest route visiting $n$ cities exactly once and returning to the origin, choosing from $n!$ paths. In practical applications, minimal cost and deviation from the optimal route are often essential for operational consistency. For example, in logistics and delivery services, routes that closely follow the optimal sequence improve loading and unloading efficiency and ensure consistent customer experience \citep{laporte2009fifty, golden2008vehicle}. Minimal sensitivity reduces the cognitive load on drivers who rely on established patterns, which is critical in repetitive, high-volume delivery operations \cite{toth2002vehicle} \cite{Young2007}. Routes with low sensitivity to deviations provide robustness in situations where consistency and predictability are priorities. Thus, sampling from flat modes allows us to propose multiple robust routes that lie within the same cost bracket.

The energy function \( U(\bm{\theta}) \), where $\bm{\theta }$ represents a specific unique route, signifies the weighted sum of the Euclidean distances between consecutive states (cities). In the Traveling Salesman Problem (TSP) and similar optimization problems, \( U(\bm{\theta}) \) is designed to capture the total cost of a particular route configuration \( \bm{\theta} = (\theta_1, \theta_2, \dots, \theta_n) \). 
The mathematical formulation of \( U(\bm{\theta}) \) can be expressed as:
\[
% \resizebox{0.60\hsize}{!}{$ 
U(\bm{\theta}) = -\left(\sum_{i=1}^{n-1} \left(w_{(\theta_i, \theta_{i+1})} \cdot \norm{\theta_i- \theta_{i+1}}\right) + w_{(\theta_n, \theta_1)} \cdot \norm{\theta_n- \theta_1} \right),
% $}
\]
where \( w_{(\theta_i, \theta_{i+1})} \) is a directional weight or scaling factor that allows for non-symmetric costs, accounting for the fact that the cost to travel from city \( \theta_i \) to \( \theta_{i+1} \) may differ from the reverse direction, and the term \( w_{(\theta_n, \theta_1)}\) represents the cost of returning from the last city \( \theta_n \) back to the starting city \( \theta_1 \), thereby completing the tour.

The energy function \( U(\bm{\theta}) \) quantifies the overall cost associated with a given route, based on the weighted Euclidean distances between consecutive cities. Maximizing \( U(\bm{\theta}) \) involves finding the optimal sequence of cities that minimizes the total travel cost. This formulation is particularly useful in real-world applications where different paths may have varying travel costs due to factors like road conditions, transportation constraints, or other contextual variables \citep{golden2008vehicle, laporte2009fifty}.

For our experimental setup, we address the 8-city TSP, where each city is represented as a 3D binary tensor. A valid solution to the TSP ensures that all cities are visited exactly once, and the path returns to the starting city. If a proposed solution violates the uniqueness of city visits, we reject the sample and remain at the current solution.

\begin{wrapfigure}{r}{0.65\textwidth}
  \centering
  \vspace{-15pt}
  \begin{tabular}{cc}
    \includegraphics[width=0.3\textwidth]{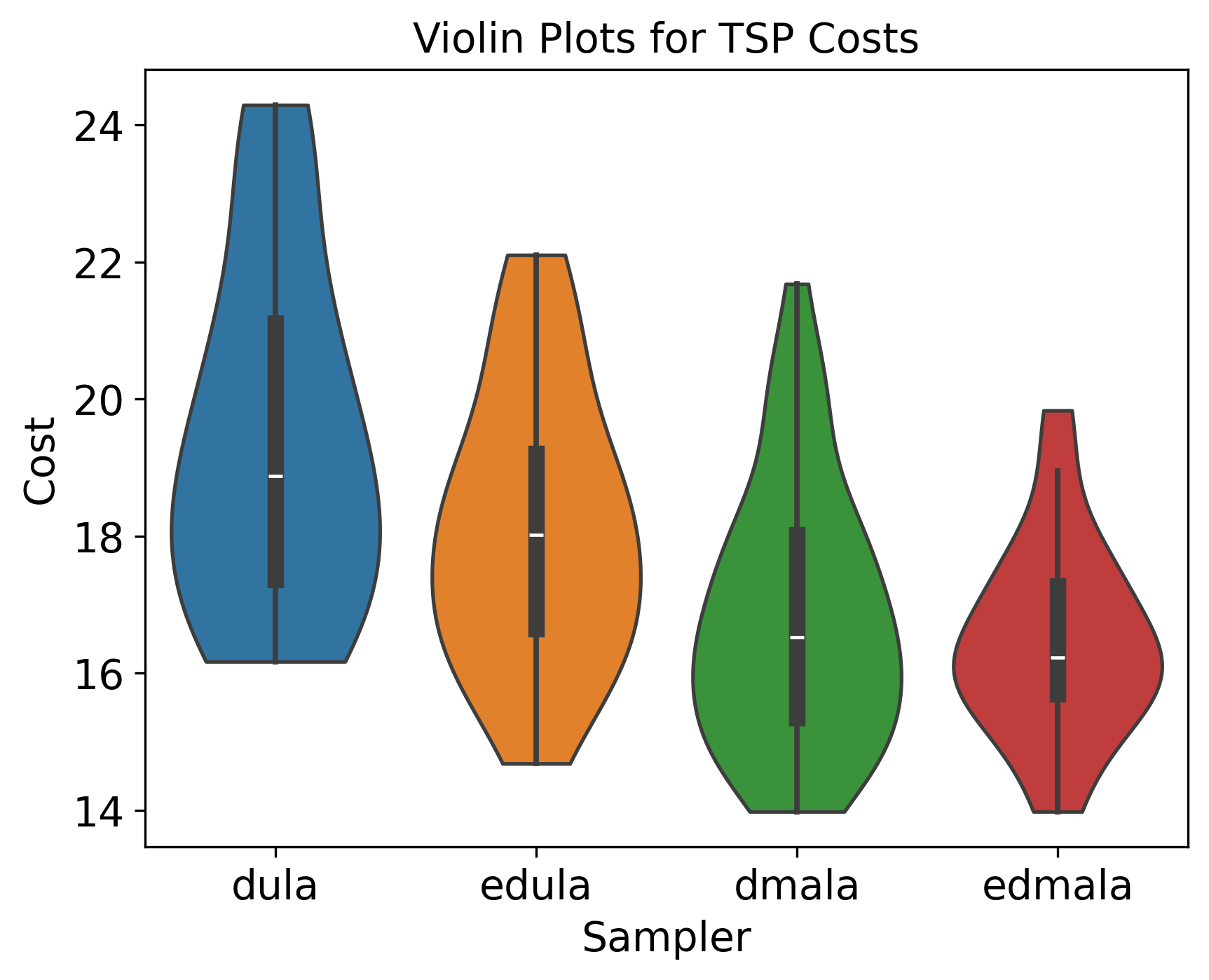} &
    \includegraphics[width=0.3\textwidth]{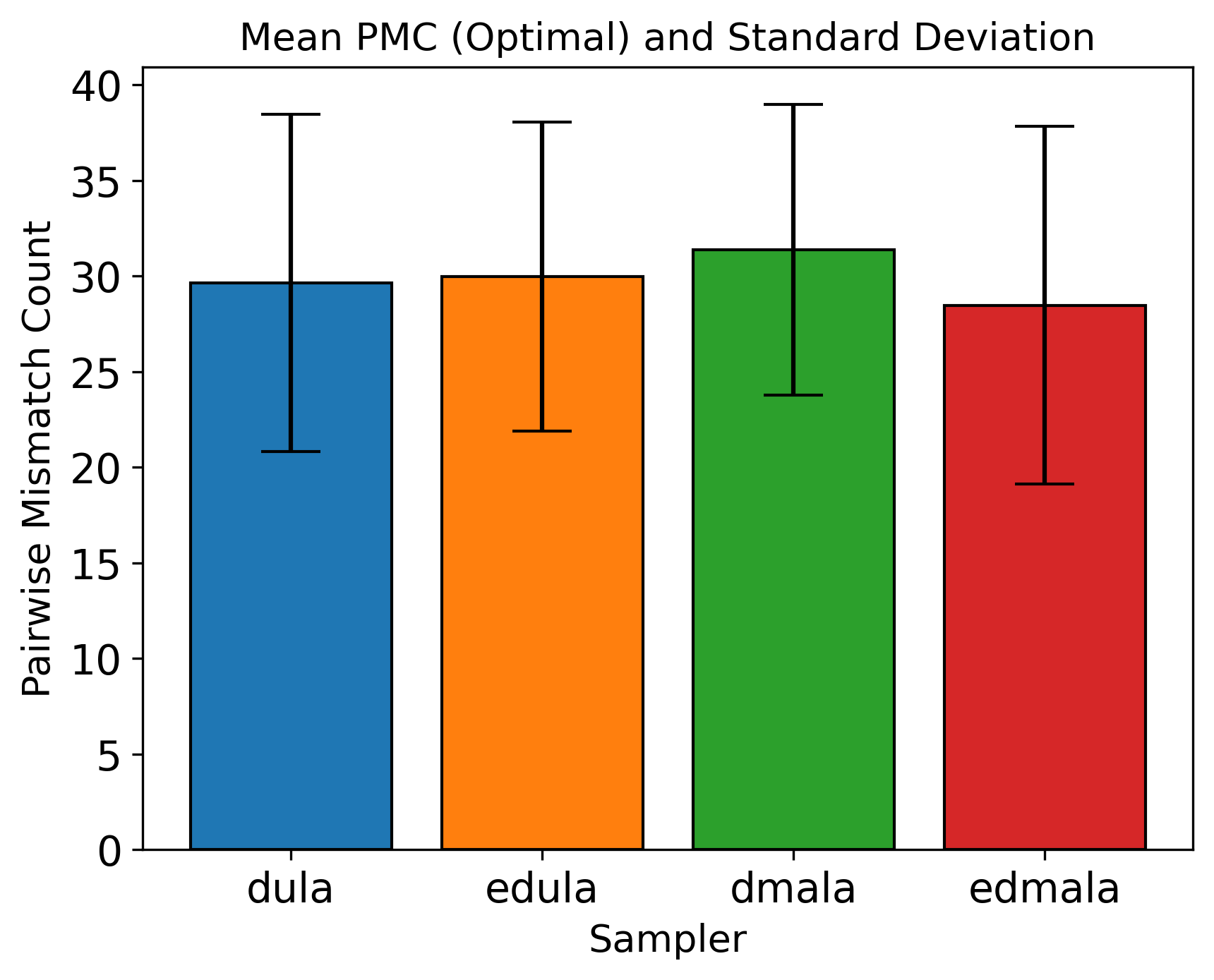}
  \end{tabular}
  \caption{Performance of various samplers on TSP.}
  \label{fig:tsp_cost}
  \vspace{-20pt}
\end{wrapfigure}

We employ four samplers: DULA, DMALA, EDULA, and EDMALA, each with a 10,000-iteration run and a 2,000-iteration burn-in period. After the burn-in, we record unique paths and plot their costs (negative of the energy function). Additionally, we identify the best path for each sampler amongst all unique solutions . Consequently, we calculate the average pairwise mismatch count (PMC) of the best path to all other sampled paths (see Figure \ref{fig:tsp_cost}), which quantifies how distinct the explored solutions are from the optimal path \citep{schiavinotto2007review, merz1997genetic}. 

\textbf{Left:}  EDULA and EDMALA, show clear superiority over their counterparts, DULA and DMALA, by achieving lower variance cost-spreads. This highlights the less variability in their sampling, demonstrating their superiority in efficiently finding consistent, robust solutions for the TSP.

\textbf{Right:}  To examine the potential variability from the optimal solution, we focus on the upper confidence band, represented as the mean discrepancy plus its standard deviation. While DULA and EDULA have similar upper bounds, EDMALA has a lower upper bound compared to DMALA.  
We provide additional results in the Appendix   \ref{subsec:app-tsp}.

\subsection{Sampling From Restricted Boltzmann Machines}
Restricted Boltzmann Machines (RBMs) are a class of generative stochastic neural networks that learn a probability distribution over their input data. 
The energy function for an RBM, which defines the joint configuration of visible and hidden units, is given by:
\[
U(\bm{\theta}) = \sum_i \text{Softplus}(\mathbf{W}\bm{\theta} + a)_i + b^{\top}\bm{\theta},
\]
where \(\{\mathbf{W}, a, b\}\) are the weight matrix and bias parameters, respectively, and \(\bm{\theta} \in \{0, 1\}^d\) represents the binary state of the visible units.

When the RBM assigns high probability to specific digit representations, a sharp mode for digit 3 (for instance) might appear as an idealized version without extraneous strokes. This configuration represents the model’s interpretation of a quintessential ‘3’ with a prominent probability peak. Any minor alteration, like flipping a single pixel, lowers the altered image’s probability. The sampler has thus learned to prioritize exact, pristine versions of each digit, marking any deviation from this high-probability state as unlikely.

For MNIST, this narrow focus limits flexibility. The model assigns high probability to only a few “perfect” digit versions, treating minor variations as less probable. This rigidity makes the generated images sensitive to small changes and limits the RBM’s ability to recognize natural, varied handwriting. In the context of RBMs, sampling from flat modes explores a wider range of latent handwritten styles, enhancing the model’s ability to capture the underlying data distribution. This reflects a broader representation of possible input variations, crucial for tasks like image generation and data reconstruction \cite{ mumurray2009evaluating}. In practice, this means that images generated from flat modes in RBMs are less likely to overfit to sharp, specific patterns in the training data and are instead more reflective of the variability inherent in the dataset.

In our experiments, we generated 5000 images per sampler for the MNIST dataset, applying a thinning factor of 1000 to ensure diversity in the samples. A simple convolutional autoencoder (CAE) was used for image generation and reconstruction, allowing us to evaluate the performance and generalization capability of sampler-generated data. To assess robustness, we trained 5 CAEs on the sampler-generated images and tested them under various conditions. Initially, clean test data was used to establish baseline performance. Subsequently, we introduced Gaussian noise (with a noise factor of 0.1) to evaluate the models’ resilience against perturbations, a common method for assessing adversarial robustness \citep{madry2018towards}. Additionally, we examined the models with occluded images, where random sections of the images were obscured by zero-valued pixel blocks. This test simulates scenarios with missing or obstructed information, a widely used technique in robustness studies to measure model performance under partial information loss \citep{zhang2019theoretically}.

For quantitative evaluation, we employed several widely accepted metrics: Mean Reconstruction Squared Error (MSE) to measure pixel-level differences between original and reconstructed images, Peak Signal Noise Ratio (PSNR) to measure the fidelity of the reconstructed images, and the Structural Similarity Index (SSIM) to assess the structural integrity of the reconstructions \citep{wang2004image}. Additionally, we computed the log-likelihood to quantify how well the reconstructed images fit the underlying data distribution. These metrics collectively offer a comprehensive assessment of the performance and robustness of the models across clean, noisy, and occluded data.

\begin{table}[t]
    \centering
    \renewcommand{\arraystretch}{0.85}
    \caption{Results of different samplers on MNIST under clean, noisy, and occluded conditions.}
      \resizebox{0.85\textwidth}{!}{%
    \begin{tabular}{|c|c|c|c|c|c|}
        \hline
        {\textbf{Sampler}} & {\textbf{Setting}} & {\textbf{MSE}(↓)} & {\textbf{PSNR}(↑)} & {\textbf{SSIM}(↑)} & {\textbf{Log-Likelihood}(↑)} \\ \hline
        HB-10-1 & Clean & 0.0253 ± 0.0005 & 16.3555 ± 0.0858 & 0.5303 ± 0.0014 & -0.0134 ± 0.0009 \\
               & Noisy & 0.0267 ± 0.0004 & 15.9763 ± 0.0697 & 0.3941 ± 0.0035 & 0.0165 ± 0.0011 \\
               & Occluded & 0.0256 ± 0.0004 & 16.2720 ± 0.0749 & 0.4963 ± 0.0017 & -0.0154 ± 0.0008 \\  \hline
        BG-1    & Clean & 0.0257 ± 0.0007 & 16.2492 ± 0.1125 & 0.5294 ± 0.0025 & -0.0157 ± 0.0014 \\
               & Noisy & 0.0270 ± 0.0006 & 15.9086 ± 0.0885 & 0.3938 ± 0.0038 & 0.0144 ± 0.0013 \\
               & Occluded & 0.0260 ± 0.0006 & 16.1613 ± 0.0992 & 0.4947 ± 0.0024 & -0.0179 ± 0.0013 \\ \hline
        DULA   & Clean & 0.0268 ± 0.0006 & 16.1160 ± 0.1022 & 0.5114 ± 0.0030 & -0.0209 ± 0.0015 \\
               & Noisy & 0.0280 ± 0.0005 & 15.7851 ± 0.0815 & 0.3907 ± 0.0041 & 0.0097 ± 0.0013 \\
               & Occluded & 0.0272 ± 0.0006 & 16.0187 ± 0.0922 & 0.4766 ± 0.0028 & -0.0233 ± 0.0014 \\  \hline
        DMALA  & Clean & 0.0256 ± 0.0004 & 16.3305 ± 0.0709 & 0.5291 ± 0.0035 & -0.0156 ± 0.0011 \\
               & Noisy & 0.0270 ± 0.0004 & 15.9547 ± 0.0623 & 0.3939 ± 0.0032 & 0.0148 ± 0.0009 \\
               & Occluded & 0.0259 ± 0.0004 & 16.2372 ± 0.0632 & 0.4950 ± 0.0035 & -0.0182 ± 0.0010 \\  \hline
        EDULA  & Clean & 0.0264 ± 0.0005 & 16.2135 ± 0.0877 & 0.5083 ± 0.0052 & -0.0179 ± 0.0014 \\
               & Noisy & 0.0276 ± 0.0004 & 15.8700 ± 0.0652 & \textbf{0.3968} ± 0.0030 & 0.0121 ± 0.0012 \\
               & Occluded & 0.0268 ± 0.0005 & 16.1115 ± 0.0797 & 0.4743 ± 0.0051 & -0.0206 ± 0.0014 \\  \hline
        EDMALA & Clean & \textbf{0.0251} ± 0.0005 & \textbf{16.3974} ± 0.0975 & \textbf{0.5368} ± 0.0016 & \textbf{-0.0117} ± 0.0009 \\
               & Noisy & \textbf{0.0266} ± 0.0004 & \textbf{15.9938} ± 0.0727 & 0.3933 ± 0.0029 & \textbf{0.0177} ± 0.0012 \\
               & Occluded & \textbf{0.0255} ± 0.0005 & \textbf{16.3022} ± 0.0839 & \textbf{0.5019} ± 0.0017 & \textbf{-0.0141} ± 0.0007 \\  \hline
    \end{tabular}%
    }
    \label{tab:rbm}
    \vspace{-8pt} 
\end{table}

The results in Table \ref{tab:rbm} indicate that EDLP methods consistently outperform their non-entropic counterparts across all test settings. Specifically, EDMALA achieves the lowest MSE, highest PSNR, highest SSIM (except for Noisy), and the best log-likelihood values among the samplers tested. These metrics together suggest that EDLP has superior generalization capabilities, making it especially effective for reconstructing unseen data accurately.
We provide additional results in the Appendix  \ref{subsec:app-rbm}.

\subsection{Binary Bayesian Neural Networks}

In alignment with the findings of \citeauthor{li2024entropymcmc} (Section 6.3), which highlight the role of flat modes in enhancing generalization in deep neural networks, we explore the training of binary Bayesian neural networks using discrete sampling techniques, leveraging the ability of flat modes to facilitate better generalization. Our experimental design involves regression tasks on four UCI datasets~\cite{Dua:2019}, with the energy function for each dataset defined as follows:
\[
U(\bm{\theta}) = -\sum_{i=1}^N ||f_{\bm{\theta}}(x_i) - y_i||^2,
\]
where $D = \{x_i, y_i\}_{i=1}^N$ is the training dataset, and $f_{\bm{\theta}}$ denotes a two-layer neural network with \texttt{Tanh} activation and 500 hidden neurons. Following the experimental setup in \cite{zhang2022langevin}, we report the average test RMSE and its standard deviation. As shown in Table~\ref{tab:bayesbnn}, EDMALA and EDULA consistently outperform their non-entropic variants across all datasets, but don't outperform GWG-1 on test RMSE on the COMPAS dataset. This exception can be attributed to overfitting, aligning with prior work~\cite{zhang2022langevin}. Overall, these results confirm that our method enhances generalization performance on unseen test data.
We provide additional results and hyperparameter settings in the Appendix \ref{subsec:app-bbnn}. 
\begin{table}[!t]
\vspace{-5pt} 
    \centering
    \renewcommand{\arraystretch}{0.85}
    \caption{Average Test RMSE for various datasets.}
      \resizebox{0.85\textwidth}{!}{%
    \begin{tabular}{|l|c|c|c|c|c|c|}
        \hline
        Dataset & \textbf{Gibbs} & \textbf{GWG} & \textbf{DULA} & \textbf{DMALA} & \textbf{EDULA} & \textbf{EDMALA} \\
        \hline
        COMPAS & \textbf{0.4752} \scriptsize{±0.0058} & 0.4756 \scriptsize{±0.0056} & 0.4789 \scriptsize{±0.0039} & 0.4773 \scriptsize{±0.0036} & 0.4778 \scriptsize{±0.0037} & 0.4768 \scriptsize{±0.0033} \\
        \hline
        News & 0.1008 \scriptsize{±0.0011} & 0.0996 \scriptsize{±0.0027} & 0.0923 \scriptsize{±0.0037} & 0.0916 \scriptsize{±0.0040} & 0.0918 \scriptsize{±0.0036} & \textbf{0.0915} \scriptsize{±0.0036} \\
        \hline
        Adult & 0.4784 \scriptsize{±0.0151} & 0.4432 \scriptsize{±0.0255} & 0.3895 \scriptsize{±0.0102} & 0.3872 \scriptsize{±0.0107} & 0.3889 \scriptsize{±0.0097} & \textbf{0.3861} \scriptsize{±0.0110} \\
        \hline
        Blog & 0.4442 \scriptsize{±0.0107} & 0.3728 \scriptsize{±0.0093} & 0.3236 \scriptsize{±0.0114} & 0.3213 \scriptsize{±0.0117} & 0.3218 \scriptsize{±0.0119} & \textbf{0.3211} \scriptsize{±0.0145} \\
        \hline
    \end{tabular}%
    }
    \label{tab:bayesbnn}
    \vspace{-8pt} 
\end{table}

\section{Discussion}
\subsection{Limitations}\label{sec:lim}
Since EDLP collects only discrete samples, it produces half as many samples per iteration as EMCMC. The coupling mechanism in Section \ref{sec:coupling} increases the computational load relative to DLP. However, as \citeauthor{li2024entropymcmc} states in their Section 4.2, the cost of gradient computation remains the same for \(d\)-dimensional models when \(\widetilde{\bm{\theta}}\) resides in a \(2d\) dimensional space. EDLP doubles memory usage compared to DLP, but the space complexity remains linear in \(d\), ensuring scalability.
\subsection{Conclusion}
We propose a simple and computationally efficient gradient-based sampler designed for sampling from flat modes in discrete spaces. The algorithm leverages a guiding variable based on local entropy. We provide non-asymptotic convergence guarantees for both the unadjusted and Metropolis-adjusted versions. Empirical results demonstrate the effectiveness of our method across a variety of applications. We hope our framework highlights the importance of flat-mode sampling in discrete systems, with broad utility across scientific and machine learning domains.
\newpage

\bibliography{main}
\bibliographystyle{icml2025}

\appendix
\onecolumn
\section{Analysis of the Effect of Flatness Parameter $\eta$ \label{sec:app_eta}}
\subsection{Intuition}
\begin{figure}[H]
    \centering
    \includegraphics[width=0.6\linewidth]{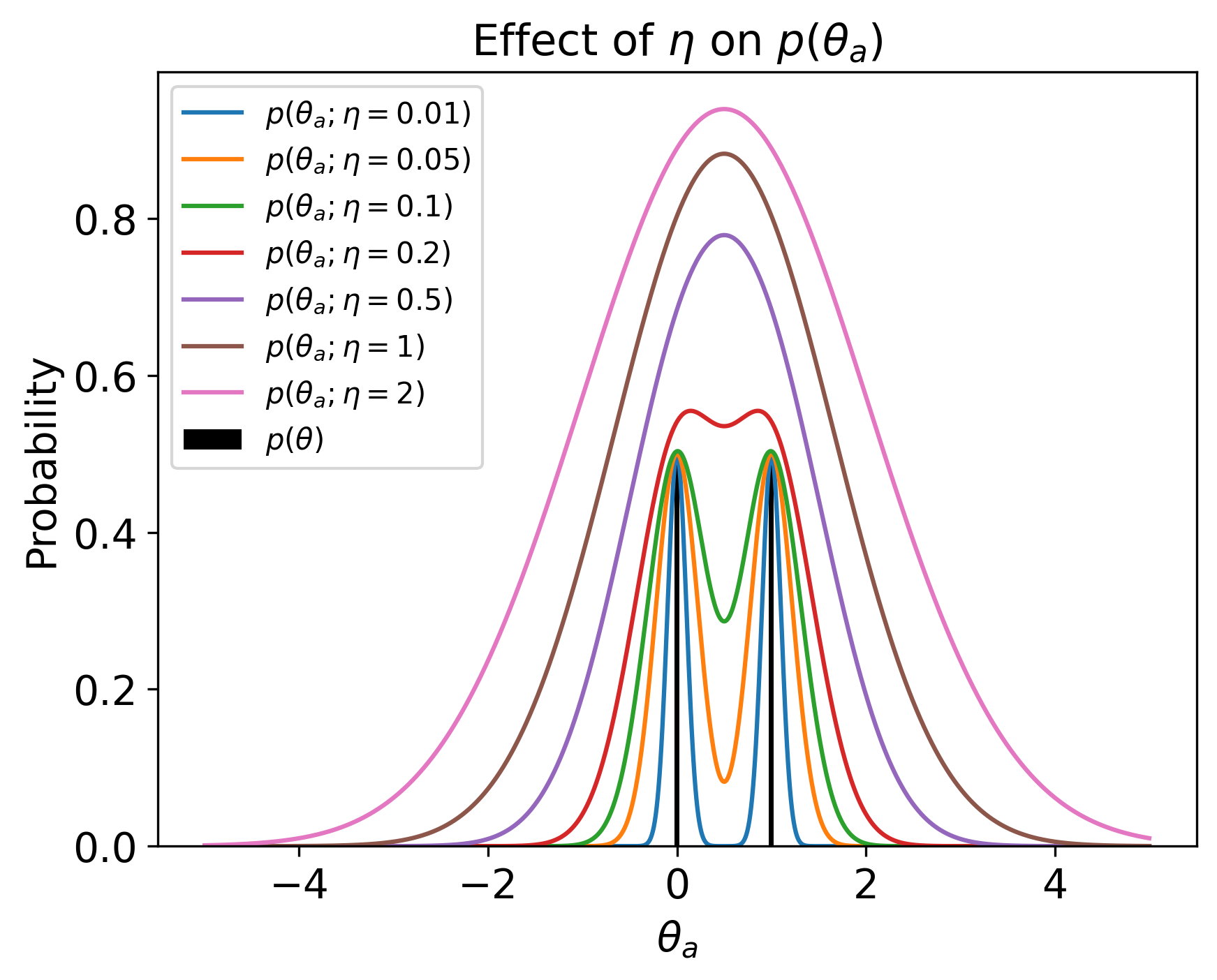}
    \caption{$p(\bm{\bm{\theta}_a})$ for $\bm{\theta} \sim \textit{Bernoulli}(0.5)$}
    \label{fig:eta_effect}
\end{figure}

Figure \ref{fig:eta_effect} illustrates the effect of varying the flatness parameter $\eta$ on the probability distribution $p(\bm{\theta}_a)$ for $\bm{\theta}$ drawn from a Bernoulli(0.5) distribution. The \textit{layered} curves represent different values of $\eta$, showing how the distribution $p(\bm{\theta}_a)$ changes as $\eta$ increases.

\subsubsection*{Effect of Small $\eta$ (Strong Coupling)}

For very small values of $\eta$ (e.g., $\eta = 0.01$, $\eta = 0.05$,  $\eta = 0.1$), the curves (blue, orange, and green) are sharply peaked and closely resemble the original $p(\bm{\theta})$.
Small $\eta$ values imply strong coupling between $\bm{\theta}$ and $\bm{\theta}_a$. The auxiliary distribution $p(\bm{\theta}_a)$ remains very close to $p(\bm{\theta})$, indicating that $\bm{\theta}_a$ is tightly bound to $\bm{\theta}$, and the variance is minimal.

\subsubsection*{Moderate $\eta$ Values (Moderate Coupling)}

As $\eta$ increases (e.g., $\eta = 0.2$), the curves (red) become wider and smoother.
These moderate $\eta$ values adequately capture the flatness of the landscape. The distribution $p(\bm{\theta}_a)$ starts to diverge from $p(\bm{\theta})$, allowing $\bm{\theta}_a$ to explore a broader region around the peaks.

\subsubsection*{Large $\eta$ (Weak Coupling)}

For larger values of $\eta$ (e.g., $\eta = 0.5$, $\eta = 1$, $\eta = 2$), the curves (purple, brown, and magenta) are much wider.
Large $\eta$ values imply weak coupling between $\bm{\theta}$ and $\bm{\theta}_a$. The auxiliary distribution $p(\bm{\theta}_a)$ is excessively smoothed out compared to $p(\bm{\theta})$, indicating that $\bm{\theta}_a$ can explore a much broader range of values with less influence from $\bm{\theta}$.

\subsubsection*{Considerations for \( \eta \) Approaching Infinity}
As \( \eta \) approaches infinity, the auxiliary distribution \( p(\bm{\theta}_a) \) flattens, and the gradient \( \nabla_{\bm{\theta}_a} U_{\eta}(\widetilde{\bm{\theta}}) \) tends toward zero. This results in an extremely weak coupling, effectively causing the EDLP framework to behave similarly to a standard DLP. The parameter \( \eta \) thus plays a critical role in determining the behavior of the sampler, necessitating careful tuning based on the specific requirements of the sampling task.

\subsection{Sensitivity Analysis}
\begin{figure}[t]
  \begin{center}
  \begin{tabular}{cc}
    \hspace{-10pt}
    \includegraphics[width=.45\textwidth]{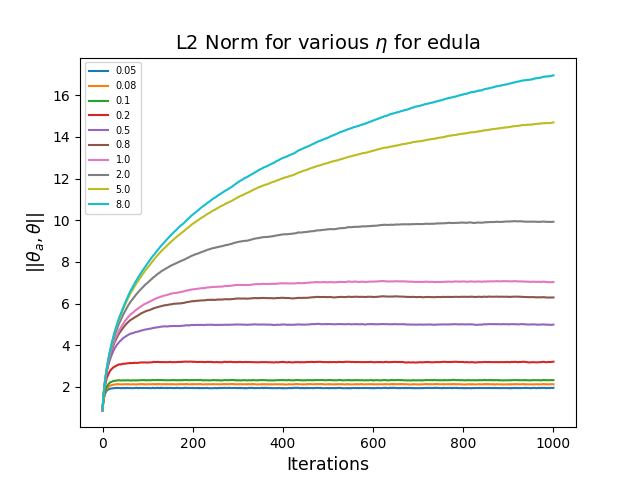} &
    \hspace{-10pt}
    \includegraphics[width=.45\textwidth]{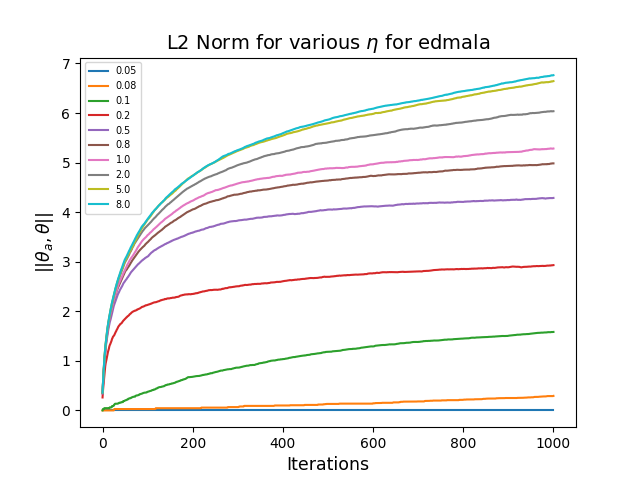} \\
    \hspace{-10pt}
    \includegraphics[width=.45\textwidth]{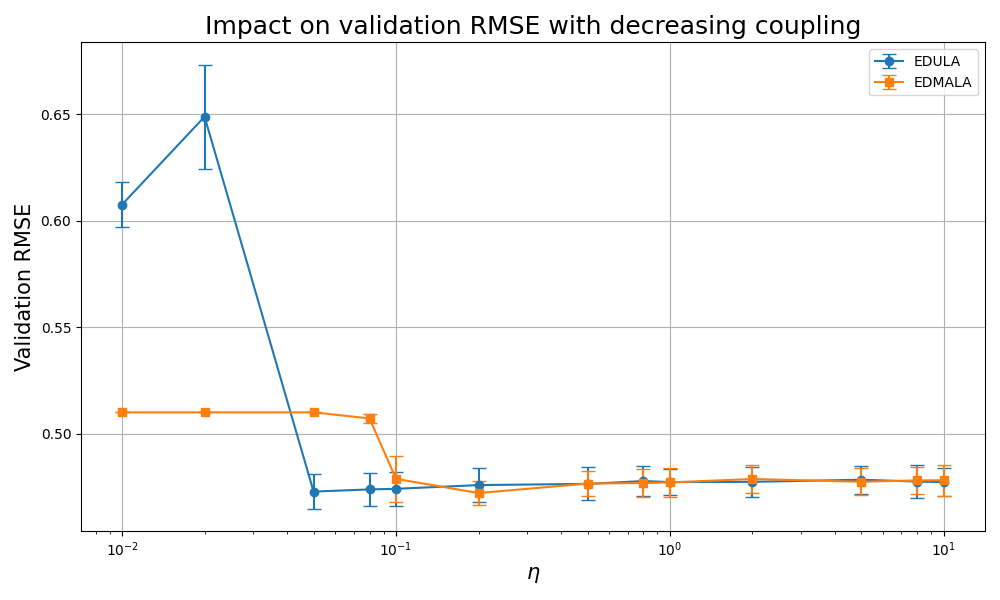} &
    \hspace{-10pt}
    \includegraphics[width=.45\textwidth]{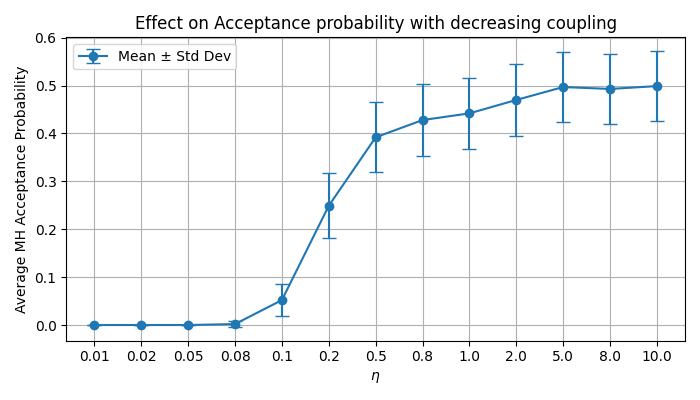}
  \end{tabular}
  \end{center}
  \vspace{-10pt}
  \caption{Diagnostics for EDLP}
  \label{fig:Diagnostics}
  \vspace{-10pt}
\end{figure}

The flatness parameter $\eta$ is arguably the most crucial hyperparameter to optimize in the EDLP algorithm (Algorithm \ref{alg:edlp}). Similar to the hyperparameter tuning ablation strategies employed in \cite{li2024entropymcmc} (Appendix, Section E), we conduct hyperparameter tuning on the COMPAS dataset’s validation data. Specifically, we monitor the L2 norm between sampled pairs of $\bm{\theta}$ and $\bm{\theta}_a$ for various values of $\eta$. Additionally, we plot the validation RMSE for both EDULA and EDMALA across different values of $\eta$. Finally, we plot the average MH acceptance ratio for EDMALA to assess the impact of $\eta$ on the joint MH acceptance step. We maintain $\alpha=0.1$ for both samplers and $\alpha_a=0.01$ for EDULA and $\alpha_a=0.001$ for EDMALA( see Figure \ref{fig:Diagnostics}).

We observe that as $\eta$ increases, the coupling between the variables weakens, allowing both variables to move more freely, thus increasing the norm. This behavior is consistent across both EDULA and EDMALA. However, EDMALA exhibits a more conservative behavior at the same coupling strength compared to EDULA due to the presence of the joint Metropolis-Hastings (MH) acceptance step, which imposes stricter alignment between the variables, hence maintaining a tighter coupling.

Both samplers demonstrate robustness across a wide range of $\eta$, with relatively stable validation RMSE performance. However, EDULA shows slightly less robustness, particularly at extremely small coupling values, resulting in increased variability and higher RMSE. EDMALA maintains a stable, consistent performance, indicating better robustness to changes in the coupling parameter.

The final plot shows how the MH acceptance probability varies with coupling strength $\eta$ for EDMALA. Initially, with very tight coupling , the acceptance probability is near zero, indicating overly restricted movements due to the strong alignment requirement between the discrete and continuous variables. As $\eta$ increases (coupling relaxes), the acceptance probability rises significantly, reflecting greater freedom in proposing moves that the joint MH criterion accepts. After a certain coupling threshold (around 0.8 here), the acceptance rate plateaus, suggesting diminishing returns from further relaxation in coupling strength. Thus, an intermediate coupling provides a balance, allowing effective exploration without overly compromising the sampler’s consistency.

\section{Gibbs-like Update Procedure \label{sec:glu}}
Gibbs-like updating procedures have been widely employed across various contexts in the sampling literature, particularly within Bayesian hierarchical models, latent variable models, and non-parametric Bayesian approaches. For instance, Gibbs sampling is a fundamental technique in hierarchical Bayesian models, where parameters are partitioned into blocks and updated conditionally on others to facilitate efficient sampling \citep{casella1992explaining}. In latent variable models, such as Hidden Markov Models (HMMs) and mixture models, Gibbs-like updates allow for alternating between sampling latent variables and model parameters, thereby simplifying the overall process \citep{diebolt1994estimation}. Additionally, these updates are crucial in non-parametric Bayesian approaches, such as Dirichlet Process Mixture Models (DPMMs), where they enable the efficient sampling of cluster assignments and hyperparameters \citep{neal2000markov}. Gibbs-like updates are also prominently used in spatial statistics, particularly in Conditional Autoregressive (CAR) models, where the value at each spatial location is updated based on its neighbors \citep{besag1974spatial}.

Since our goal is to sample from a joint distribution, rather than simultaneously updating \(\bm{\theta}\) and \(\bm{\theta}_a\), we alternatively update these variables iteratively. The conditional distribution for the primary variable \(\bm{\theta}\) is given by:

\[
p(\bm{\theta}|\bm{\theta}_a) \propto \frac{1}{Z_{\bm{\theta}_a}} \exp{\left\{ U(\bm{\theta}) - \frac{1}{2\eta} \|\bm{\theta} - \bm{\theta}_a\|^2 \right\}},
\]

where \(Z_{\bm{\theta}_a} = \exp{\mathcal{F}(\bm{\theta}_a; \eta)}\) serves as the normalization constant. Correspondingly, the conditional distribution for the auxiliary variable \(\bm{\theta}_a\) is:

\[
p(\bm{\theta}_a|\bm{\theta}) \propto \frac{1}{Z_{\bm{\theta}}} \exp{\left\{ -\frac{1}{2\eta} \|\bm{\theta} - \bm{\theta}_a\|^2 \right\}},
\]

where \(Z_{\bm{\theta}} = \exp{(U(\bm{\theta}))}\) is the associated normalization constant. This formulation reveals that \(\bm{\theta}_a\) is sampled from \(\mathcal{N}(\bm{\theta}, \eta\bm{I})\), with the variance \(\eta\) controlling the expected distance between \(\bm{\theta}\) and \(\bm{\theta}_a\). During the Metropolis-Hastings (MH) step, the acceptance probability is now calculated as:

\[
\min\left(1,  \frac{q_{\alpha}({\bm{\theta}}|\widetilde{\bm{\theta'}})}{q_{\alpha}({\bm{\theta'}}|\widetilde{\bm{\theta}})}\frac{\pi(\widetilde{\bm{\theta'}})}{\pi(\widetilde{\bm{\theta}})}\right).
\]

This Gibbs-like alternating update scheme offers distinct advantages: (1) exact sampling of \(\bm{\theta}_a\), (2) elimination of the need for the \(\alpha_a\) parameter, (3) a less intensive computation of the MH acceptance probability, and (4) reduced overall computational overhead, especially when the proposal step involves an MH correction. This gibbs-like updating also shares similarities with the proximal sampling methods \citep{pereyra2016proximal, liang2023a}. This innovation can potentially allow DLP to generalize effectively to more complex, high-dimensional, and non-differentiable discrete target distributions such as the discrete Laplace distribution, which is commonly used in privacy-preserving mechanisms\citep{dwork2006calibrating, ghosh2012universally}. We leave out the theoretical analysis of the GLU versions for future work.

\section{Considerations for Excluding Focussed Belief Propogation from Benchmarking \label{sec:app_fbp}}

 \textbf{1. Fundamental Differences in Sampling Mechanism: } Most of the sampling algorithms we use generate samples sequentially, with each sample \( x_{t+1} \) derived from the previous sample \( x_t \). This sequential dependency is essential for building a Markov Chain that explores the distribution space and gradually converges to the target distribution. fBP produces samples sequentially, but instead employs a \textit{message-passing algorithm} aimed at converging to a fixed solution or configuration. It operates to converge deterministically to a solution, rather than generating a sequence of probabilistic samples. Moreover, fBP lacks a formal proof of convergence, relying instead on heuristic principles rooted in replica theory. This absence of theoretical guarantees or established convergence rates means that even if fBP appears to perform well, we cannot interpret or quantify its reliability, efficiency, or consistency across varying datasets and tasks. In contrast, MCMC-based methods like Langevin dynamics and Gibbs sampling come with well-understood convergence properties, enabling meaningful performance evaluations and robust benchmarking. This interpretability gap makes fBP less suitable for our study, where theoretical soundness and predictable behavior are critical.
    
\textbf{2. Technical and Practical Constraints with using fBP: } While fBP is originally implemented in Julia\footnote{Carlo Baldassi, \textit{BinaryCommitteeMachinefBP.jl}, GitHub repository, \url{https://github.com/carlobaldassi/BinaryCommitteeMachinefBP.jl}, accessed November 8, 2024.}, a Python wrapper\footnote{Curti, Nico and Dall’Olio, Daniele and Giampieri, Enrico, \textit{ReplicatedFocusingBeliefPropagation}, GitHub repository, \url{https://github.com/Nico-Curti/rFBP}, accessed November 8, 2024.} is also available. However, this wrapper still depends on the underlying Julia or C++ implementations, introducing potential cross-language communication overhead. This dependency complicates integration in Python workflows and creates an inherent performance disparity when compared to purely Pythonic implementations, making direct runtime comparisons less meaningful. Despite fBP’s speed advantage, its execution becomes slow as sample dimensions increase and network ensembles grow larger. The volume of message-passing in high-dimensional contexts limits its scalability. As task complexity increases, fBP faces challenges in achieving stable convergence, further limiting its suitability for our high-dimensional setup. Past studies have excluded computationally expensive methods from experimental evaluations \cite{zhang2022langevin}.

\textbf{3. Computational Overhead and Efficiency Concerns Resource Demands for Multiple Runs:} 
If we were to use fBP to generate multiple samples, we would need to reinitialize and re-run the algorithm for each sample with a new seed, effectively solving the problem from scratch each time. This is highly inefficient compared to MCMC methods, where each subsequent sample builds on the previous one without needing to restart the entire algorithm. For larger models and datasets, this repeated initialization and execution would result in a significant computational burden.

\textbf{4. Nature of Tasks:} 
In certain structured sampling tasks, such as the TSP, we enforce constraints to ensure that each proposed state is a valid TSP solution. This entails accepting only those configurations that satisfy specific requirements of the TSP. However, fBP  does not adhere to such constraints, as it lacks mechanisms for directly enforcing the validity of the sampled states. Consequently, fBP is unsuitable for tasks where such structural constraints are critical, placing it outside the scope for comparison in these applications.

\begin{figure}[H]
\centering
    \begin{tabular}{ccc}       
        \includegraphics[width=0.2\textwidth]{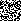}  &
        \includegraphics[width=0.2\textwidth]{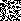}  &
        \includegraphics[width=0.2\textwidth]{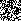}
        \\
        (a)  &
        (b) &
        (c)
    \end{tabular}
    \caption{Random Image Samples for MNIST using fBP}
    \label{fig:fbp_mnist}
\end{figure}

\begin{figure}[H]
\centering
    \begin{tabular}{ccc}       
        \includegraphics[width=0.2\textwidth]{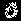}  &
        \includegraphics[width=0.2\textwidth]{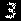}  &
        \includegraphics[width=0.2\textwidth]{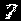}
        \\
        (a)  &
        (b) &
        (c)
    \end{tabular}
    \caption{Random Image Samples for MNIST using DMALA}
    \label{fig:DMALA_mnist}
\end{figure}

\begin{figure}[H]
\centering
    \begin{tabular}{ccc}       
        \includegraphics[width=0.2\textwidth]{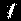}  &
        \includegraphics[width=0.2\textwidth]{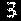}  &
        \includegraphics[width=0.2\textwidth]{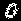}
        \\
        (a)  &
        (b) &
        (c)
    \end{tabular}
    \caption{Random Image Samples for MNIST using EDMALA}
    \label{fig:EDMALA_mnist}
\end{figure}

We conducted preliminary experiments using fBP for Restricted Boltzmann Machine (RBM) sampling on the MNIST dataset to assess its effectiveness in image generation. Figure \ref{fig:fbp_mnist} shows random image samples generated by fBP on MNIST, which resemble random unstructured noise rather than recognizable digits, compared to MNIST samples by DMALA and EDMALA in Figures \ref{fig:DMALA_mnist},  \ref{fig:EDMALA_mnist} respectively. These outputs suggest that fBP doesn’t capture the underlying structure of the MNIST data.

\begin{figure}[H]
    \centering
    \includegraphics[width=0.5\textwidth]{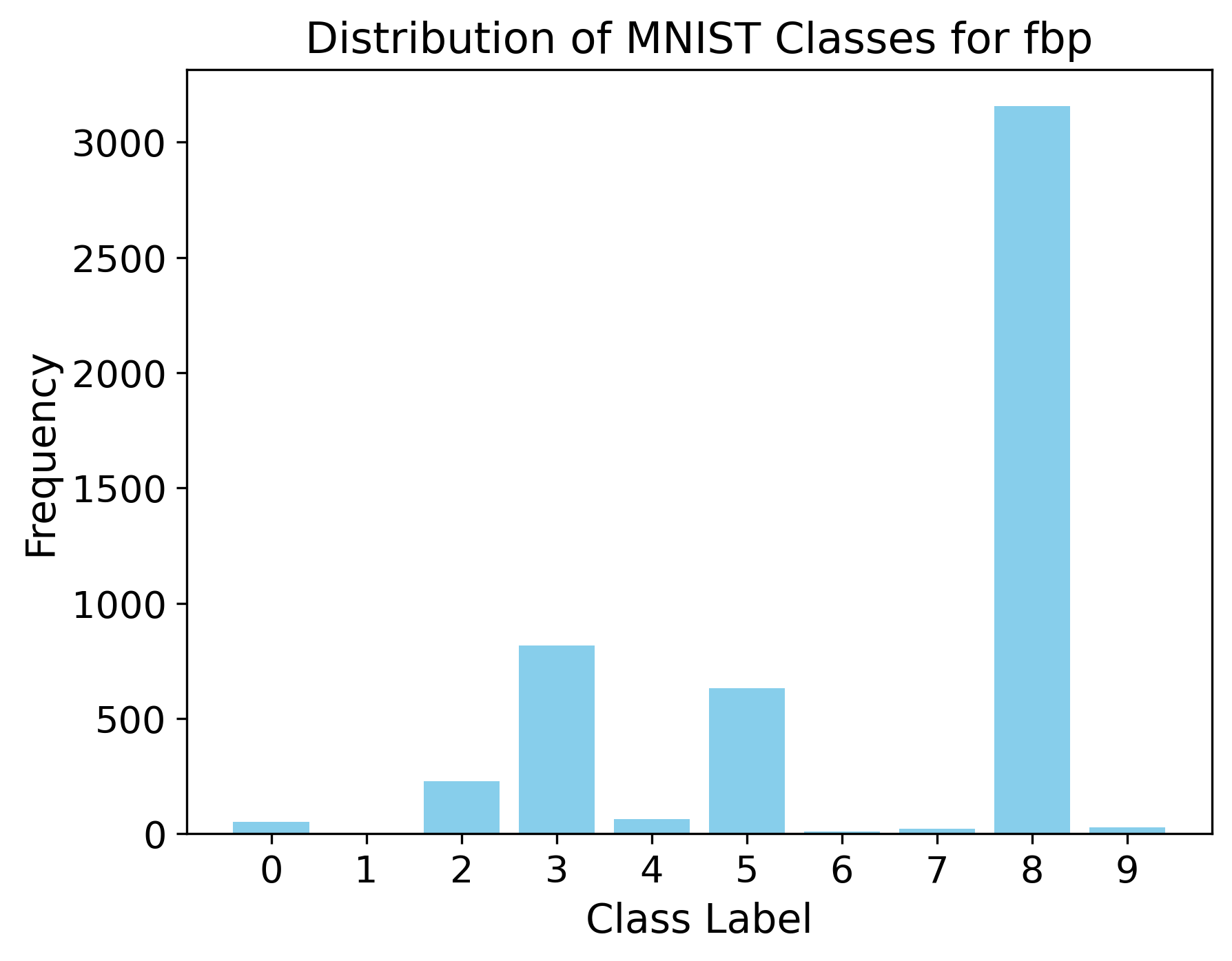}
    \caption{Mode Collapse using fBP}
    \label{fig:fbp_mode_analysis}
\end{figure}

fBP lacks direct use of the energy function $U(.)$ during optimization, preventing accurate data modeling. Figure \ref{fig:fbp_mode_analysis} illustrates this through a distribution analysis of generated MNIST classes, showing significant mode collapse. Most generated samples cluster around a few classes, with an imbalance favoring certain digits and ignoring others.

These findings highlight a fundamental issue with fBP in image generation tasks. Mode collapse suggests fBP struggles to explore diverse data regions, making it unsuitable for generating realistic, structured outputs that adhere to specific distribution characteristics, like image data in the MNIST dataset. 

In summary, fBP diverges significantly from the MCMC-based sampling methods used in our study due to its deterministic message-passing mechanism, which converges to fixed configurations rather than generating sequential probabilistic samples. While a Python wrapper exists, its reliance on the underlying Julia or C++ implementations introduces potential cross-language communication overhead, creating performance inconsistencies when compared to native Python implementations. Moreover, fBP’s lack of constraint adherence and dependence on spin-like variable encoding make it unsuitable for complex, structured sampling tasks like TSP or data-driven applications requiring diverse sampling, such as image generation on MNIST. Our preliminary experiments confirm that fBP struggles with mode collapse and fails to capture essential data distribution characteristics.

\section{Proofs \label{sec:app_proof}}
\subsection{Proof of Lemma~\ref{lemma:joint_posterior}}
Assume $\widetilde{\bm{\theta}}=[\bm{\theta}^T,\bm{\theta}_a^T]^T$ is sampled from the joint posterior distribution:
\begin{equation}
    p(\widetilde{\bm{\theta}})=p(\bm{\theta},\bm{\theta}_a)\propto\exp\left\{U(\bm{\theta})-\frac{1}{2\eta}\|\bm{\theta}-\bm{\theta}_a\|^2\right\}.
\end{equation}
Then the marginal distribution for $\bm{\theta}$ is:
\begin{equation}
\begin{aligned}
    p(\bm{\theta})&=\int_{}p(\bm{\theta},\bm{\theta}_a)d\bm{\theta}_a \\
    &=(2\pi\eta)^{-\frac{d}{2}}Z^{-1}\int_{}\exp\left\{U(\bm{\theta})-\frac{1}{2\eta}\|\bm{\theta}-\bm{\theta}_a\|^2\right\}d\bm{\theta}_a \\
    &=Z^{-1}\exp(U(\bm{\theta}))(2\pi\eta)^{-\frac{d}{2}}\int_{}\exp\left\{-\frac{1}{2\eta}\|\bm{\theta}-\bm{\theta}_a\|^2\right\}d\bm{\theta}_a \\
    &=Z^{-1}\exp(U(\bm{\theta})),
\end{aligned}
\end{equation}
where $Z=\sum_{\bm{\Theta}}\exp(U(\bm{\theta}))$ is the normalizing constant, and it is obtained by:
\begin{equation}
    \sum_{\bm{\Theta}}\int_{}\exp\left\{U(\bm{\theta})-\frac{1}{2\eta}\|\bm{\theta}-\bm{\theta}_a\|^2\right\}d\bm{\theta}_a=(2\pi\eta)^{\frac{d}{2}}\sum_{\bm{\Theta}}\exp(U(\bm{\theta})):=(2\pi\eta)^{\frac{d}{2}}Z.
\end{equation}
This verifies that the joint posterior distribution $p(\bm{\theta},\bm{\theta}_a)$ is mathematically well-defined\footnote{The exact form of the joint posterior is $p(\bm{\theta},\bm{\theta}_a)=(2\pi\eta)^{-\frac{d}{2}}Z^{-1}\exp(U(\bm{\theta})-\frac{1}{2\eta}\|\bm{\theta}-\bm{\theta}_a\|^2)$.}. Similarly, the marginal distribution for $\bm{\theta}_a$ is:
\begin{equation}
\begin{aligned}
p(\bm{\theta}_a)&=\sum_{\bm{\Theta}}p(\bm{\theta},\bm{\theta}_a) \\
    &\propto\sum_{\bm{\Theta}}\exp\left\{U(\bm{\theta})-\frac{1}{2\eta}\|\bm{\theta}-\bm{\theta}_a\|^2\right\} \\
    &=\exp{\mathcal{F}(\bm{\theta}_a;\eta)}.
\end{aligned}
\end{equation}

\subsection{Proof of Proposition \ref{prop:1}}
We follow a similar-style analysis as seen in Theorem 5.1 of \cite{zhang2022langevin}.

Using   Equation \eqref{eq:proposal},
\[
 \scalebox{0.8}{
$\begin{aligned}
q_{\gamma}(\widetilde{\bm{\theta}}'|\widetilde{\bm{\theta}}) & \propto \exp\left(\frac{1}{2} \nabla_{\bm{\theta}}U_{\eta}(\widetilde{\bm{\theta}})^{\top}(\bm{\theta}' - \bm{\theta}) - \frac{1}{2\alpha}\|\bm{\theta}' - \bm{\theta\|}^2\right) \cdot \frac{1}{\sqrt{2\pi\alpha_a}^d} \exp\left( -\frac{1}{2\alpha_a}\|\bm{\theta_a}' - \bm{\theta_a} - \frac{\alpha_a}{2}\nabla_{\bm{\theta_a}}U_{\eta}(\widetilde{\bm{\theta}})\|^2\right)\\
&= \frac{1}{\sqrt{2\pi\alpha_a}^d} \exp\left(\frac{1}{2} \nabla_{\bm{\theta}}U_{}(\bm{\theta})^{\top}(\bm{\theta}' - \bm{\theta}) - \frac{1}{2\alpha}\|\bm{\theta}' - \bm{\theta}\|^2   -\frac{1}{2\eta}(\bm{\theta} - \bm{\theta}_a)^{\top}(\bm{\theta}' - \bm{\theta})\right)\cdot\\
&\quad \left(  - \frac{1}{2\alpha_a} \|\bm{\theta}_a' - \bm{\theta}_a\|^2 + \frac{1}{2\eta} (\bm{\theta} - \bm{\theta}_a)^{\top} (\bm{\theta}'_a - \bm{\theta}_a) - \frac{\alpha_a}{8\eta^2} \|\bm{\theta} - \bm{\theta}_a\|^2 \right)\\
&= \frac{1}{\sqrt{(2\pi\alpha_a)^d}} \exp \left( \frac{1}{2}( -U(\bm{\theta}) + U(\bm{\theta}') ) - (\bm{\theta}-\bm{\theta}')^{\top} \left(\frac{1}{2\alpha}I + \frac{1}{4}\int_{0}^{1} \nabla^2 U((1-s)\bm{\theta} + s\bm{\theta}') ds \right)(\bm{\theta}-\bm{\theta}') \right. \\
&\quad \left. - \frac{1}{2\eta} (\bm{\theta} - \bm{\theta}_a)^{\top}(\bm{\theta}' - \bm{\theta} + \bm{\theta}_a - \bm{\theta}_a') - \frac{1}{2\alpha_a} \|\bm{\theta}_a' - \bm{\theta}_a\|^2 - \frac{\alpha_a}{8\eta^2} \|\bm{\theta} - \bm{\theta}_a\|^2 \right)\\
&= \frac{1}{\sqrt{(2\pi\alpha_a)^d}} \exp \left( \frac{1}{2}( -U(\bm{\theta}) + U(\bm{\theta}') ) - (\bm{\theta}-\bm{\theta}')^{\top}\left(\frac{1}{2\alpha}I + \frac{1}{4}\int_{0}^{1} \nabla^2 U((1-s)\bm{\theta}+s\bm{\theta}')ds \right)(\bm{\theta}-\bm{\theta}') \right. \\
&\quad \left. - \frac{1}{2\eta}(\bm{\theta} - \bm{\theta}_a)^{\top}(\bm{\theta}' - \bm{\theta}_a') - \frac{1}{2\alpha_a} \|\bm{\theta}_a' - \bm{\theta}_a\|^2 + \frac{4\eta-\alpha_a}{8\eta^2} \|\bm{\theta} - \bm{\theta}_a\|^2 \right)
\end{aligned}$}
\]

The normalizing constant  for  Equation \eqref{eq:proposal} \({Z}_{\bm{\widetilde{\Theta}}}(\widetilde{\bm{\theta}})\) is computed by integrating over \(\mathbb{R}^d\) and summing over \(\bm{\Theta}\):

{\scriptsize
\begin{align}{\label{eq:normalizing_constant}}
{Z_{\bm{\widetilde{\Theta}}}}(\widetilde{\bm{\theta}}) = \frac{1}{\sqrt{2\pi\alpha_a}^d} \int_{\bm{\theta'}_a} \sum_{\bm{\theta'} \in \Theta} \exp\bigg(\frac{1}{2} \nabla_{\bm{\theta}} U_{\eta}(\widetilde{\bm{\theta}})^{\top}(\bm{\theta'}-\bm{\theta}) -\frac{1}{2\alpha}{\norm{\bm{\theta'}-\bm{\theta}}^2}- \frac{1}{2\alpha_a} \| \bm{\theta'}_a - \bm{\theta}_a - \frac{\alpha_a}{2} \nabla_{\bm{\theta}_a} U_{\eta}(\widetilde{\bm{\theta}}) \|^2 \bigg)d\bm{\theta'}_a
\end{align}
}

We note that since $\nabla^2 U(\cdot)$ is continuous( from Assumption \ref{assumption:5.2}), we know that
\[\min_{x,y \in \bm{\Theta}} (x-y)^{T} \left(\int_{0}^{1} \nabla^2 U ((1-s)x+sy) ds\right) (x-y)\]
is well-defined.

Consequently, the modified normalizing constant(Equation \eqref{eq:normalizing_constant}), $Z_\gamma(\widetilde{\bm{\theta}})$, becomes
\[
 \scalebox{0.8}{
$\begin{aligned}
Z_\gamma(\widetilde{\bm{\theta}}) &= \frac{1}{\sqrt{(2\pi\alpha_a)^d}}
\int_{\bm{\theta'}_a} \sum_{\bm{\theta'} \in \bm{\Theta}} 
\exp \Bigg(
    \frac{1}{2} \big( -U(\bm{\theta}) + U(\bm{\theta}') \big) 
    - (\bm{\theta}-\bm{\theta}')^{\top} 
    \left( \frac{1}{2\alpha}I + \frac{1}{4} \int_{0}^{1} 
    \nabla^2 U \big( (1-s)\bm{\theta} + s\bm{\theta}' \big) \, ds \right) (\bm{\theta}-\bm{\theta}') \\
& \quad - \frac{1}{2\eta}(\bm{\theta} - \bm{\theta}_a)^{\top}(\bm{\theta}' - \bm{\theta}_a') 
    - \frac{1}{2\alpha_a} \|\bm{\theta}_a' - \bm{\theta}_a\|^2 
    + \frac{4\eta-\alpha_a}{8\eta^2} \|\bm{\theta} - \bm{\theta}_a\|^2
\Bigg).
\end{aligned}$}
\]

Now, we establish that $q(\widetilde{\bm{\theta}}|\widetilde{\bm{\theta}}')$ is reversible with respect to $\pi_{\gamma}$, where 

$\pi_{\gamma}=\frac{Z_\gamma(\widetilde{\bm{\theta}})\exp \{\frac{\alpha_a}{8\eta^2}\norm{\bm{\theta}-\bm{\theta}_a}^2\} \pi(\widetilde{\bm{\theta}})}{\int_{y} \sum_{x \in \bm{\bm{\Theta}}} Z_\gamma([x^{\top},y^{\top}]^{\top})\exp{\frac{\alpha_a}{8\eta^2}\norm{x-y}^2}  \pi([x^{\top},y^{\top}]^{\top}) dy}$.

Note that,
\[
 \scalebox{0.9}{
$\begin{aligned}
    \pi_{\gamma}(\widetilde{\bm{\theta}}) q_{\gamma}(\widetilde{\bm{\theta}}'|\widetilde{\bm{\theta}}) &= \frac{Z_\gamma(\widetilde{\bm{\theta}}) \exp \left( \frac{\alpha_a}{8\eta^2} \| \bm{\theta}-\bm{\theta}_a \|^2 \right) \pi(\widetilde{\bm{\theta}})}{\int_{y} \sum_{x \in \bm{\Theta}} Z_\gamma([x^{\top}, y^{\top}]^{\top}) \exp\left(\frac{\alpha_a}{8\eta^2} \|x - y\|^2\right) \pi([x^{\top}, y^{\top}]^{\top}) \, dy} \frac{1}{ Z_{\gamma}(\widetilde{\bm{\theta}})} \frac{1}{(\sqrt{2\pi\alpha_a})^d} \\
    &\quad \exp \Bigg( \frac{1}{2} \big( -U(\bm{\theta}) + U(\bm{\theta}') \big) - (\bm{\theta} - \bm{\theta}')^{\top} \left(\frac{1}{2\alpha}I + \frac{1}{4} \int_{0}^{1} \nabla^2 U((1-s)\bm{\theta} + s\bm{\theta}') \, ds \right) (\bm{\theta} - \bm{\theta}') \\
    &\quad - \frac{1}{2\eta} (\bm{\theta} - \bm{\theta}_a)^{\top} (\bm{\theta}' - \bm{\theta_a'}) - \frac{1}{2\alpha_a} \|\bm{\theta}_a' - \bm{\theta}_a\|^2 + \frac{4\eta - \alpha_a}{8\eta^2} \| \bm{\theta}-\bm{\theta}_a \|^2 \Bigg) \\
    &= \frac{1}{\int_{y} \sum_{x \in \bm{\Theta}} Z_{\gamma}([x^{\top}, y^{\top}]^{\top}) \exp\left(\frac{\alpha_a}{8\eta^2} \|x - y\|^2\right) \pi([x^{\top}, y^{\top}]^{\top}) \, dy} \frac{1}{(\sqrt{2\pi\alpha_a})^d} \\
    &\quad \exp \Bigg( \frac{1}{2} \big( U(\bm{\theta}) + U(\bm{\theta}') \big) - \frac{1}{2} (\bm{\theta} - \bm{\theta}')^{\top} \left(\frac{1}{\alpha}I + \frac{1}{2} \int_{0}^{1} \nabla^2 U((1-s)\bm{\theta} + s\bm{\theta}') \, ds \right) (\bm{\theta} - \bm{\theta}') \\
    &\quad - \frac{1}{2\eta} (\bm{\theta} - \bm{\theta}_a)^{\top} (\bm{\theta}' - \bm{\theta}_a') - \frac{1}{2\alpha_a} \|\bm{\theta}_a' - \bm{\theta}_a\|^2 \Bigg) \\
    &= \pi_{\gamma}(\bm{\theta}') q_{\gamma}(\bm{\theta}|\bm{\theta}').
\end{aligned}$}
\]
{Chain looks symmetric and reversible with respect to $\pi_\gamma$}.

 Now, given this, note that $Z'_{\gamma}(\widetilde{\bm{\theta}})$ converges to $1$ as $\alpha \to 0$ and $\alpha_a \to 0$.
 \[
 \scalebox{0.8}{
$\begin{aligned}
Z'_{\gamma}(\widetilde{\bm{\theta}}) &= Z_{\gamma}(\widetilde{\bm{\theta}}) \exp\left(\frac{\alpha_a}{8\eta^2} \| \bm{\theta}-\bm{\theta}_a \|^2\right) \\
&= \frac{1}{\sqrt{(2\pi\alpha_a)^d}} \int_{y} \sum_{x} \exp \Bigg( 
- \frac{1}{2} \left( U(\bm{\theta}) - U(x) \right) - (\bm{\theta} - x)^{\top}\left(\frac{1}{2\alpha} I + \frac{1}{4} \int_{0}^{1} \nabla^2 U((1-s)\bm{\theta} + s\bm{\theta}') \, ds \right)(\bm{\theta} - x) \\
&\quad - \frac{1}{2\alpha_a} \| y - \bm{\theta}_a \|^2 + \frac{4\eta}{8\eta^2} \| \bm{\theta} - \bm{\theta}_a \|^2 
\Bigg) \, dy \\
&\underset{\alpha \to 0}{=} \frac{1}{\sqrt{(2\pi\alpha_a)^d}} \int_{y} \sum_x \exp\left(\frac{1}{2} \left(U(x) - U(\bm{\theta})\right) - \frac{1}{2\alpha_a} \|y - \bm{\theta}_a\|^2 \right. + \frac{1}{2\eta}\| \bm{\theta} - \bm{\theta}_a\|^2 - \frac{1}{2\eta} (\bm{\theta} - \bm{\theta}_a)^{\top} (x - y) \Bigg) \delta_{\bm{\theta}}(x) \, dy \\
&= \int_{y} \exp\left(\frac{1}{2\eta} \| \bm{\theta}-\bm{\theta}_a \|^2 - \frac{1}{2\eta} (\bm{\theta} - \bm{\theta}_a)^{\top} (\bm{\theta} - y)\right) \, dy \\
&\underset{\alpha_a \to 0}{=} \int_{y} \exp\left(\frac{1}{2\eta} \| \bm{\theta}-\bm{\theta}_a \|^2 - \frac{1}{2\eta} (\bm{\theta} - \bm{\theta}_a)^{\top} (\bm{\theta} - \bm{\theta}_a)\right) \, dy \\
&= 1.
\end{aligned}$}
\]

where $\delta_{\theta}(.)$ is a Dirac delta. It follows that $\pi_{\gamma}$ converges pointwisely to $\pi(\widetilde{\bm{\theta}})$. By Scheffé’s Lemma,
it immediately implies $\pi_{\gamma}(\widetilde{\bm{\theta}}) \to \pi(\widetilde{\bm{\theta}})$ as $\alpha \to 0$ and $\alpha_a \to 0$.

Let us consider the convergence rate in terms of the $L_1$-norm 

\begin{align*}
    \norm{\pi_{\gamma} -\pi}_1=\int_{{\theta_a}}\sum_{{\theta} \in {\bm{\Theta}} }\Abs{\frac{Z'_{\gamma}(\widetilde{\bm{\theta}})\pi(\widetilde{\bm{\theta}})}{\int_{y} \sum_{x \in \bm{\Theta}} Z'_{\gamma}([x^{\top},y^{\top}]^{\top}) \pi([x^{\top},y^{\top}]^{\top}) dy} - \pi(\widetilde{\bm{\theta}})} d\bm{\theta}_a
\end{align*}

We write out each absolute value term 
\begingroup\makeatletter\def\f@size{8}\check@mathfonts
\def\maketag@@@#1{\hbox{\m@th\large\normalfont#1}}
\begin{align*}
\Abs{\frac{Z'_{\gamma}(\widetilde{\bm{\theta}})\pi(\widetilde{\bm{\theta}})}{\int_{y} \sum_{x \in \bm{\Theta}} Z'_{\gamma}([x^{\top},y^{\top}]^{\top}) \pi([x^{\top},y^{\top}]^{\top}) dy} - \pi(\widetilde{\bm{\theta}})}
     &=
      \pi(\widetilde{\bm{\theta}})\Abs{\frac{Z'_{\gamma}(\widetilde{\bm{\theta}})}{\int_{y} \sum_{x \in \bm{\Theta}} Z'_{\gamma}([x^{\top},y^{\top}]^{\top}) \pi([x^{\top},y^{\top}]^{\top}) dy} - 1}
\end{align*}
\endgroup
First, we note that 
    since $U$ is M-gradient Lipschitz and $\frac{\alpha}{2}<\frac{1}{M}$,  the matrix \[\frac{1}{2\alpha}I-\frac{1}{4}\int_{0}^{1} \nabla^2 U((1-s)\bm{\theta}+s\bm{\theta}')ds>\frac{1}{4}\left(\frac{2}{\alpha}-M\right) I\] is positive definite.

Second, for \( x' \in \bm{\Theta} \) and \( y' \in \bm{\Theta}_a \) (under Assumptions \ref{assumption:5.1} and \ref{assumption:5.3}), we know that the following minimum exists and is well-defined:
$\min_{\substack{x \in \bm{\Theta} \setminus \{x'\} \\ y \in \bm{\Theta}_a \setminus \{y'\}}} (x - y)^{\top}(x' - y')$

Thus when, 
$
\frac{Z'_{\gamma}(\widetilde{\bm{\theta}})}{\int_{y} \sum_{x \in \bm{\Theta}} Z'_{\gamma}\left(\begin{bmatrix}x^{\top} \\ y^{\top}\end{bmatrix}\right) \pi\left(\begin{bmatrix}x^{\top} \\ y^{\top}\end{bmatrix}\right) dy} - 1 \geq 0$, we get,
\[
\scalebox{0.55}{
$\begin{aligned}
\left|\frac{Z'_{\gamma}(\widetilde{\bm{\theta}})\pi(\widetilde{\bm{\theta}})}{\int_{y} \sum_{x \in \bm{\Theta}} Z'_{\gamma}\left(\begin{bmatrix}x^{\top} \\ y^{\top}\end{bmatrix}\right) \pi\left(\begin{bmatrix}x^{\top} \\ y^{\top}\end{bmatrix}\right) dy} - \pi(\widetilde{\bm{\theta}})\right|
&= \pi(\widetilde{\bm{\theta}}) \left| \frac{Z'_{\gamma}(\widetilde{\bm{\theta}})}{\int_{y} \sum_{x \in \bm{\Theta}} Z'_{\gamma}\left(\begin{bmatrix}x^{\top} \\ y^{\top}\end{bmatrix}\right) \pi\left(\begin{bmatrix}x^{\top} \\ y^{\top}\end{bmatrix}\right) dy} - 1 \right| \\
&\leq \pi(\widetilde{\bm{\theta}}) \left( 1 + \frac{1}{\sqrt{(2\pi\alpha_a)^d}} \int_{y \neq \theta_a} \sum_{x \neq \theta} \exp \left( \frac{1}{2} (U(x) - U(\bm{\theta}))- \frac{1}{2} (\bm{\theta} - x)^{\top} \left( \frac{1}{\alpha} I + \frac{1}{2} \int_{0}^{1} \nabla^2 U((1-s)\theta + sx) \, ds \right) (\bm{\theta} - x) \right. \right. \\
&\quad \left. \left. - \frac{1}{2\alpha_a} \| y - \bm{\theta}_a \|^2 + \frac{4\eta}{8\eta^2} \| \bm{\theta} - \bm{\theta}_a \|^2 - \frac{1}{2\eta} (\bm{\theta} - \bm{\theta}_a)^{\top} (x - y) \right) dy - 1 \right) \\
&\leq \frac{\pi(\widetilde{\bm{\theta}})}{\sqrt{(2\pi\alpha_a)^d}} \exp\left( \frac{M}{4} - \frac{1}{2\alpha} + \frac{1}{2\eta} \| \bm{\theta} - \bm{\theta}_a \|^2 - \frac{\vartheta(\bm{\Theta}, \bm{\Theta}_a)}{2\eta} \right)\cdot \left( \int_{y \neq \theta_a} \sum_{x \neq \theta} \exp\left( \frac{1}{2} U(x) - \frac{1}{2} U(\bm{\theta}) - \frac{1}{2\alpha_a} \| y - \theta_a \|^2 \right) \, dy \right) \\
&\leq \pi(\widetilde{\bm{\theta}}) \exp\left( \frac{M}{4} - \frac{1}{2\alpha} + \frac{1}{2\eta} \| \bm{\theta} - \bm{\theta}_a \|^2 - \frac{\vartheta(\bm{\Theta}, \bm{\Theta}_a)}{2\eta} \right) \left( \sum_{x} \exp\left( U(x) \right) \right) \\
&= \pi(\widetilde{\bm{\theta}}) Z \exp\left( \frac{M}{4} - \frac{1}{2\alpha} + \frac{1}{2\eta} \| \bm{\theta} - \bm{\theta}_a \|^2 - \frac{\vartheta(\bm{\Theta}, \bm{\Theta}_a)}{2\eta} \right) \\
&\leq \pi(\widetilde{\bm{\theta}}) Z \exp\left( \frac{M}{4} - \frac{1}{2\alpha} + \frac{\Delta(\bm{\Theta}, \bm{\Theta}_a)^2 - \vartheta(\bm{\Theta}, \bm{\Theta}_a)}{2\eta} \right).
\end{aligned}$
}
\]
Similarly, when $ \frac{Z'_{\gamma}(\widetilde{\bm{\theta}})}{\int_{y} \sum_{x \in \bm{\Theta}} Z'_{\gamma}\left(\begin{bmatrix}x^{\top} \\ y^{\top}\end{bmatrix}\right) \pi\left(\begin{bmatrix}x^{\top} \\ y^{\top}\end{bmatrix}\right) dy} - 1 < 0 $, we get

\[
\scalebox{0.4}{
$\begin{aligned}
      &\left| 
      \frac{Z'_{\gamma}(\widetilde{\bm{\theta}})\pi(\widetilde{\bm{\theta}})}
      {\int_{y} \sum_{x \in \bm{\Theta}} Z'_{\gamma}
      \left(
      \begin{bmatrix}
      x^{\top} \\
      y^{\top}
      \end{bmatrix}
      \right)
      \pi
      \left(
      \begin{bmatrix}
      x^{\top} \\
      y^{\top}
      \end{bmatrix}
      \right) dy} 
      - \pi(\widetilde{\bm{\theta}})
      \right| \\[1.5em]
     &=
     \pi(\widetilde{\bm{\theta}})
     \left(
     1 -
     \frac{
     1 + 
     \frac{1}{\sqrt{(2\pi\alpha_a)^d}}
     \int_{y \neq \theta_a} \sum_{x\neq \theta} 
     \exp\left(
     \frac{1}{2} (U(x)- U(\bm{\theta})) 
     - \frac{1}{2}  (\bm{\theta} - x)^{\top} 
     \left(
     \frac{1}{\alpha}I + \frac{1}{2}\int_{0}^{1} \nabla^2 U((1-s)\theta+sx)ds 
     \right) (\bm{\theta} - x)
     - \frac{1}{2\alpha_a} \|y - \bm{\theta_a}\|^2
     + \frac{4\eta}{8\eta^2}\| \bm{\theta}-\bm{\theta}_a \|^2 
     -\frac{1}{2\eta} (\bm{\theta}-\bm{\theta}_a)^{\top}( x-y)
     \right) dy
     }
     {1 + 
     \frac{1}{\sqrt{2\pi\alpha_a}^d}
     \int_{p}\frac{1}{\sqrt{\pi}^d}\exp\left(-p^2\right) 
     \int_{q \neq p} \sum_r \frac{1}{Z}\exp\left(U(r)\right)
     \sum_{s\neq r}
     \exp\left(
     \frac{1}{2} \left(U(s)-\frac{1}{2} U(r)\right)
     - \frac{1}{2}  (r-s)^{\top} 
     \left(
     \frac{1}{\alpha}I + \frac{1}{2}\int_{0}^{1} \nabla^2 U((1-l)r+ls)dl 
     \right) (r-s) 
     - \frac{1}{2\alpha_a} \|q - p\|^2
     + \frac{4\eta}{8\eta^2}\|r - p\|^2 
     -\frac{1}{2\eta}(r-p)^{\top}( s-q)
     \right) \, dq \, dp 
     }
     \right) \\[1.5em]
     &\le
    \pi(\widetilde{\bm{\theta}})
     \left(
     1 -
     \frac{
     1
     }{
     1 + 
     \frac{1}{\sqrt{2\pi\alpha_a}^d}
     \int_{p}\frac{1}{\sqrt{\pi}^d}\exp\left(-p^2\right) 
     \int_{q \neq p} 
     \exp\left(- \frac{1}{2\alpha_a} \|q-p\|^2 \right) 
     \sum_r 
     \exp\left(\frac{4\eta}{8\eta^2}\|r - p\|^2\right) 
     \frac{1}{Z}\exp\left(U(r)\right)
     \sum_{s\neq r}
     \exp\left(
     \frac{1}{2}( U(s)-U(r)) - 
     \frac{1}{2}  (r-s)^{\top} 
     \left(
     \frac{1}{\alpha}I + \frac{1}{2}\int_{0}^{1} \nabla^2 U((1-l)r+ls)dl 
     \right) (r-s) 
     -\frac{1}{2\eta}(r-p)^{\top}( s-q)
     \right) \, dq \, dp
     }
     \right) \\[1.5em]
     &=
     \pi(\widetilde{\bm{\theta}})
     \left(
     \frac{
     \frac{1}{\sqrt{2\pi\alpha_a}^d}
     \int_{p}\frac{1}{\sqrt{\pi}^d}\exp\left(-p^2\right) 
     \int_{q \neq p} 
     \exp\left(- \frac{1}{2\alpha_a} \|q-p\|^2 \right) 
     \sum_r 
     \exp\left(\frac{4\eta}{8\eta^2}\|r - p\|^2\right) 
     \frac{1}{Z}\exp\left(U(r)\right)
     \sum_{s\neq r}
     \exp\left(
     \frac{1}{2} (U(s)- U(r)) 
     - \frac{1}{2}  (r-s)^{\top} 
     \left(
     \frac{1}{\alpha}I + \frac{1}{2}\int_{0}^{1} \nabla^2 U((1-l)r+ls)dl 
     \right) (r-s) 
     -\frac{1}{2\eta}(r-p)^{\top}( s-q)
     \right) \, dq \, dp
     }{
     1 + 
     \frac{1}{\sqrt{2\pi\alpha_a}^d}
     \int_{p}\frac{1}{\sqrt{\pi}^d}\exp\left(-p^2\right) 
     \int_{q \neq p} 
     \exp\left(- \frac{1}{2\alpha_a} \|q-p\|^2 \right) 
     \sum_r 
     \exp\left(\frac{4\eta}{8\eta^2}\|r - p\|^2\right) 
     \frac{1}{Z}\exp\left(U(r)\right)
     \sum_{s\neq r}
     \exp\left(
     \frac{1}{2}( U(s)-U(r)) - 
     \frac{1}{2}  (r-s)^{\top} 
     \left(
     \frac{1}{\alpha}I + \frac{1}{2}\int_{0}^{1} \nabla^2 U((1-l)r+ls)dl 
     \right) (r-s) 
     -\frac{1}{2\eta}(r-p)^{\top}( s-q)
     \right) \, dq \, dp
     }
     \right) \\[1.5em]
     &\le \frac{\pi(\widetilde{\bm{\theta}})}{\sqrt{2\pi\alpha_a}^d} \left(\int_{p} \frac{1}{\sqrt{\pi}^d}\exp\left(-p^2\right) \int_{q \neq p} \exp\left(- \frac{1}{2\alpha_a} \norm{q-p}^2 \right) \sum_r \exp\left( \frac{4\eta}{8\eta^2}\norm{r - p}^2\right) \frac{1}{Z}\exp\left(U(r)\right)\sum_{s\neq r}\exp\left(\frac{1}{2} (U(s)- U(r))- \frac{1}{2}(r-s)^{\top} \left(\frac{1}{\alpha}I + \frac{1}{2}\int_{0}^{1} \nabla^2 U((1-l)r+ls)dl\right) (r-s)
     -\frac{1}{2\eta}(r-p)^{\top}( s-q)\right) \right) dq  dp \\[1.5em]
     &\le \frac{\pi(\widetilde{\bm{\theta}})}{\sqrt{(2\pi\alpha_a)^d}} 
    \exp\left( \frac{M}{4}- \frac{1}{2\alpha} \right)
    \left( \int_{p} \frac{1}{\sqrt{\pi}^d} \exp\left(-p^2\right) \right) \int_{q \neq p} \exp\left(- \frac{1}{2\alpha_a} \|q-p\|^2 \right) 
    \sum_r \exp\left( \frac{1}{2\eta}\|r - p\|^2\right) 
    \frac{1}{Z} \exp\left(U(r)\right)\sum_{s \neq r} \exp\left(\frac{1}{2} (U(s)- U(r)) - \frac{1}{2\eta}(r-p)^{\top}( s-q)\right) \Bigg) dq dp\\[1.5em]
     &\le
      \frac{\pi(\widetilde{\bm{\theta}})}{\sqrt{2\pi\alpha_a}^d}\exp\left( \frac{M}{4}- \frac{1}{2\alpha} +\frac{\Delta(\bm{\Theta}, \bm{\Theta}_a)^2-\vartheta{(\bm{\Theta}, \bm{\Theta}_a)}}{2\eta}\right) \left(\int_{p} \frac{1}{\sqrt{\pi}^d}\exp\left(-p^2\right) \int_{q \neq p} \exp\left(- \frac{1}{2\alpha_a} \norm{q-p}^2 \right)\sum_r \frac{1}{Z}\left(U(r) \right)\sum_{s\neq r} \exp\left(\frac{1}{2} (U(s)-U(r))
     \right)\, dq \, dp \right)\\[1.5em]
     &\le
     \frac{\pi(\widetilde{\bm{\theta}})}{\sqrt{2\pi\alpha_a}^d}Z\exp\left(\frac{M}{4}- \frac{1}{2\alpha} +\frac{\Delta(\bm{\Theta}, \bm{\Theta}_a)^2-\vartheta{(\bm{\Theta}, \bm{\Theta}_a)}}{2\eta}\right)\left(\int_{p} \frac{1}{\sqrt{\pi}^d}\exp\left(-p^2\right) \int_{q \neq p} \exp\left(- \frac{1}{2\alpha_a} \norm{q-p}^2 \right) 
      \, dq \, dp \right)\\[1.5em]
      &=
    {\pi(\widetilde{\bm{\theta}})}Z\exp\left(\frac{M}{4}- \frac{1}{2\alpha} +\frac{\Delta(\bm{\Theta}, \bm{\Theta}_a)^2-\vartheta{(\bm{\Theta}, \bm{\Theta}_a)}}{2\eta}\right)\int_{ p}\left(\frac{1}{\sqrt{\pi}^d}\exp\left(-p^2\right)  \right) \, dp \\[1.5em]
      &={\pi(\widetilde{\bm{\theta}})}Z\exp\left(\frac{M}{4}- \frac{1}{2\alpha} +\frac{\Delta(\bm{\Theta}, \bm{\Theta}_a)^2-\vartheta{(\bm{\Theta}, \bm{\Theta}_a)}}{2\eta}\right)
\end{aligned}$}
\]

Therefore, the difference between $\pi_{\gamma}$ and $\tilde{\pi}$ can be bounded as follows
\begin{align*}
    \norm{\pi_{\gamma} -\tilde{\pi}}_1 &\le \int_{\theta_a}\sum_{{\theta} \in {\bm{\Theta}}}  {\pi(\widetilde{\bm{\theta}})}Z\exp\left(\frac{M}{4} -\frac{1}{2\alpha}+\frac{\Delta(\bm{\Theta}, \bm{\Theta}_a)^2-\vartheta{(\bm{\Theta}, \bm{\Theta}_a)}}{2\eta} \right) d\theta_a\\
    &\leq Z\exp\left(\frac{M}{4} -\frac{1}{2\alpha}+\frac{\Delta(\bm{\Theta}, \bm{\Theta}_a)^2-\vartheta{(\bm{\Theta}, \bm{\Theta}_a)}}{2\eta} \right)
\end{align*}

\subsection{Proofs for EDULA}
We start by establishing results for a more general case in which Assumption~\ref{assumption:5.3} is dropped. We establish that in this setting geometric rates of convergence exist. However, in this case proving that the stationary distribution is close to the target remains an open problem. .
\begin{theorem}\label{thm:drift:1}
   Let Assumption~\ref{assumption:5.1} hold. Then for the Markov chain with transition operator \( P \) as in Algorithm \ref{alg:edlp}, the drift condition is satisfied as follows:
   \[
   P V(\tilde{\bm\theta}) \le \alpha_a \, d + 2\, \left(1 - \frac{\alpha_a}{\eta}\right)^2 V(\tilde{\bm\theta}) + 2\, \frac{\alpha^2_a}{\eta^2} \sup_{\bm{\theta} \in \bm{\Theta}} \|\bm{\theta}\|^2.
   \]
\end{theorem}

\emph{Proof.} We establish an explicit drift and minorization condition for the joint chain, which confirms the convergence rate. Note that
\[
p((\bm{\theta}'_{a}, \bm{\theta}') \mid (\bm{\theta}'_{a}, \bm{\theta}')) = p(\bm{\theta}'_{a} \mid \bm{\theta}, \bm{\theta}_{a}) \cdot p(\bm{\theta}' \mid \bm{\theta}_{a}, \bm{\theta}).
\]
Now,
\begin{align*}
    p(\bm{\theta}'_{a} \mid \bm{\theta}, \bm{\theta}_{a}) &= \frac{1}{(2\pi \alpha_a)^{d/2}} \exp\left\{-\frac{1}{2 \alpha_a} \left\|\bm{\theta}'_{a} - \bm{\theta}_{a} \left(1 - \frac{\alpha_a}{\eta}\right) - \frac{\alpha_a}{\eta} \bm{\theta}\right\|^2\right\}
\end{align*}
and 
\begin{align*}
p(\bm{\theta}' \mid \bm{\theta}_{a}, \bm{\theta}) = \frac{\exp\left\{-\frac{1}{2 \alpha} \left\|\bm{\theta}' - \bm{\theta} + \alpha \nabla U(\bm{\theta}) - \frac{\alpha}{\eta} \left(\bm{\theta} - \bm{\theta}_{a}\right)\right\|^2\right\}}{\sum_{\bm{x} \in \bm{\Theta}} \exp\left\{-\frac{1}{2 \alpha} \left\|\bm{x} - \bm{\theta} + \alpha \nabla U(\bm{\theta}) - \frac{\alpha}{\eta} \left(\bm{\theta} - \bm{\theta}_{a}\right)\right\|^2\right\}}.
\end{align*}
Therefore, our Markov transition operator \( P \) is given as
\[
P((\bm{\theta}_{a}, \bm{\theta}), A) = \int_A p((\bm{\theta}'_{a}, \bm{\theta}') \mid (\bm{\theta}, \bm{\theta}_{a})) \, d\mu,
\]
where \( A \in \bm{\Theta} \times \mathbb{R}^d \) and \( \mu \) is the product of the counting measure and Lebesgue measure. 

We shall first establish a drift condition:
\[
P V \le \lambda V + b,
\]
where we choose the Lyapunov function \( V(\bm{x}_1, \bm{x}_2) = \|\bm{x}_1\|^2 \) and some constant \( b > 0 \). 

We note that 
\begin{align*}
    P V(\bm{\theta}_{a}, \bm{\theta}) &= \frac{1}{(2\pi \alpha_a)^{d/2}} \sum_{\bm{\theta}' \in \bm{\Theta}} \int \|\bm{\theta}'_{a}\|^2 \, \exp \left\{-\frac{1}{2 \alpha_a} \left\|\bm{\theta}'_{a} - \bm{\theta}_{a} \left(1 - \frac{\alpha_a}{\eta}\right) - \frac{\alpha_a}{\eta} \bm{\theta}\right\|^2\right\} \\
    & \quad \cdot \frac{\exp\left\{-\frac{1}{2 \alpha} \left\|\bm{\theta}' - \bm{\theta} + \alpha \nabla U(\bm{\theta}) - \frac{\alpha}{\eta} \left(\bm{\theta} - \bm{\theta}_{a}\right)\right\|^2\right\}}{\sum_{\bm{x} \in \bm{\Theta}} \exp\left\{-\frac{1}{2 \alpha} \left\|\bm{x} - \bm{\theta} + \alpha \nabla U(\bm{\theta}) - \frac{\alpha}{\eta} \left(\bm{\theta} - \bm{\theta}_{a}\right)\right\|^2\right\}} \, d\bm{\theta}_a.
\end{align*}
Using a change of variables, we have
\begin{align*}
    P V(\bm{\theta}_{a}, \bm{\theta}) &= \frac{1}{(2\pi \alpha_a)^{d/2}} \sum_{\bm{\theta}' \in \bm{\Theta}} \int \left\|u + \bm{\theta}_{a} \left(1 - \frac{\alpha_a}{\eta}\right) + \frac{\alpha_a}{\eta} \bm{\theta}\right\|^2 \, \exp \left\{-\frac{1}{2 \alpha_a} \|u\|^2\right\} \\
    & \quad \cdot \frac{\exp\left\{-\frac{1}{2 \alpha} \left\|\bm{\theta}' - \bm{\theta} + \alpha \nabla U(\bm{\theta}) - \frac{\alpha}{\eta} \left(\bm{\theta} - \bm{\theta}_{a}\right)\right\|^2\right\}}{\sum_{\bm{x} \in \bm{\Theta}} \exp\left\{-\frac{1}{2 \alpha} \left\|\bm{x} - \bm{\theta} + \alpha \nabla U(\bm{\theta}) - \frac{\alpha}{\eta} \left(\bm{\theta} - \bm{\theta}_{a}\right)\right\|^2\right\}} \, du \\
    &\le \alpha_a \, d + 2 \, \left(1 - \frac{\alpha_a}{\eta}\right)^2 \|\bm{\theta}_{a}\|^2 + 2 \, \frac{\alpha^2_a}{\eta^2} \sup_{\bm{\theta} \in \bm{\Theta}} \|\bm{\theta}\|^2.
\end{align*}
Note that when \( \lambda = 2 \left(1 - \frac{\alpha_a}{\eta}\right)^2 < 1 \), then this is a proper drift condition with \( b = \alpha_a \, d + 2 \, \frac{\alpha^2_a}{\eta^2} \sup_{\bm{\theta} \in \bm{\Theta}} \|\bm{\theta}\|^2 \).

\begin{theorem}\label{thm:minorization:1}
Under Assumption \ref{assumption:5.1}, the Markov chain with transition operator \( P \) as in Algorithm \ref{alg:edlp} satisfies,
    \[P(\tilde{\bm\theta}, A) \ge \bar{\eta} \mu(A)\] where $\bar{\eta}>0$ is defined in \eqref{eta:def} and $\mu(\cdot)$ is the product of Lebesgue measure and counting measure and $\tilde{\bm\theta} \in C_{\alpha}$ as in \eqref{C:def} .
\end{theorem}

\emph{Proof.} 
We establish a minorization on the set,
\begin{align}\label{C:def}
C_{\alpha_a}=\left\{x: V(x) \le \frac{2\, \left(\alpha_a \, d+2\, \frac{\alpha^2_a}{\eta^2} \sup_{\bm{\theta} \in \bm{\Theta}} \|\bm{\theta}\|^2\right)}{\left(1-\frac{\alpha_a}{\eta}\right)^2}\right\}
\end{align}
We define
{\scriptsize
\begin{align}\label{eta:def}
\begin{split}
    \bar{\eta}&= \frac{1}{(2\pi \alpha_a)^{d/2}} \exp\left\{-\frac{4}{\alpha_a}\frac{\left(\alpha_a \, d+2\, \frac{\alpha^2_a}{\eta^2} \sup_{\bm{\theta} \in \bm{\Theta}} \|\bm{\theta}\|^2\right)}{\left(1-\frac{\alpha_a}{\eta}\right)^2}\right\}\cdot\frac{1}{|\bm{\Theta}|} \\
&\cdot\exp\left\{-\frac{1}{2\alpha }\left[\left(\left(\alpha \, M +1\right)^2+\alpha \, M^2 \right) \text{diam}(\bm{\Theta})^2 +\left(2\, (M +\alpha)+2\alpha M\right) \|\nabla U(a)\| \text{diam}(\bm{\Theta}) +\left(\alpha^2+\alpha\right) \|\nabla U(a)\|^2 \right.\right.\\
&\left.\left. \quad \quad \quad \quad \quad  +2 \frac{\alpha}{\eta} \left[\left(\alpha \, M +1\right)^2 \text{diam}(\bm{\Theta})^2 +2\, (M +\alpha) \|\nabla U(a)\| \text{diam}(\bm{\Theta}) +\alpha^2 \|\nabla U(a)\|^2\right]^{1/2} \text{diam}(\bm{\Theta})\right]\right\}    
\end{split}
\end{align}
}

We start with considering any $(\bm{\theta_1}, \bm{\theta_2}) \in C_{\alpha}$. Further, we also have $(\bm{\theta_a}, \bm{\theta}) \in C_{\alpha_a}$.  Therefore
    \begin{align*}
    p((\bm{\theta_1},\bm{\theta_2})\mid (\bm{\theta_a}, \bm{\theta}))&=\frac{1}{(2\pi \alpha_a)^{d/2}} \, \exp \left\{-\frac{1}{2 \alpha_a} \left\|\bm{\theta_1}-\bm{\theta_a} \left(1-\frac{\alpha_a}{\eta}\right)-\frac{\alpha_a}{\eta} \bm{\theta}\right\|^2\right\}\\
    & \quad \quad \quad \quad \quad \cdot \frac{\exp\left\{-\frac{1}{2\, \alpha}\left\|\bm{\theta_2}-\bm{\theta} +\alpha \nabla U(\bm{\theta})-\frac{\alpha}{\eta}\left(\bm{\theta}-\bm{\theta_a}\right)\right\|^2\right\}}{\sum_{x \in \bm{\Theta}}\exp\left\{-\frac{1}{2\, \alpha}\left\|x-\bm{\theta} +\alpha \nabla U(\bm{\theta})-\frac{\alpha}{\eta}\left(\bm{\theta}-\bm{\theta_a}\right)\right\|^2\right\}}.
    \end{align*}
For the first term, we note that 
\begin{align*}
 \left\|\bm{\theta_1}-\bm{\theta_a} \left(1-\frac{\alpha_a}{\eta}\right)-\frac{\alpha_a}{\eta} \bm{\theta}\right\|^2 & \le 2\,\left\|\bm{\theta_1}\right\|^2+2\, \left\|\left(1-\frac{\alpha_a}{\eta}\right)\bm{\theta_a} +\frac{\alpha_a}{\eta} \bm{\theta}\right\|^2 \\
 & \le 2\,\left\|\bm{\theta_1}\right\|^2+2\, \left(1-\frac{\alpha_a}{\eta}\right)\left\|\bm{\theta_a} \right\|^2+2\, \frac{\alpha_a}{\eta} \left\|\bm{\theta}\right\|^2\\
 & \le 8 \, \frac{\left(\alpha_a \, d+2\, \frac{\alpha^2_a}{\eta^2} \sup_{\bm{\theta} \in \bm{\Theta}} \|\bm{\theta}\|^2\right)}{\left(1-\frac{\alpha_a}{\eta}\right)^2}.
\end{align*}
Therefore, the first term is greater than 
\begin{align*}
& \frac{1}{(2\pi \alpha_a)^{d/2}} \, \exp \left\{-\frac{1}{2 \alpha_a} \left\|\bm{\theta_1}-\bm{\theta_a} \left(1-\frac{\alpha_a}{\eta}\right)-\frac{\alpha_a}{\eta} \bm{\theta_2}\right\|^2\right\} \\
& \ge \frac{1}{(2\pi \alpha_a)^{d/2}} \exp\left\{-\frac{4}{\alpha_a}\frac{\left(\alpha_a \, d+2\, \frac{\alpha^2_a}{\eta^2} \sup_{\bm{\theta} \in \bm{\Theta}} \|\bm{\theta}\|^2\right)}{\left(1-\frac{\alpha_a}{\eta}\right)^2}\right\}.
\end{align*}
For the second term, note that 
{\scriptsize
\begin{align*}
\frac{\exp\left\{-\frac{1}{2\, \alpha}\left\|\bm{\theta_2}-\bm{\theta} +\alpha \nabla U(\bm{\theta})-\frac{\alpha}{\eta}\left(\bm{\theta}-\bm{\theta_a}\right)\right\|^2\right\}}{\sum_{x \in \bm{\Theta}}\exp\left\{-\frac{1}{2\, \alpha}\left\|x-\bm{\theta} +\alpha \nabla U(\bm{\theta})-\frac{\alpha}{\eta}\left(\bm{\theta}-\bm{\theta_a}\right)\right\|^2\right\}} & \ge \frac{1}{\left|\bm{\Theta}\right|} \exp\left\{-\frac{1}{2\, \alpha}\left\|\bm{\theta_2}-\bm{\theta} +\alpha \nabla U(\bm{\theta})-\frac{\alpha}{\eta}\left(\bm{\theta}-\bm{\theta_a}\right)\right\|^2\right\}.
\end{align*}
}
For the numerator, one sees,
{
\begin{align*}
    \left\|\bm{\theta_2}-\bm{\theta} +\alpha \nabla U(\bm{\theta}) -\frac{\alpha}{\eta} \left(\bm{\theta}-\bm{\theta_a}\right)\right\|^2 
    &\le  \left\|\bm{\theta_2}-\bm{\theta} +\alpha \nabla U(\bm{\theta})\right\|^2+ \frac{\alpha^2}{\eta^2} \left\|\bm{\theta}-\bm{\theta_a}\right\|^2 \\
    &+2 \frac{\alpha}{\eta}\left\|\bm{\theta_2}-\bm{\theta} +\alpha \nabla U(\bm{\theta})\right\|\left\|\bm{\theta}-\bm{\theta_a}\right\|.
\end{align*}
}
For the first term, we have 
\begin{align*}
    \left\|\bm{\theta_2}-\bm{\theta} +\alpha \nabla U(\bm{\theta})\right\|^2 \le \left\|\bm{\theta_2}-\bm{\theta}\right\|^2+\alpha^2 \|\nabla U(\bm{\theta})\|^2+2\, \alpha \left\|\bm{\theta_2}-\bm{\theta}\right\|  \|\nabla U(\bm{\theta})\|.
\end{align*}
Define $a=\text{argmin}_{\bm{\theta} \in \bm{\Theta}} \|\nabla U(\bm{\theta})\|$. Therefore, the above expression is less than 
\begin{align*}
    \left\|\bm{\theta_2}-\bm{\theta} +\alpha \nabla U(\bm{\theta})\right\|^2 &\le \text{diam}(\bm{\Theta})^2 + \alpha^2 \left(M^2 \text{diam}(\bm{\Theta})^2 +\|\nabla U(a)\|^2 + 2\, M \, \text{diam}(\bm{\Theta}) \|\nabla U(a)\|\right)\\
    &\quad \quad +2\alpha \, \text{diam}(\bm{\Theta}) \left(M\, \text{diam}(\bm{\Theta})+\|\nabla U(a)\|\right)\\
    & \le \left(\alpha \, M +1\right)^2 \text{diam}(\bm{\Theta})^2 +2\, (M +\alpha) \|\nabla U(a)\| \text{diam}(\bm{\Theta}) +\alpha^2 \|\nabla U(a)\|^2.
\end{align*}
For the second term, we have 
\begin{align*}
\alpha \|\nabla U(\bm{\theta})\|^2 \le \alpha M^2 \text{diam}(\bm{\Theta})^2 + \alpha \|\nabla U(a)\|^2 +2 \alpha \, M\,  \text{diam}(\bm{\Theta}) \|\nabla U(a)\|
\end{align*}
and for the final term we have 

\begin{align}
  2 \frac{\alpha}{\eta}\left\|\bm{\theta_2}-\bm{\theta} +\alpha \nabla U(\bm{\theta})\right\|\left\|\bm{\theta}-\bm{\theta_a}\right\| 
  &\leq  2 \frac{\alpha}{\eta} \left[\left(\alpha M + 1\right)^2 \text{diam}(\bm{\Theta})^2 \right. \nonumber + 2 (M + \alpha) \|\nabla U(a)\| \text{diam}(\bm{\Theta}) \\
  &\quad \left. 
  + \alpha^2 \|\nabla U(a)\|^2 \right]^{1/2} \text{diam}(\bm{\Theta}).
\end{align}

Therefore we have
{\scriptsize
\begin{align*}
&\frac{\exp\left\{-\frac{1}{2\, \alpha}\left\|\bm{\theta_2}-\bm{\theta} +\alpha \nabla U(\bm{\theta})-\frac{\alpha}{\eta}\left(\bm{\theta}-\bm{\theta_a}\right)\right\|^2\right\}}{\sum_{x \in \bm{\Theta}}\exp\left\{-\frac{1}{2\, \alpha}\left\|x-\bm{\theta} +\alpha \nabla U(\bm{\theta})-\frac{\alpha}{\eta}\left(\bm{\theta}-\bm{\theta_a}\right)\right\|^2\right\}}\\
& \ge \frac{1}{|\bm{\Theta}|} \exp\left\{-\frac{1}{2\alpha }\left[\left(\left(\alpha \, M +1\right)^2+\alpha \, M^2 \right) \text{diam}(\bm{\Theta})^2 +\left(2\, (M +\alpha)+2\alpha M\right) \|\nabla U(a)\| \text{diam}(\bm{\Theta}) +\left(\alpha^2+\alpha\right) \|\nabla U(a)\|^2 \right.\right.\\
&\left.\left. \quad \quad \quad \quad \quad  +2 \frac{\alpha}{\eta} \left[\left(\alpha \, M +1\right)^2 \text{diam}(\bm{\Theta})^2 +2\, (M +\alpha) \|\nabla U(a)\| \text{diam}(\bm{\Theta}) +\alpha^2 \|\nabla U(a)\|^2\right]^{1/2} \text{diam}(\bm{\Theta})\right]\right\}.
\end{align*}
}
This finally gives $\tilde{\eta}$ as 
{\scriptsize
\begin{align*}
    \bar{\eta}&= \frac{1}{(2\pi \alpha_a)^{d/2}} \exp\left\{-\frac{4}{\alpha_a}\frac{\left(\alpha_a \, d+2\, \frac{\alpha^2_a}{\eta^2} \sup_{\bm{\theta} \in \bm{\Theta}} \|\bm{\theta}\|^2\right)}{\left(1-\frac{\alpha_a}{\eta}\right)^2}\right\}\\
&\cdot \frac{1}{|\bm{\Theta}|} \exp\left\{-\frac{1}{2\alpha }\left[\left(\left(\alpha \, M +1\right)^2+\alpha \, M^2 \right) \text{diam}(\bm{\Theta})^2 +\left(2\, (M +\alpha)+2\alpha M\right) \|\nabla U(a)\| \text{diam}(\bm{\Theta}) +\left(\alpha^2+\alpha\right) \|\nabla U(a)\|^2 \right.\right.\\
&\left.\left. \quad \quad \quad \quad \quad  +2 \frac{\alpha}{\eta} \left[\left(\alpha \, M +1\right)^2 \text{diam}(\bm{\Theta})^2 +2\, (M +\alpha) \|\nabla U(a)\| \text{diam}(\bm{\Theta}) +\alpha^2 \|\nabla U(a)\|^2\right]^{1/2} \text{diam}(\bm{\Theta})\right]\right\}
\end{align*}
}
with the reference measure $\mu(\cdot)$ is the product measure of the Lebesgue measure and the counting measure.

\begin{lemma}\label{lemma:Harris:1}
The Markov chain defined by Algorithm \ref{alg:edlp} is irreducible, aperiodic and Harris recurrent.
\end{lemma}

\emph{Proof. }For any Borel measurable $A$ with $\lambda(A)>0$ and any $\bm{\theta} \in \bm{\Theta}$, we have 
    \begin{align*}
        \mathbb{P}\left(\bm{\theta}'_{a} \in A, \, \bm{\theta}'=\bm{\theta^*} \mid \bm{\theta}_a, \, \bm{\theta}\right)&=\mathbb{P}\left(\bm{\theta}'_{a} \in A \mid \bm{\theta}_a, \,\theta\right) \, \mathbb{P}\left(\bm{\theta}'=\bm{\theta^*} \mid \bm{\theta}_a, \, \bm{\theta}\right).
    \end{align*}
    Note that both the above terms are positive since the first distribution is Gaussian and the second term is positive by definition. We can similarly establish aperiodicity by noting that there is no partition of $\bm{\Theta} \times \mathbb{R}^d$ such that the previous probability is $1$. Finally, due to the fact that the algorithm satisfies a drift condition, the Markov chain is Harris.

We may leverage the above results to obtain a rate of convergence of the sampler using~\cite{10.1214/21-EJS1800}.

\begin{theorem}\label{thm:main}
   The Markov chain has a stationary distribution dependent on $\gamma=(\alpha,\alpha_a)$, $\pi_{\gamma}$, and is $(M,\rho)$ geometrically ergodic with 
	\begin{align*}
	\|P^k (x,\cdot)-\pi_{\gamma}(\cdot)\|_{TV}\le M(x)\rho^k
	\end{align*}
	where \[M(x)=2+\frac{\tilde{b}}{1-\tilde{\lambda}}+\tilde{V}(x)\] and \[\rho\le \max\left\{(1-\bar\eta)^r, \left(\frac{1+2\tilde{b}+\tilde{\lambda}+\tilde{\lambda}d}{1+d}\right)^{1-r}\left(1+2\tilde{b}+2\tilde{\lambda}d\right)^r\right\}\] for some free parameter $0<r<1$ and where $\bar{\eta},\,  b, \, \lambda $ are previously defined.
\end{theorem}

\emph{Proof.} The proof follows directly from Theorem~\ref{thm:drift:1}, Theorem~\ref{thm:minorization:1} and Lemma~\ref{lemma:Harris:1}~\cite{10.1214/21-EJS1800}.

\begin{theorem}\label{thm:clt2}
        For any function $f:\mathbb{R}^p \to \mathbb{R}$ with $f^2(x) \le V(x)$ for all $x \in \mathbb{R}^p$ one has 
    \[\sqrt{n}\left(\bar{f}-\mathbb{E}_{\pi_{\gamma}} f\right) \overset{d}{\rightarrow} N(0, \sigma^2_f)\]
    as $n \to \infty$, where $\sigma^2_f \in [0, \infty)$. , where \[\bar{f}=\frac{1}{n}\sum_{i=1}^{n} f(X_i).\]
\end{theorem}

 \emph{Proof.} The proof follows from Theorem~\ref{thm:drift:1} by noting that 
 $PV \le \lambda V +b$ implies 
 \begin{align*}
     P\left(V+1\right) \le \lambda \left(V+1\right)+\left(b+1-\lambda\right).
 \end{align*}
 This implies a drift condition holds with $V:\mathbb{R}^d \to [1,\infty)$. Hence the result follows via~\cite{10.1214/154957804100000051}.
Note that $\sigma^2_f=0$ implies convergence to a Gaussian degenerate at $0$.

Define
{\small
\begin{align}\label{eta:minor:def}
    \bar{\eta}^* &= \frac{1}{\Phi_{\alpha_a}(\bm{\Theta}_a)} \exp\left\{-\frac{1}{\alpha_a} \operatorname{diam}(\bm{\Theta}_a)^2 - \frac{\alpha_a}{\eta^2} \, \Delta(\bm{\Theta}, \bm{\Theta}_a)^2\right\} \notag \\
    &\quad \times \frac{1}{|\bm{\Theta}|} \exp\left\{-\frac{1}{2 \alpha} \left[ \left(\left(\alpha \, M + 1\right)^2 + \alpha \, M^2\right) \operatorname{diam}(\bm{\Theta})^2 \right.\right. \notag \\
    &\quad\quad\quad\quad\quad + \left(2 (M + \alpha) + 2 \alpha M\right) \|\nabla U(a)\| \operatorname{diam}(\bm{\Theta}) \notag \\
    &\quad\quad\quad\quad\quad + \left(\alpha^2 + \alpha\right) \|\nabla U(a)\|^2 \notag \\
    &\left.\left. \quad\quad\quad\quad\quad + 2 \frac{\alpha}{\eta} \left[\left(\alpha \, M + 1\right)^2 \operatorname{diam}(\bm{\Theta})^2 + 2 (M + \alpha) \|\nabla U(a)\| \operatorname{diam}(\bm{\Theta}) + \alpha^2 \|\nabla U(a)\|^2\right]^{1/2} \operatorname{diam}(\bm{\Theta}) \right]\right\}.
\end{align}
}
\begin{lemma}\label{lemma:minorization}
    Under Assumptions \ref{assumption:5.1} and \ref{assumption:5.3}, the Markov chain with transition operator \( P \) as in Algorithm \ref{alg:edlp} satisfies,\[P((\bm{\theta}_{a},\bm{\theta}), A) \ge \bar{\eta}^* \mu(A)\] where $\bar{\eta}^*>0$ is as defined in \eqref{eta:minor:def} and $\mu(\cdot)$ is the product of Lebesgue measure and counting measure.  
\end{lemma}

 \emph{Proof.} We consider the case where $\bm{\theta}_a$ is restricted to some compact subset of $\mathbb{R}^d$, which we refer to as $\bm{\Theta}_a$. In this case, note that the transition kernel changes to 
\begin{align*}
    p((\bm{\theta}_1, \bm{\theta}_2) \mid (\bm{\theta}_a, \bm{\theta})) &= \frac{1}{\Phi_{\alpha_a}(\bm{\Theta}_a)} \exp\left\{-\frac{1}{2 \alpha_a} \left\|\bm{\theta}_1 - \bm{\theta}_a \left(1 - \frac{\alpha_a}{\eta}\right) - \frac{\alpha_a}{\eta} \bm{\theta} \right\|^2\right\} \\
    &\quad \times \frac{\exp\left\{-\frac{1}{2 \alpha} \left\|\bm{\theta}_2 - \bm{\theta} + \alpha \nabla U(\bm{\theta}) - \frac{\alpha}{\eta} \left(\bm{\theta} - \bm{\theta}_a\right)\right\|^2\right\}}{\sum_{\bm{x} \in \bm{\Theta}} \exp\left\{-\frac{1}{2 \alpha} \left\|\bm{x} - \bm{\theta} + \alpha \nabla U(\bm{\theta}) - \frac{\alpha}{\eta} \left(\bm{\theta} - \bm{\theta}_a\right)\right\|^2\right\}}.
\end{align*}

The proof is similar to Theorem~\ref{thm:minorization:1}. The key difference is that we can minorize on the entire set. Noting that 
    \begin{align*}
       \left\|\bm{\theta}_{1}-\bm{\theta}_a \left(1-\frac{\alpha_a}{\eta}\right)-\frac{\alpha_a}{\eta} \bm{\theta}\right\|^2 
       & \le 2\, \left\|\bm{\theta}_1-\bm{\theta}_a\right\|^2+2\frac{\alpha_a^2}{\eta^2}\left\|\bm{\theta}_a-\bm{\theta}\right\|^2\\
       & \le 2\, \text{diam}(\bm{\Theta}_a)^2+2\frac{\alpha_a^2}{\eta^2}\,  \Delta(\bm{\Theta}, \bm{\Theta}_a)^2.
    \end{align*}
Using the same argument as Theorem~\ref{thm:minorization:1}, we get a uniform minorization with 
{\small
\begin{align*}
    \bar{\eta}^* &= \frac{1}{\Phi_{\alpha_a}(\bm{\Theta}_a)} \exp\left\{-\frac{1}{\alpha_a} \operatorname{diam}(\bm{\Theta}_a)^2 - \frac{\alpha_a}{\eta^2} \, \Delta(\bm{\Theta}, \bm{\Theta}_a)^2\right\} \notag \\
    &\quad \times \frac{1}{|\bm{\Theta}|} \exp\left\{-\frac{1}{2 \alpha} \left[ \left(\left(\alpha \, M + 1\right)^2 + \alpha \, M^2\right) \operatorname{diam}(\bm{\Theta})^2 \right.\right. \notag \\
    &\quad\quad\quad\quad\quad + \left(2 (M + \alpha) + 2 \alpha M\right) \|\nabla U(a)\| \operatorname{diam}(\bm{\Theta}) \notag \\
    &\quad\quad\quad\quad\quad + \left(\alpha^2 + \alpha\right) \|\nabla U(a)\|^2 \notag \\
    &\left.\left. \quad\quad\quad\quad\quad + 2 \frac{\alpha}{\eta} \left[\left(\alpha \, M + 1\right)^2 \operatorname{diam}(\bm{\Theta})^2 + 2 (M + \alpha) \|\nabla U(a)\| \operatorname{diam}(\bm{\Theta}) + \alpha^2 \|\nabla U(a)\|^2\right]^{1/2} \operatorname{diam}(\bm{\Theta}) \right]\right\}.
\end{align*}
}
with the reference measure $\mu(\cdot)$ is the product measure of the Lebesgue measure and the counting measure.

\emph{Proof of Theorem \ref{thm:edula}.} Using Lemma~\ref{lemma:minorization} and Proposition ~\ref{prop:1}, we further have
    \begin{align*}
        \|P^k(x,\cdot)-\tilde{\pi}\|_{TV} \le (1-\bar\eta^*)^k + Z\exp\left(\frac{M}{4} -\frac{1}{2\alpha}+\frac{\Delta(\bm{\Theta}, \bm{\Theta}_a)^2-\vartheta{(\bm{\Theta}, \bm{\Theta}_a)}}{2\eta} \right)
    \end{align*}
    for all  $x \in \mathbb{R}^d$ and $M(x), \rho$ is as defined in Theorem ~\ref{thm:drift:1} itself. Hence we are done.

\begin{theorem}\label{thm:clt}
  Let assumptions~\ref{assumption:5.1}, \ref{assumption:5.3} hold. Then, for any function $f:\mathbb{R}^p \to \mathbb{R}$ with $\|f\|_{\mathbb{L}_{\pi}^2}<\infty$, one has 
    \[\sqrt{n}\left(\bar{f}-\mathbb{E}_{\pi_{\gamma}} f\right) \overset{d}{\rightarrow} N(0, \sigma^2_f)\]
    as $n \to \infty$, where $\sigma^2_f \in [0, \infty)$.
\end{theorem}

 \emph{Proof.} Using Theorem~\ref{thm:edula}, the proof follows directly from \cite{10.1214/154957804100000051}.

\subsection {Proofs for EDMALA}

\begin{proposition}\label{thm:drift:2}
For EDMALA( EDLP with MH step, refer Algorithm~\ref{alg:edlp}) the drift condition is satisfied with drift function $V(x_1,x_2)=\|x_1\|^2$.
\end{proposition}

\emph{Proof.}    The proof follows from Theorem~\ref{thm:drift:1} by observing that 
    \begin{align*}
        P V(\theta_{a},\theta) &\le \int \|\bm{\theta}_{a_1}\|^2 q((\bm{\theta}_{a},\bm{\theta}),(\bm{\theta}_{a_1},\bm{\theta}_1)) d\bm{\theta}_{a_1}+1\\
        & \le \lambda V(\bm{\theta}_{a},\bm{\theta}) +(b+1).
    \end{align*}

\begin{lemma}\label{lemma:minorization_bounded_edmala}
Under Assumptions \ref{assumption:5.1}, \ref{assumption:5.2}, \ref{assumption:5.3}, and $\alpha < \frac{2}{M}$,  for Markov chain P in Algorithm \ref{alg:edlp}, we have for any $\widetilde{\bm{\theta}}, \widetilde{\bm{\theta}'} \in \widetilde{\bm{\Theta}}$, 
\[p(\widetilde{\bm{\theta}}|\widetilde{\bm{\theta}'}) \geq \epsilon_{ \gamma} \frac{\exp \left\{ \frac{1}{2} U(\bm{\theta}')\right\} }{\sum_{x \in \bm{\Theta}} \exp\left( \frac{U(x)}{2}\right) }.\frac{\exp \left\{ -\frac{1}{2\alpha_a} \text{diam}(\bm{\Theta}_a)^2 \right\}}{\Phi_{\alpha_a}(\bm{\bm{\Theta}}_a)} \]
, where 
\[
\epsilon_{\gamma} = 
\exp \left\{
    \begin{aligned}
        &-\left(\frac{M}{2} + \frac{1}{\alpha} - \frac{m}{4}\right) \text{diam}(\bm{\Theta})^2 - \frac{1}{2} \|\nabla U(a)\| \, \text{diam}(\bm{\Theta}) \\
        &- \left(\frac{3\alpha_a}{8\eta^2} + \frac{2}{\eta}\right) \Delta(\bm{\Theta}, \bm{\Theta}_a)^2 + \frac{\vartheta(\bm{\Theta}, \bm{\Theta}_a)}{\eta}
    \end{aligned}
\right\},
\]
$\text{with } a \in \arg\min_{\bm{\theta} \in \bm{\Theta}} \|\nabla U(\bm{\theta})\|$
\end{lemma}

\emph{Proof. }
We follow a similar minorization proof style as of Lemma 5.3 from \cite{pynadath2024gradientbased}.

Notice, 
\begin{align*}
Z_{\gamma}(\widetilde{\bm{\bm{\theta}}}) &\leq \frac{1}{\sqrt{2\pi\alpha_a}^d}\exp\left(-  \frac{U(\bm{\theta})}{2} - \frac{\alpha_a}{8\eta^2} \| \bm{\theta} - \bm{\theta}_a \|^2 + \frac{1}{2\eta} \| \bm{\theta} - \bm{\theta}_a \|^2 \right)\sum_{x \in \bm{\Theta}} \exp\left(  \frac{U(x)}{2}\right) \\
&\int_{y} \sum_{x}  \exp \Bigg( -\frac{1}{2\alpha_a}\norm{y-\bm{\theta}_a}^2- \frac{1}{2\eta} (\bm{\theta} - \bm{\theta}_a)^{\top} (x - y)\Bigg) \, dy\\
&\leq \sum_{x \in \bm{\Theta}} \exp\left(  \frac{U(x)}{2}\right) \exp\left(-  \frac{U(\bm{\theta})}{2} + \frac{1}{2\eta} (\| \bm{\theta} - \bm{\theta}_a \|^2-\vartheta{(\bm{\Theta},\bm{\Theta}_a)}) \right)\\
&\leq \sum_{x \in \bm{\Theta}} \exp\left(  \frac{U(x)}{2}\right)  \exp\left(-  \frac{U(\bm{\theta})}{2} + \frac{\Delta(\bm{\Theta}, \bm{\Theta}_a)^2-\vartheta{(\bm{\Theta},\bm{\Theta}_a)}}{2\eta} \right)
\end{align*}

Since Assumption~\ref{assumption:5.2} holds true in this setting, we have an $m>0$ such that for any $\bm{\theta} \in conv(\bm{\Theta})$
    \[\nabla^2 U(\bm{\theta}) \ge m\, I.\]
From this, one notes that 
{\small    
\begin{align*}
Z_{\gamma}(\widetilde{\bm{\bm{\theta}}}) &\geq \frac{1}{\sqrt{2\pi\alpha_a}^d} \exp\left\{-\frac{U(\bm{\theta})}{2}- \frac{\alpha_a}{8\eta^2} \| \bm{\theta} - \bm{\theta}_a \|^2 + \frac{1}{2\eta} \| \bm{\theta} - \bm{\theta}_a \|^2\right\} \exp\left\{-\frac{1}{2}\left(\frac{1}{ \alpha}- \frac{m}{2}\right)\, \text{diam($\bm{\Theta}$)}^2\right\}\\
&\sum_{x \in \bm{\Theta}} \exp\left(  \frac{U(x)}{2}\right) \int_{y} \sum_{x}  \exp \Bigg( -\frac{1}{2\alpha_a}\norm{y-\bm{\theta}_a}^2 - \frac{1}{2\eta} (\bm{\theta} - \bm{\theta}_a)^{\top} (x - y)\Bigg) \, dy\\
&\geq\sum_{x \in \bm{\Theta}} \exp\left(  \frac{U(x)}{2}\right) \exp\left\{-\frac{U(\bm{\theta})}{2}- \frac{\alpha_a}{8\eta^2} \| \bm{\theta} - \bm{\theta}_a \|^2-\frac{1}{2}\left(\frac{1}{ \alpha}- \frac{m}{2}\right)\, \text{diam($\bm{\Theta}$)}^2-\frac{1}{2\eta}\Delta(\bm{\Theta}, \bm{\Theta}_a)^2\right\}\\
&\geq\sum_{x \in \bm{\Theta}} \exp\left(  \frac{U(x)}{2}\right) \exp\left\{-\frac{U(\bm{\theta})}{2}- \frac{\alpha_a}{8\eta^2} \Delta(\bm{\Theta}, \bm{\Theta}_a)^2-\frac{1}{2}\left(\frac{1}{ \alpha}- \frac{m}{2}\right)\, \text{diam($\bm{\Theta}$)}^2-\frac{1}{2\eta}\Delta(\bm{\Theta}, \bm{\Theta}_a)^2 \right\}
\end{align*}
}
In other words,
{\scriptsize
    \begin{align*}
        \exp\left( (- \frac{\alpha_a}{8\eta^2}-\frac{1}{2\eta}) \Delta(\bm{\Theta}, \bm{\Theta}_a)^2-\frac{1}{2}\left(\frac{1}{ \alpha}- \frac{m}{2}\right)\, \text{diam($\bm{\Theta}$)}^2 \right) \le \frac{Z_{\gamma}(\widetilde{\bm{\bm{\theta}}})}{\sum_{x \in \bm{\Theta}} \exp\left(  \frac{U(x)}{2}\right) \exp\left(-  \frac{U(\bm{\theta})}{2}\right)} \le \exp\left( \frac{\Delta(\bm{\Theta}, \bm{\Theta}_a)^2-\vartheta{(\bm{\Theta},\bm{\Theta}_a)}}{2\eta}\right)
    \end{align*}
}
Consequently,
    \begin{align*}
        \frac{\frac{Z_{\gamma}(\widetilde{\bm{\bm{\theta}}})}{\sum_{x \in \bm{\Theta}} \exp\left(  \frac{U(x)}{2}\right) \exp\left(-  \frac{U(\bm{\theta})}{2}\right)}}{\frac{Z_{\gamma}(\widetilde{\bm{\theta}'})}{\sum_{x \in \bm{\Theta}} \exp\left(  \frac{U(x)}{2}\right) \exp\left(-  \frac{U(\bm{\theta}')}{2}\right)}} &\geq \frac{\exp\left((- \frac{\alpha_a}{8\eta^2} -\frac{1}{2\eta})\Delta(\bm{\Theta}, \bm{\Theta}_a)^2 -\frac{(2-m\alpha)\text{diam($\bm{\Theta}$)}^2}{4\alpha} \right)}{\exp\left( \frac{\Delta(\bm{\Theta}, \bm{\Theta}_a)^2-\vartheta{(\bm{\Theta},\bm{\Theta}_a)}}{2\eta} \right)}
    \end{align*}
This implies
\begin{align*}
        \frac{Z_{\gamma}(\widetilde{\bm{\bm{\theta}}})}{Z_{\gamma}(\widetilde{\bm{\theta}'})} &\geq {\exp \left( \frac{1}{2} (-U(\bm{\theta})+U(\bm{\theta}'))\right)}\frac{\exp\left( (-\frac{\alpha_a}{8\eta^2} -\frac{1}{2\eta})\Delta(\bm{\Theta}, \bm{\Theta}_a)^2 -\frac{(2-m\alpha)\text{diam($\bm{\Theta}$)}^2}{4\alpha} \right)}{\exp\left( \frac{\Delta(\bm{\Theta}, \bm{\Theta}_a)^2-\vartheta{(\bm{\Theta},\bm{\Theta}_a)}}{2\eta} \right)}
\end{align*}

One notices from \eqref{eq:proposal},
{\footnotesize
\begin{align*}
q_{\gamma}(\widetilde{\bm{\theta}'}|\widetilde{\bm{\bm{\theta}}}) 
&= \frac{Z_{\gamma}(\widetilde{\bm{\bm{\theta}}})^{-1}}{\sqrt{(2\pi\alpha_a)^d}} 
\exp \Bigg( 
    \frac{1}{2}( -U(\bm{\theta}) + U(\bm{\theta}') ) 
    - (\bm{\theta}-\bm{\theta}')^{\top}
    \left(\frac{1}{2\alpha}I + \frac{1}{4}\int_{0}^{1} \nabla^2 U((1-s)\bm{\theta}+s\bm{\theta}')ds \right)(\bm{\theta}-\bm{\theta}') \\
&\quad 
    - \frac{1}{2\eta}(\bm{\theta} - \bm{\theta}_a)^{\top}(\bm{\theta}' - \bm{\theta}_a') 
    - \frac{1}{2\alpha_a} \|\bm{\theta}_a' - \bm{\theta}_a\|^2 
    + \frac{4\eta-\alpha_a}{8\eta^2} \|\bm{\theta} - \bm{\theta}_a\|^2 
\Bigg)\\
&\geq 
\frac{Z_{\gamma}(\widetilde{\bm{\bm{\theta}}})^{-1}}{\sqrt{(2\pi\alpha_a)^d}} 
\exp \Bigg( 
    \frac{1}{2}\left\langle\nabla U(\bm{\theta}),\bm{\theta}'-\bm{\theta}\right\rangle 
    - \frac{1}{2\alpha}\|\bm{\theta}-\bm{\theta}'\|^2 - \frac{1}{2\eta}(\bm{\theta} - \bm{\theta}_a)^{\top}(\bm{\theta}' - \bm{\theta}_a') \\
& \quad
    - \frac{1}{2\alpha_a} \|\bm{\theta}_a' - \bm{\theta}_a\|^2 
    - \frac{\alpha_a}{8\eta^2} \|\bm{\theta} - \bm{\theta}_a\|^2 
\Bigg)
\end{align*}
}

We also note that
{\scriptsize
    \begin{align*}
        - \frac{1}{2} \left\langle\nabla U(\bm{\theta}),\bm{\theta}'-\bm{\theta}\right\rangle+\frac{1}{2\alpha}\|\bm{\theta}-\bm{\theta}'\|^2
        & = \frac{1}{2} \left\langle - \nabla U(\bm{\theta}) + \nabla U(a),\bm{\theta}'-\bm{\theta}\right\rangle+ \frac{1}{2} \left\langle - \nabla U(a),\bm{\theta}'-\bm{\theta}\right\rangle + \frac{1}{2\alpha}\|\bm{\theta}-\bm{\theta}'\|^2\\
        & \le \frac{1}{2} \left\langle - \nabla U(\bm{\theta}) + \nabla U(a),\bm{\theta}'-\bm{\theta}\right\rangle+ \frac{1}{2} \left\langle - \nabla U(a),\bm{\theta}'-\bm{\theta}\right\rangle + \frac{1}{2\alpha}diam(\bm{\Theta})^2 \\
        & \le \frac{1}{2} \left\| - \nabla U(\bm{\theta}) + \nabla U(a)\| \| \bm{\theta}'-\bm{\theta}\right\|+ \frac{1}{2} \left\|\nabla U(a) \| \| \bm{\theta}'-\bm{\theta}\right\| + \frac{1}{2\alpha}diam(\bm{\Theta})^2 \\
        & \le \frac{1}{2} \| - \nabla U(\bm{\theta}) + \nabla U(a)\|   diam(\bm{\Theta}) + \frac{1}{2} \|\nabla U(a) \| diam(\bm{\Theta}) + \frac{1}{2\alpha}diam(\bm{\Theta})^2 \\
        & \le \left(\frac{1}{2} M+\frac{1}{2\alpha}\right)\, diam(\bm{\Theta})^2+ \frac{1}{2} \|\nabla U(a)\|\, diam(\bm{\Theta}).
    \end{align*}
}
This is because,
From Assumption \ref{assumption:5.1} (U is $M$-gradient Lipschitz), we have 
\begin{align*}
  \frac{1}{2} \int_{0}^{1} \nabla^{2} U ( (1-s) \theta + s \theta' ) \, ds ) (\bm{\theta} - \bm{\theta}')  + \frac{1}{\alpha} I  \ge  \left( \frac{1}{ \alpha}  -  \frac{M}{2} \right) I \\
\end{align*}
Since $\alpha < \frac{2}{M}$, the matrix $\left(\frac{1}{2 \alpha} - \frac{ M}{2}\right) I $    is positive definite.\\

Combining, we get
{\scriptsize
\begin{align*}
q_{\gamma}(\widetilde{\bm{\theta}'}|\widetilde{\bm{\bm{\theta}}})
&\geq
\frac{Z_{\gamma}(\widetilde{\bm{\bm{\theta}}})^{-1}}{\sqrt{(2\pi\alpha_a)^d}}\exp  \left\{ (-\frac{M}{2}-\frac{1}{2\alpha}) \text{diam($\bm{\Theta}$)}^2- \frac{1}{2} \|\nabla U(a)\|\text{diam($\bm{\Theta}$)} - \frac{1}{2\eta}(\bm{\theta} - \bm{\theta}_a)^{\top}(\bm{\theta}' - \bm{\theta}_a') - \frac{1}{2\alpha_a} \|\bm{\theta}_a' - \bm{\theta}_a\|^2 - \frac{\alpha_a}{8\eta^2} \|\bm{\theta} - \bm{\theta}_a\|^2\right \}\\
&\geq \frac{ \frac{1}{\sqrt{(2\pi\alpha_a)^d}}\exp  \left\{ (-\frac{M}{2}-\frac{1}{2\alpha}) \text{diam($\bm{\Theta}$)}^2- \frac{1}{2} \|\nabla U(a)\|\text{diam($\bm{\Theta}$)} - \frac{1}{2\eta}(\bm{\theta} - \bm{\theta}_a)^{\top}(\bm{\theta}' - \bm{\theta}_a') - \frac{1}{2\alpha_a} \|\bm{\theta}_a' - \bm{\theta}_a\|^2 - \frac{\alpha_a}{8\eta^2} \|\bm{\theta} - \bm{\theta}_a\|^2\right \}}{\sum_{x \in \bm{\Theta}} \exp\left(  \frac{U(x)}{2}\right) \exp\left(-  \frac{U(\bm{\theta})}{2} +\frac{\Delta(\bm{\Theta}, \bm{\Theta}_a)^2-\vartheta{(\bm{\Theta},\bm{\Theta}_a)}}{2\eta} \right)}\\
&\geq \frac{\exp \left\{ -\frac{1}{2\alpha_a} \text{diam}(\bm{\Theta}_a)^2 \right\}}{\Phi_{\alpha_a}(\bm{\bm{\Theta}}_a)}\frac{ \exp  \left\{(-\frac{M}{2}-\frac{1}{2\alpha}) \text{diam($\bm{\Theta}$)}^2- \frac{1}{2} \|\nabla U(a)\|\text{diam($\bm{\Theta}$)}   + (-\frac{1}{2\eta}-\frac{\alpha_a}{8\eta^2}) \Delta( \bm{\Theta}, \bm{\Theta}_a)^2\right \}}{\sum_{x \in \bm{\Theta}} \exp\left(  \frac{U(x)}{2}\right) \exp\left(-  \frac{U(\bm{\theta})}{2}  +\frac{\Delta(\bm{\Theta}, \bm{\Theta}_a)^2-\vartheta{(\bm{\Theta},\bm{\Theta}_a)}}{2\eta} \right)}\\
\end{align*}
}

 Acceptance Ratio,
 \begin{align*}
 \mathcal{\rho}(\widetilde{\bm{\theta}'}\mid \widetilde{\bm{\theta}})&= \left(\frac{\pi(\widetilde{\bm{\theta}'})q_{\gamma}(\widetilde{\bm{\theta}}\mid \widetilde{\bm{\theta}'})}{\pi(\widetilde{\bm{\theta}})q_{\gamma}(\widetilde{\bm{\theta}'}\mid \widetilde{\bm{\theta}})}\right)\\
 &= 
  \exp\left\{ U({\bm{\theta}'})- U(\bm{\theta}) +\frac{1}{2\eta}(\norm{\bm{\theta}-\bm{\theta}_a}^2-\norm{\bm{\theta}'-\bm{\theta}_a'}^2)\right\}\frac{\tilde{Z}}{\tilde{Z}}\cdot\\
  &\exp\left\{ U({\bm{\theta}})- U(\bm{\theta}')  -\frac{1}{2\eta}(\norm{\bm{\theta}-\bm{\theta}_a}^2-\norm{\bm{\theta}'-\bm{\theta}_a'}^2) - \frac{\alpha_a}{8\eta^2} (\norm{\bm{\theta}' - \bm{\theta}_a'}^2-\norm{\bm{\theta} - \bm{\theta}_a}^2 )\right \}\frac{Z_{\gamma}(\widetilde{\bm{\bm{\theta}}})}{Z_{\gamma}(\widetilde{\bm{\theta}'})}\\
 &= 
  \exp\left\{ - \frac{\alpha_a}{8\eta^2} (\norm{\bm{\theta}' - \bm{\theta}_a'}^2-\norm{\bm{\theta} - \bm{\theta}_a}^2 )\right \}\frac{Z_{\gamma}(\widetilde{\bm{\bm{\theta}}})}{Z_{\gamma}(\widetilde{\bm{\theta}'})}
\end{align*}
where $\tilde{Z}$ is the normalizing constant for $\pi(\widetilde{\bm{\theta}})$.

with Acceptance Probability
\begin{align*}
 \mathcal{A}(\widetilde{\bm{\theta}'}\mid \widetilde{\bm{\bm{\theta}}})&= \left(\mathcal{\rho}(\widetilde{\bm{\theta}'}\mid \widetilde{\bm{\bm{\theta}}})\wedge 1\right)\\
 \end{align*}
and consider the transition kernel as 
\begin{align*}
p(\widetilde{\bm{\theta}'}\mid \widetilde{\bm{\bm{\theta}}})&=\left(\mathcal{A}(\widetilde{\bm{\theta}'}\mid \widetilde{\bm{\bm{\theta}}})\right)q_{\gamma}(\widetilde{\bm{\theta}'}\mid \widetilde{\bm{\bm{\theta}}})+\left(1-L(\widetilde{\bm{\bm{\theta}}})\right)\delta_{\widetilde{\bm{\bm{\theta}}}}(\widetilde{\bm{\theta}'})
\end{align*}
where $\delta_{\widetilde{\bm{\theta}}}(\widetilde{\bm{\theta}'})$ is the Kronecker delta function and $L(\widetilde{\bm{\theta}})$ is the total acceptance probability from the point $\widetilde{\bm{\theta}}$ with 
\[L(\widetilde{\bm{\theta}})=\int_{ \bm{\theta}_a'\in \bm{\Theta}_a}\sum_{ \bm{\theta}'\in \bm{\Theta} } \left(\mathcal{\rho}([\bm{\theta'}^T, \bm{\theta}_a'^T]^T \mid \widetilde{\bm{\bm{\theta}}})\wedge 1\right) q_{\gamma}([\bm{\theta'}^T, \bm{\theta}_a'^T]^T\vert \widetilde{\bm{\bm{\theta}}}) \quad d\bm{\theta}_a'\]  

\label{proof:lmj}
We note that
{\small
\begin{align*}
    p(\widetilde{\bm{\theta}'} \mid \widetilde{\bm{\bm{\theta}}}) 
    &= \left(\mathcal{A}(\widetilde{\bm{\theta}'} \mid \widetilde{\bm{\bm{\theta}}})\right) q_{\gamma}(\widetilde{\bm{\theta}'} \mid \widetilde{\bm{\bm{\theta}}}) + \left(1 - L(\widetilde{\bm{\bm{\theta}}})\right) \delta_{\widetilde{\bm{\bm{\theta}}}}(\widetilde{\bm{\theta}'}) \\
    &\geq \left(\mathcal{A}(\widetilde{\bm{\theta}'} \mid \widetilde{\bm{\bm{\theta}}})\right) q_{\gamma}(\widetilde{\bm{\theta}'} \mid \widetilde{\bm{\bm{\theta}}}) \\ 
    &= \left(\rho(\widetilde{\bm{\theta}'} \mid \widetilde{\bm{\bm{\theta}}}) \wedge 1\right) q_{\gamma}(\widetilde{\bm{\theta}'} \mid \widetilde{\bm{\bm{\theta}}}) \\
    &= \exp\left\{ - \frac{\alpha_a}{8\eta^2} \left(\|\bm{\theta}' - \bm{\theta}_a'\|^2 - \|\bm{\theta} - \bm{\theta}_a\|^2\right) \right\} \frac{Z_{\gamma}(\widetilde{\bm{\bm{\theta}}})}{Z_{\gamma}(\widetilde{\bm{\theta}'})} q_{\gamma}(\widetilde{\bm{\theta}'} \mid \widetilde{\bm{\bm{\theta}}}) \\
    &\geq \exp\left\{ - \frac{\alpha_a}{8\eta^2} \|\bm{\theta}' - \bm{\theta}_a'\|^2 \right\} \frac{Z_{\gamma}(\widetilde{\bm{\bm{\theta}}})}{Z_{\gamma}(\widetilde{\bm{\theta}'})} q_{\gamma}(\widetilde{\bm{\theta}'} \mid \widetilde{\bm{\bm{\theta}}}) \\
    &\geq \exp\left\{ - \frac{\alpha_a}{8\eta^2} \Delta(\bm{\Theta}, \bm{\Theta}_a)^2 + \frac{1}{2} (-U(\bm{\theta}) + U(\bm{\theta}')) \right\}\frac{\exp\left( -\frac{\alpha_a}{8\eta^2} - \frac{1}{2\eta})\Delta(\bm{\Theta}, \bm{\Theta}_a)^2 - \frac{(2 - m\alpha)\text{diam}(\bm{\Theta})^2}{4\alpha} \right)}{\exp\left( \frac{\Delta(\bm{\Theta}, \bm{\Theta}_a)^2 - \vartheta{(\bm{\Theta},\bm{\Theta}_a)}}{2\eta} \right)} q_{\gamma}(\widetilde{\bm{\theta}'} \mid \widetilde{\bm{\bm{\theta}}}) \\
    &\geq \exp\left\{ - \frac{\alpha_a}{8\eta^2} \Delta(\bm{\Theta}, \bm{\Theta}_a)^2 + \frac{1}{2} (-U(\bm{\theta}) + U(\bm{\theta}')) \right\} \frac{\exp\left( -\frac{\alpha_a}{8\eta^2} - \frac{1}{2\eta})\Delta(\bm{\Theta}, \bm{\Theta}_a)^2 - \frac{(2 - m\alpha)\text{diam}(\bm{\Theta})^2}{4\alpha} \right)}{\exp\left( \frac{\Delta(\bm{\Theta}, \bm{\Theta}_a)^2 - \vartheta{(\bm{\Theta},\bm{\Theta}_a)}}{2\eta} \right)} \\
    &\quad \cdot \frac{\exp \left\{ -\frac{1}{2\alpha_a} \text{diam}(\bm{\Theta}_a)^2 \right\}}{\Phi_{\alpha_a}(\bm{\bm{\Theta}}_a)} \frac{\exp \left\{( -\frac{M}{2} - \frac{1}{2\alpha}) \text{diam}(\bm{\Theta})^2 - \frac{1}{2} \|\nabla U(a)\| \text{diam}(\bm{\Theta}) + \left(-\frac{1}{2\eta} - \frac{\alpha_a}{8\eta^2}\right) \Delta(\bm{\Theta}, \bm{\Theta}_a)^2 \right\}}{\sum_{x \in \bm{\Theta}} \exp\left( \frac{U(x)}{2}\right) \exp\left(- \frac{U(\bm{\theta})}{2} + \frac{\Delta(\bm{\Theta}, \bm{\Theta}_a)^2 - \vartheta{(\bm{\Theta},\bm{\Theta}_a)}}{2\eta} \right)}\\
    &= \frac{\exp \left\{ -\frac{1}{2\alpha_a} \text{diam}(\bm{\Theta}_a)^2 \right\}}{\Phi_{\alpha_a}(\bm{\bm{\Theta}}_a)} \frac{\exp \left\{ \frac{1}{2} U(\bm{\theta}')\right\} }{\sum_{x \in \bm{\Theta}} \exp\left( \frac{U(x)}{2}\right) } \exp\left\{ (- \frac{3\alpha_a}{8\eta^2}- \frac{2}{\eta})\Delta(\bm{\Theta}, \bm{\Theta}_a)^2 +\frac{\vartheta(\bm{\Theta}, \bm{\Theta}_a)}{\eta}\right\}\\
    &\quad \cdot \exp \left\{ ( -\frac{M}{2} - \frac{1}{\alpha}+\frac{m}{4}) \text{diam}(\bm{\Theta})^2 - \frac{1}{2} \|\nabla U(a)\| \text{diam}(\bm{\Theta})   \right\}\\
    &= \epsilon_{ \gamma} \frac{\exp \left\{ \frac{1}{2} U(\bm{\theta}')\right\} }{\sum_{x \in \bm{\Theta}} \exp\left( \frac{U(x)}{2}\right) } \frac{\exp \left\{ -\frac{1}{2\alpha_a} \text{diam}(\bm{\Theta}_a)^2 \right\}}{\Phi_{\alpha_a}(\bm{\bm{\Theta}}_a)}
\end{align*} 
}
\emph{Proof.} Proof follows from using Lemma $\ref{lemma:minorization_bounded_edmala}$ .

\section{Additional Experimental Results \label{sec:app_add}}

\subsection{4D Joint Bernoulli}\label{subsec:app-bern}

To provide additional insights into the functionality of EDLP samplers, we explore their behavior on the 4D Joint Bernoulli Distribution, which serves as the simplest low-dimensional case among our experiments. This aids in visualizing and understanding the sampling process.

\subsubsection*{Target Distribution}
The following represents the probability mass function (PMF) for the 4D Joint Bernoulli Distribution used in our test case. The distribution has 16 states with the corresponding probabilities:

\begin{figure}[h]
    \centering
    \begin{minipage}{0.4\textwidth}
\[
P_{\bm{\Theta}}(\bm{\theta}) =
\begin{cases} 
0.07688 & \text{if } \bm{\theta} = 0000, \\
0.04725 & \text{if } \bm{\theta} = 0001, \\
0.12500 & \text{if } \bm{\theta} = 0010, \\
0.01667 & \text{if } \bm{\theta} = 0011, \\
0.08688 & \text{if } \bm{\theta} = 0100, \\
0.07688 & \text{if } \bm{\theta} = 0101, \\
0.07688 & \text{if } \bm{\theta} = 0110, \\
0.16756 & \text{if } \bm{\theta} = 0111, \\
0.04725 & \text{if } \bm{\theta} = 1000, \\
0.05825 & \text{if } \bm{\theta} = 1001, \\
0.01667 & \text{if } \bm{\theta} = 1010, \\
0.04725 & \text{if } \bm{\theta} = 1011, \\
0.07688 & \text{if } \bm{\theta} = 1100, \\
0.04725 & \text{if } \bm{\theta} = 1101, \\
0.01900 & \text{if } \bm{\theta} = 1110, \\
0.01335 & \text{if } \bm{\theta} = 1111. \\
\end{cases}
\]
    \end{minipage}%
    \begin{minipage}{0.59\textwidth }
        \centering
        \includegraphics[width=\linewidth]{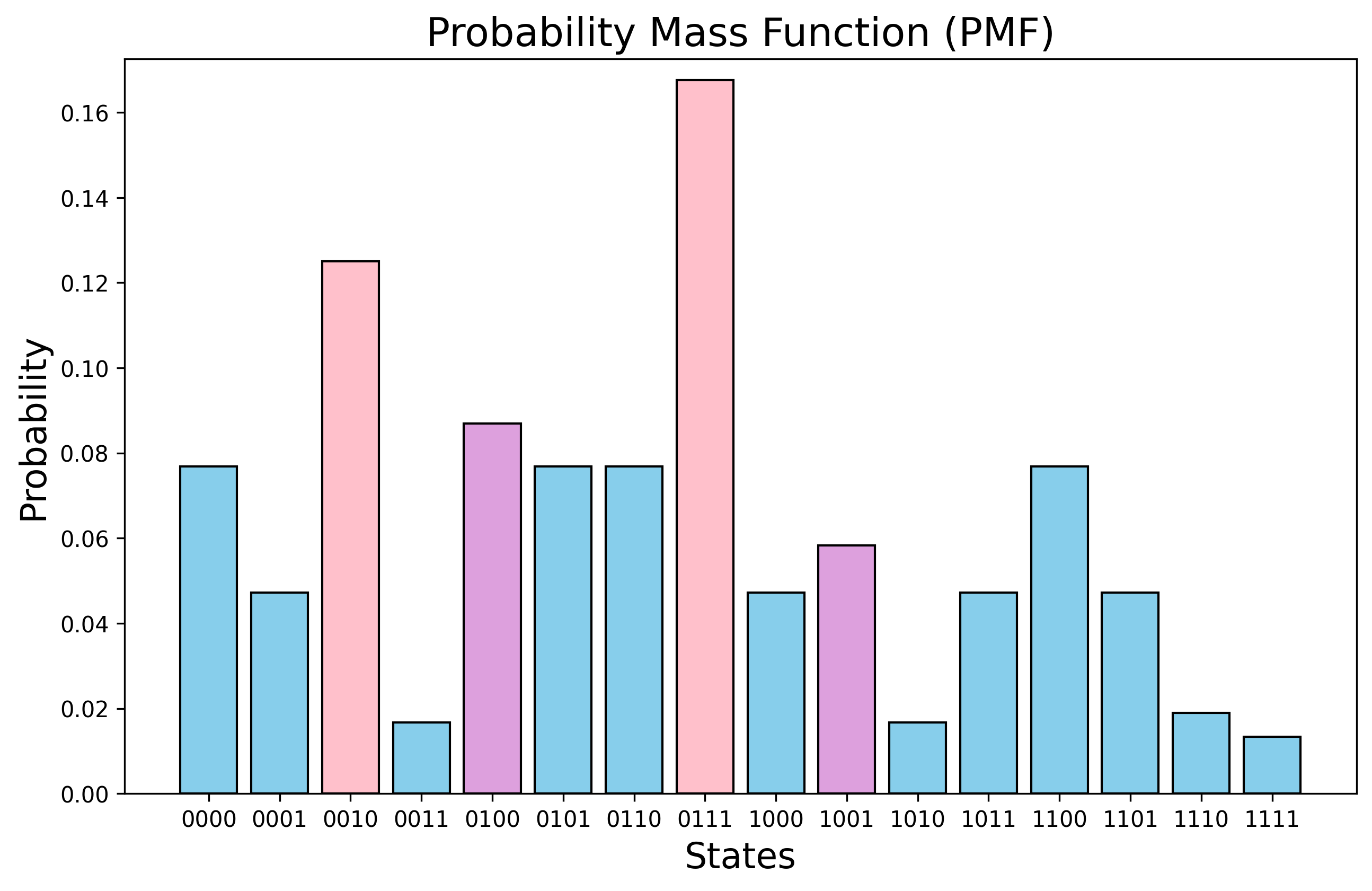}
        \caption{Target Distribution for 4D Joint Bernoulli}
        \label{fig:bern_target}
    \end{minipage}
\end{figure}
\subsubsection*{Flatness Diagnostics}

\begin{figure}[t]
  \begin{center}
  \begin{tabular}{cc}
    \hspace{-10pt}
    \includegraphics[width=.45\textwidth]{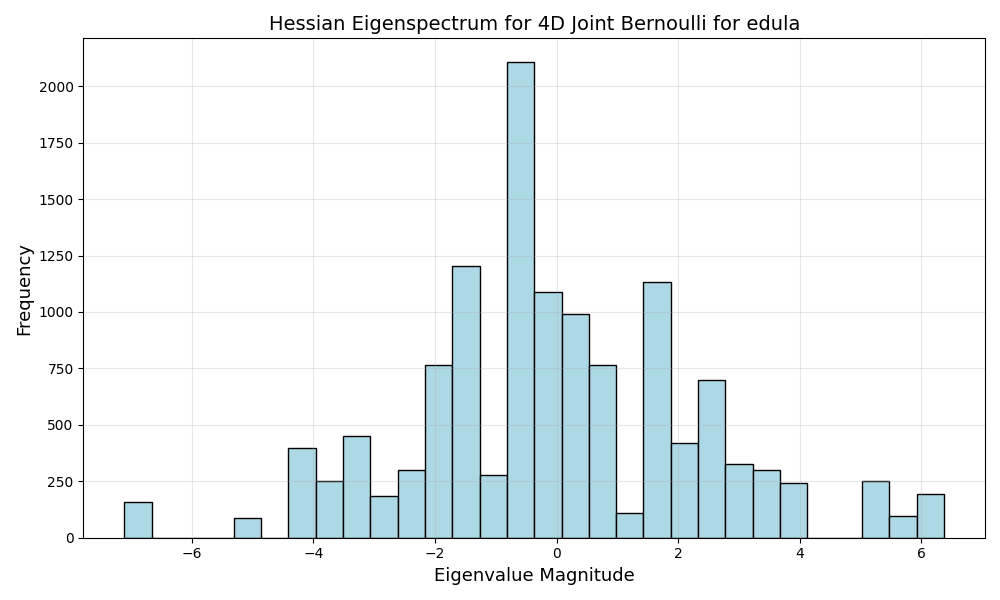} &
    \hspace{-10pt}
    \includegraphics[width=.45\textwidth]{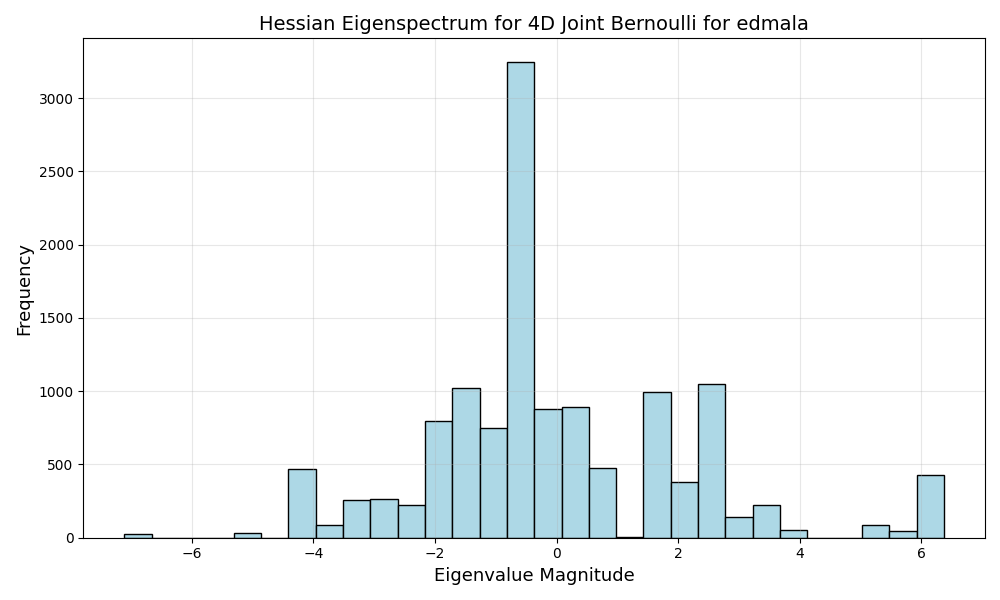} \\
    \hspace{-10pt}
    \includegraphics[width=.45\textwidth]{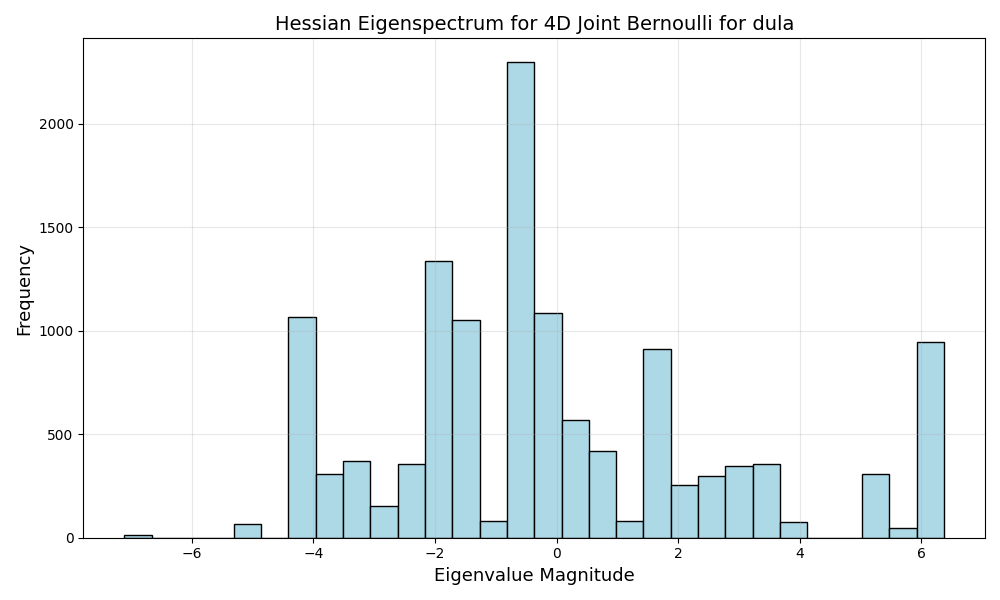} &
    \hspace{-10pt}
    \includegraphics[width=.45\textwidth]{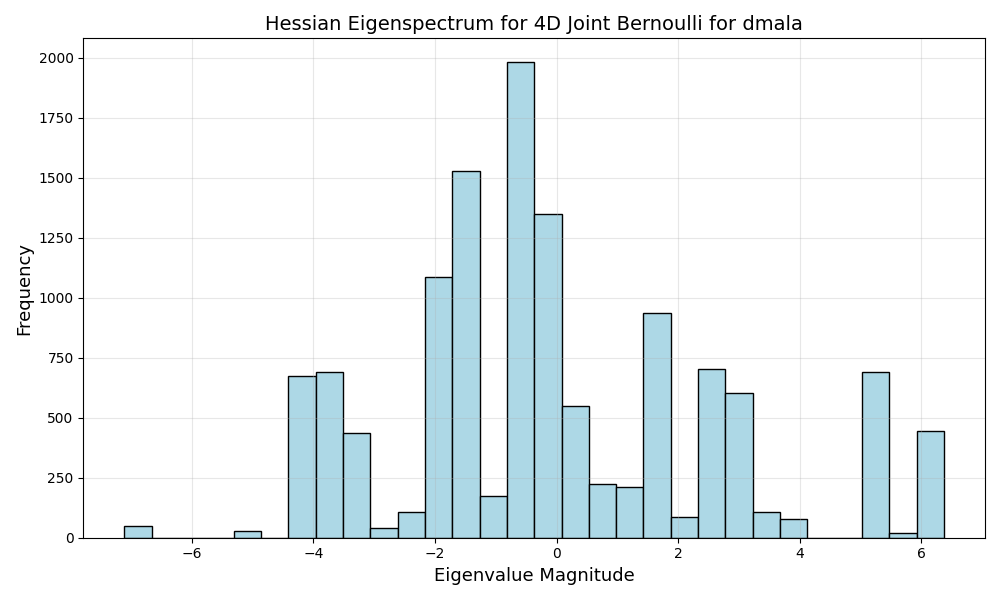}
  \end{tabular}
  \end{center}
  \vspace{-10pt}
  \caption{Eigenspectra of EDULA, EDMALA, DULA, and DMALA's  performance on a Bernoulli distribution.}
  \label{fig:eigenspectra}
  \vspace{-10pt}
\end{figure}

Under the experimental setup outlined in Section \ref{sec:exp}, we present the true Eigenspectrum of the Hessian, derived from the discrete samples collected for EDULA, EDMALA, DULA, and DMALA (Figure \ref{fig:eigenspectra}).We manually
tune the stepsizes for EDULA and EDMALA to 0.1 and 0.4 respectively. This visualization is inspired by Section 6.3 of \citep{li2024entropymcmc}, where diagonal Fisher information matrix approximation was used to plot the Eigenvalues. The alignment of the Eigenvalues closer to 0 indicates that the sampled data corresponds to a flatter curvature of the energy function.

EDMALA and EDULA, specifically designed with entropy-aware flatness optimization, exhibit eigenvalue distributions that are notably tighter and more concentrated around zero compared to their non-entropic counterparts, DMALA and DULA.

Quantitatively, EDULA demonstrates a lower spectral dispersion, evidenced by a lower standard deviation (std = 2.401) and narrower interquartile range (IQR = 3.031), relative to DULA (std = 2.832, IQR = 3.466). Similarly, EDMALA outperforms DMALA in terms of spectral concentration, achieving a standard deviation of 2.197 and IQR of 2.747, compared to DMALA's standard deviation of 2.700 and IQR of 3.224. Furthermore, visual inspection corroborates these quantitative findings; EDMALA and EDULA feature fewer extreme eigenvalues and outliers, reflecting biasing into sampling from flatter regions. Collectively, these results affirm that our entropy-guided methods (EDMALA, EDULA)  effectively traverse flatter, aligning well with their intended design objectives.

\subsection{TSP}\label{subsec:app-tsp}

\begin{figure}[H]
\centering
\centering
    \includegraphics[width=0.5\textwidth]{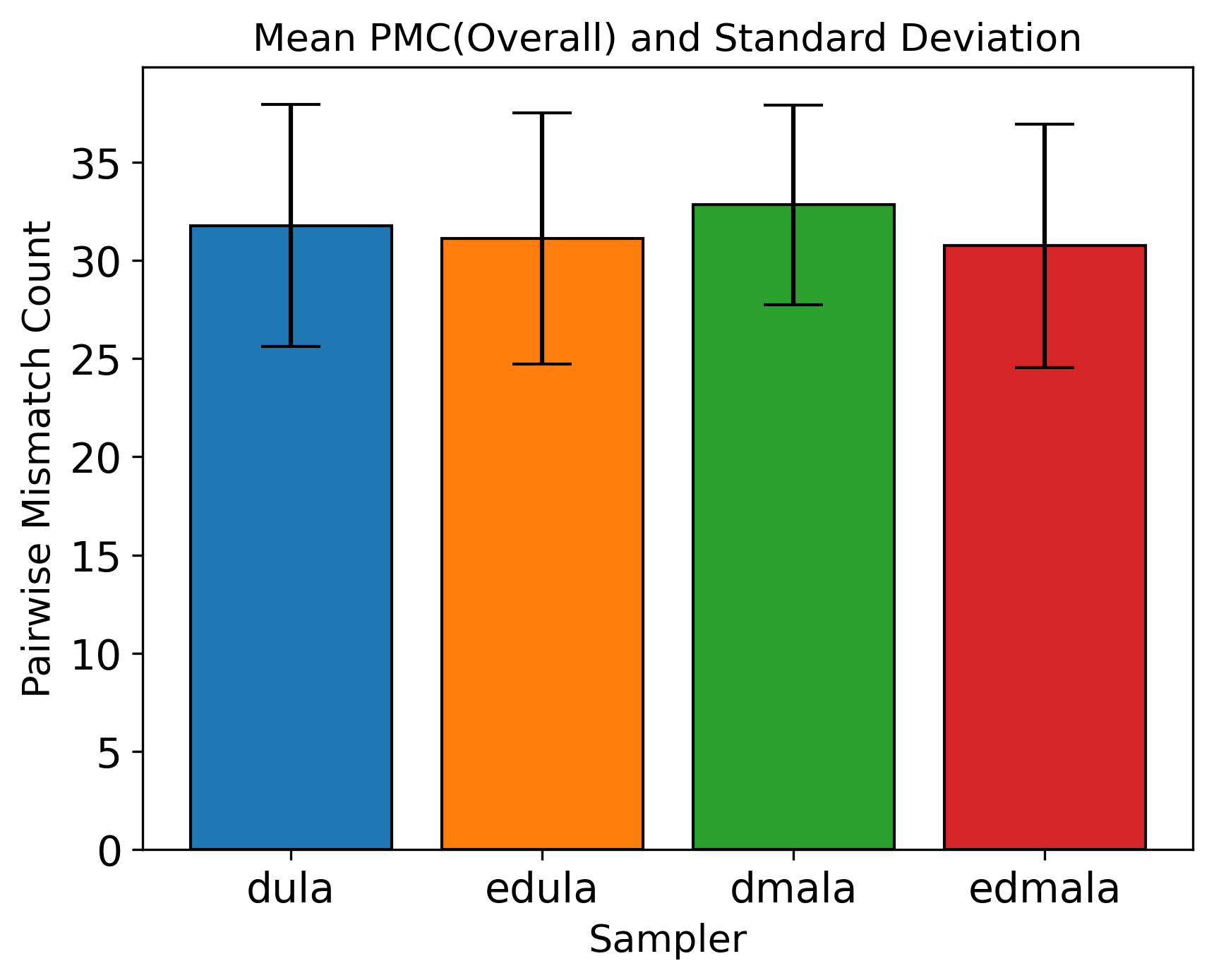}
    \caption{Variation in Solutions}
    \label{fig:tsp_var_sol}
\end{figure}

Figure \ref{fig:tsp_var_sol} presents the average PMC between solutions generated by each sampler, along with their standard deviations. DULA and EDULA exhibit nearly identical mean swap distances, whereas EDMALA demonstrates a notably lower mean swap distance compared to DMALA. This suggests that the solutions proposed by EDMALA are structurally more similar, indicating a higher degree of consistency across its sampled solutions. 

\begin{figure}[H]
\centering
\centering
    \includegraphics[width=0.6\textwidth]{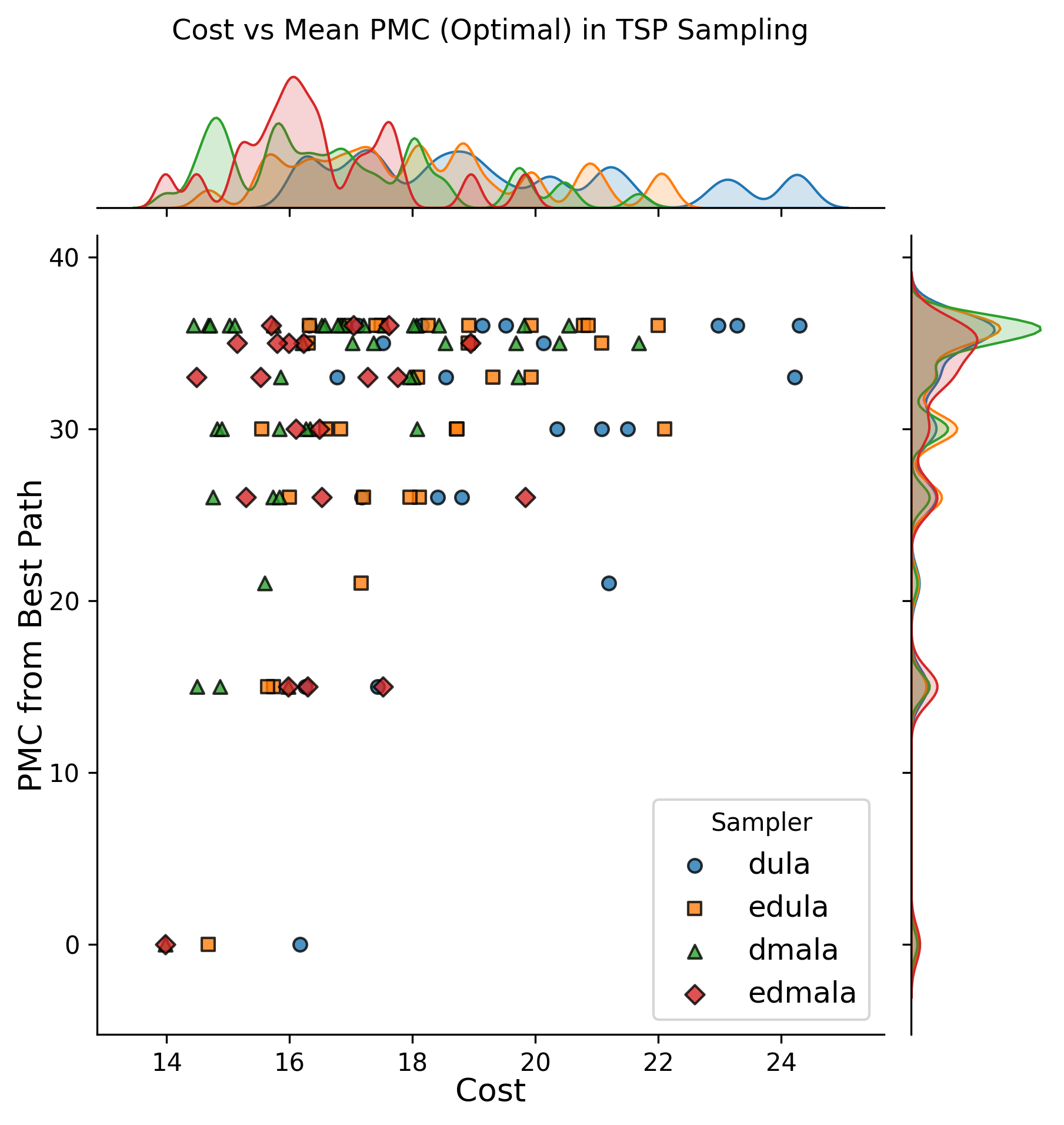}
    \caption{Marginal Plot}
    \label{fig:tsp_cost_swapd}
\end{figure}

Figure \ref{fig:tsp_cost_swapd} showcases the performance characteristics of different samplers in terms of cost and solution diversity for the TSP. EDMALA and EDULA exhibit a narrower cost distribution, suggesting that they consistently identify solutions within a tighter range of costs. This stability implies a focused exploration within a particular solution quality band \cite{camm1997constrained}. In contrast, DMALA and DULA have a broader cost spread, indicating more variability in the quality of solutions they find. 

When examining diversity in relation to the best solution, both DULA and DMALA maintain a similar spread, signifying comparable exploration depths relative to optimality. However, EDMALA stands out with a significantly smaller diversity spread compared to DMALA, indicating that EDMALA tends to produce solutions that are closer to the optimal path. This characteristic suggests that EDMALA is better suited for tasks requiring proximity to optimal solutions.

\subsection{RBM}\label{subsec:app-rbm}
\subsection*{Mode Analysis}
We performed mode analysis to validate the diversity and quality of MNIST digit samples generated by various samplers. Mode analysis assesses whether each sampler can capture the full range of MNIST digit classes (0-9) without falling into \textit{mode collapse}, a phenomenon where a generative model fails to represent certain data modes, thus limiting diversity. We leveraged a \textit{LeNet-5 convolutional neural network} \cite{lecun1998gradient} trained on MNIST to classify each generated sample and produce a class distribution for each sampler. The choice of LeNet-5, a reliable architecture for digit recognition, ensures accurate class predictions, thus providing a robust method to assess the representativeness of the samples. We train the model for 10 epochs, and achieve a $98.85\%$ accuracy on test data.

\begin{figure}[H]
\centering
    \begin{tabular}{ccc}       
        \includegraphics[width=0.3\textwidth]{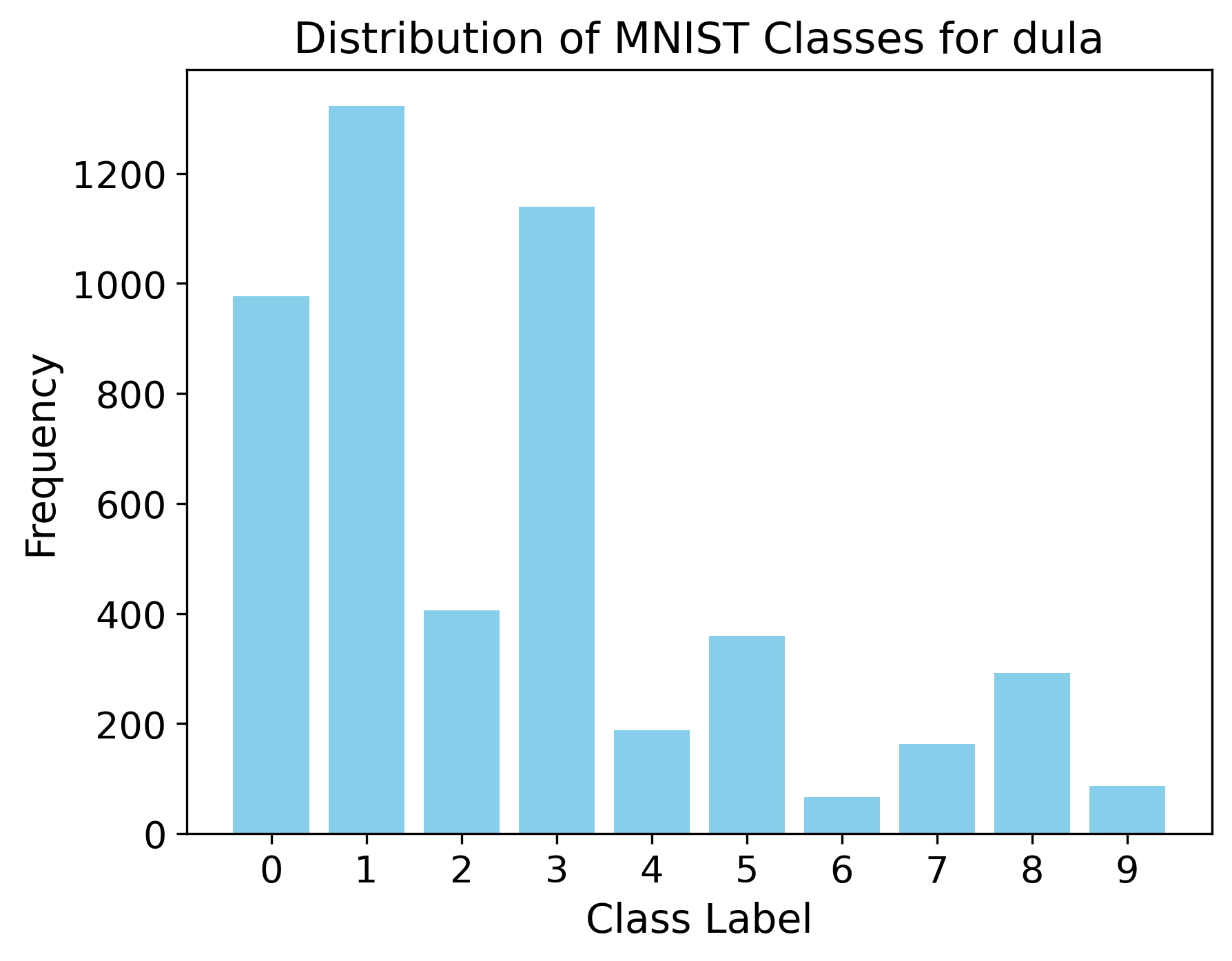}  &
        \includegraphics[width=0.3\textwidth]{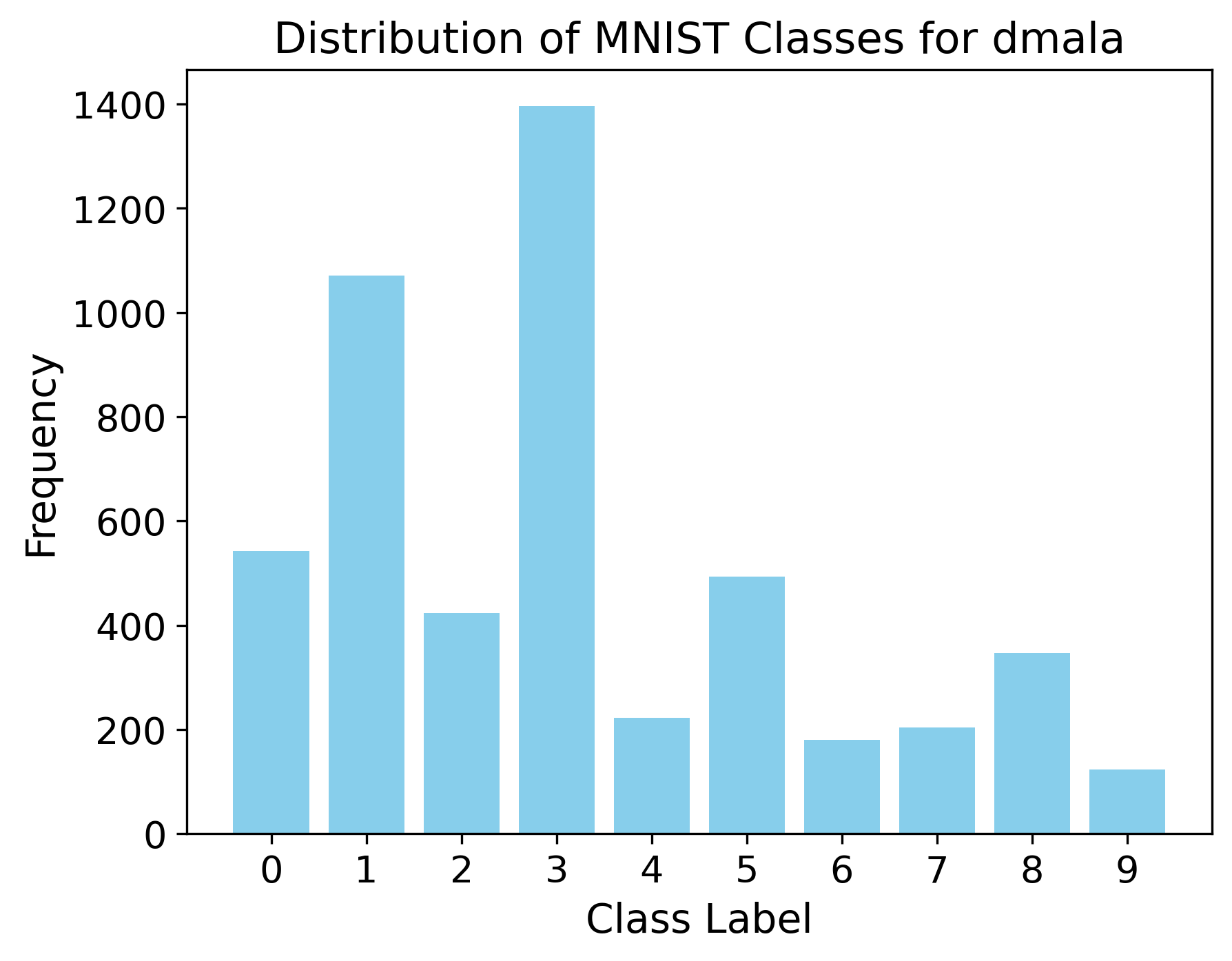}  &
        \includegraphics[width=0.3\textwidth]{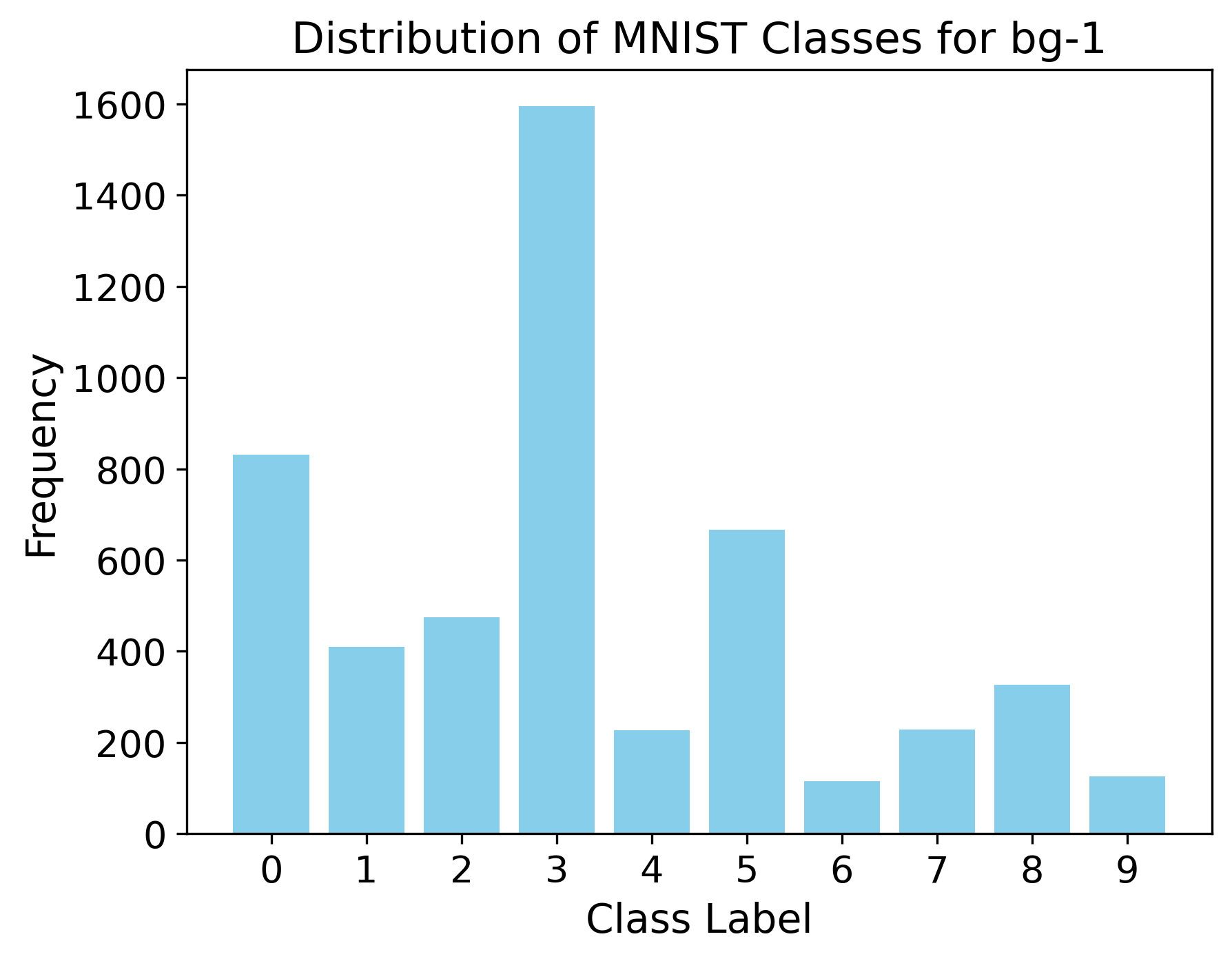}
        \\
        (a) DULA &
        (b) DMALA&
        (c) BG-1
        \\
        \includegraphics[width=0.3\textwidth]{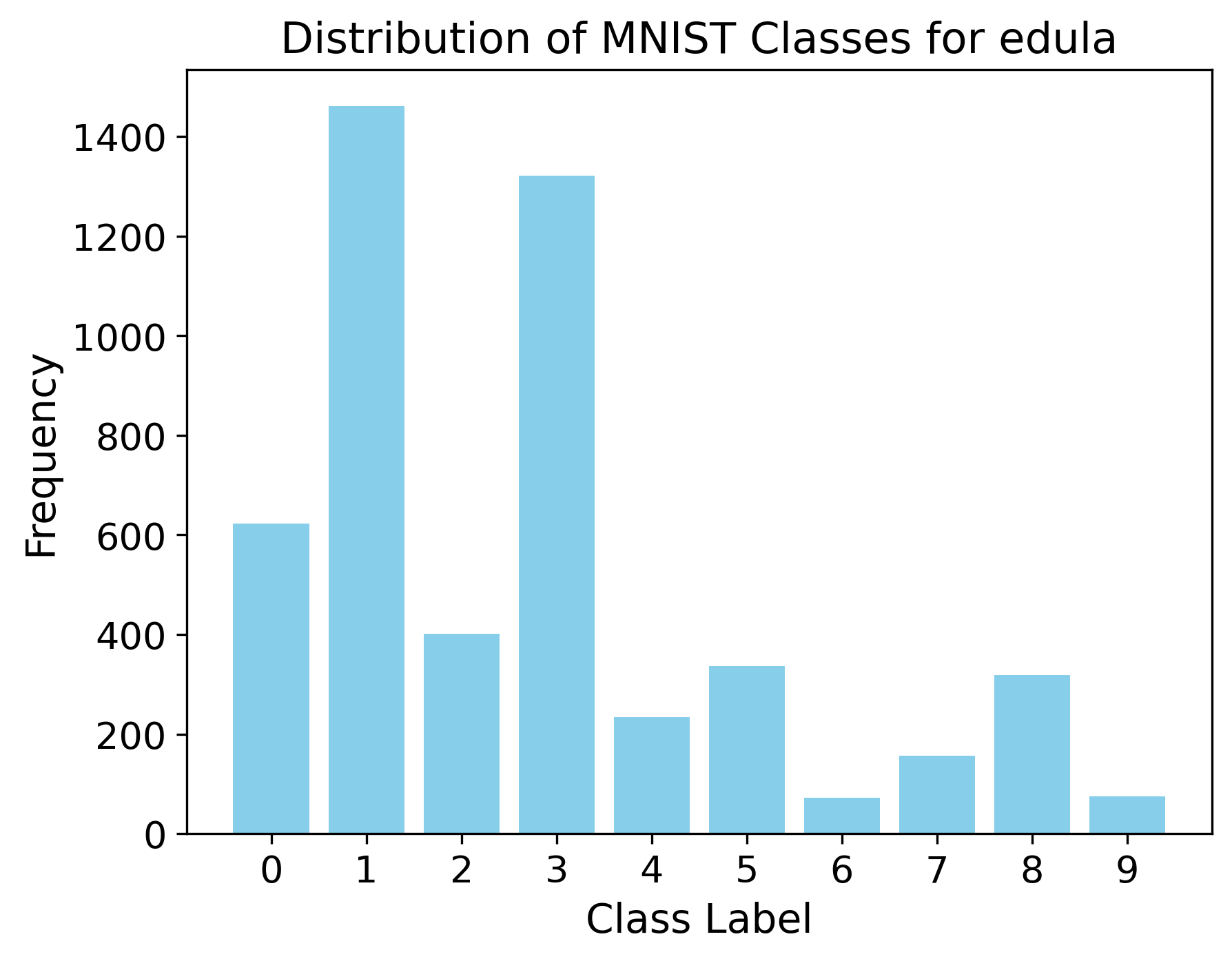}& 
        \includegraphics[width=0.3\textwidth]{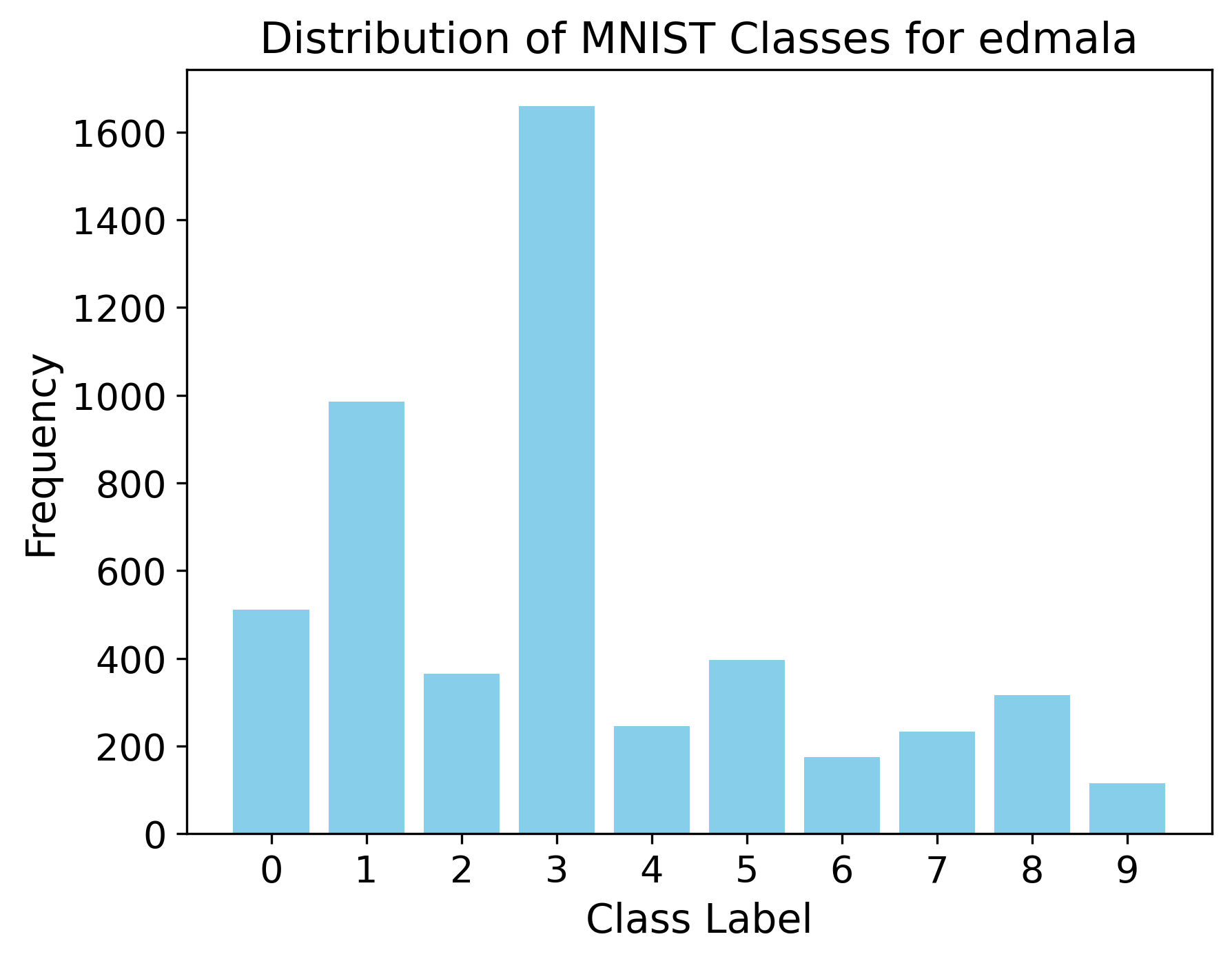}&
        \includegraphics[width=0.3\textwidth]{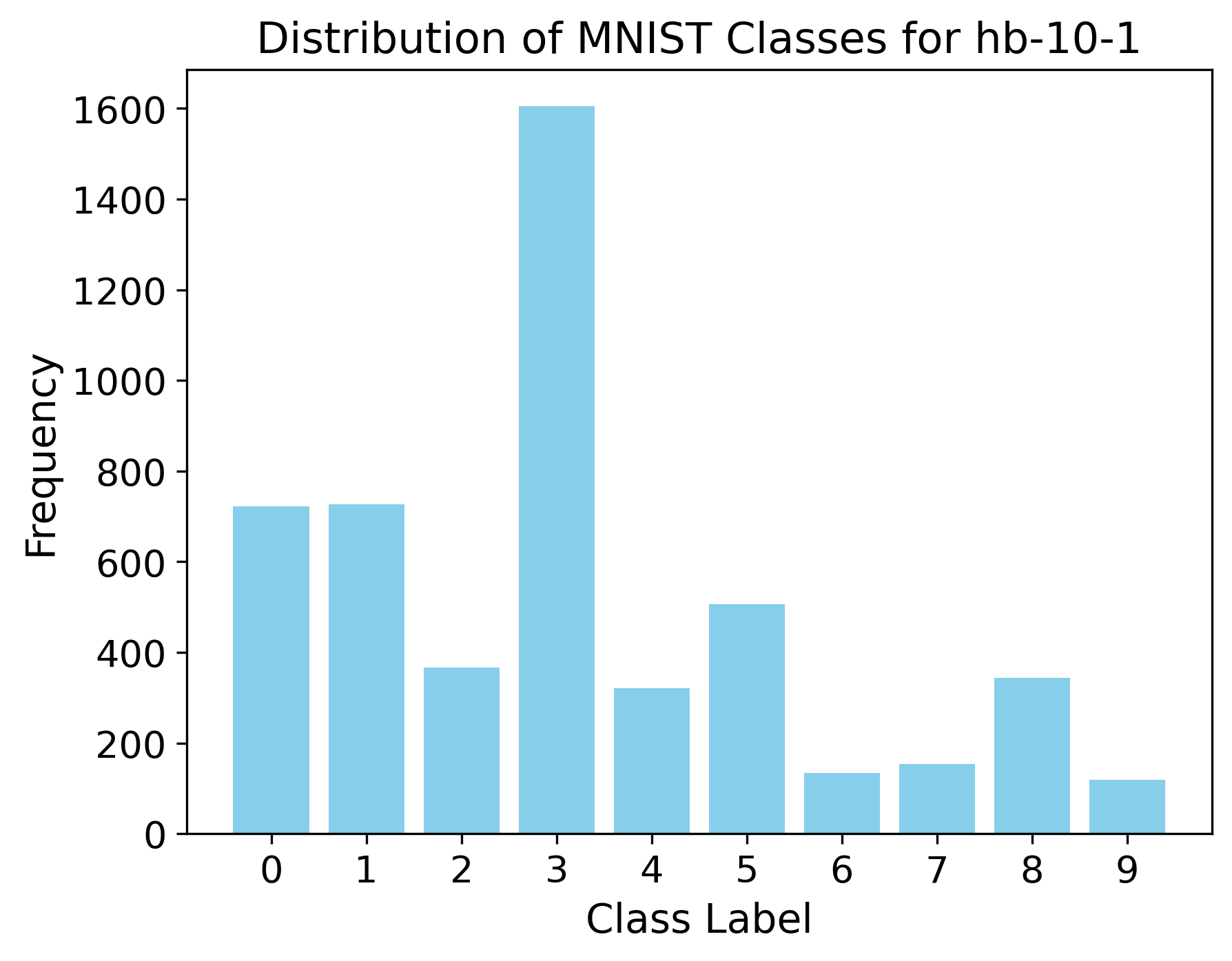}
        \\
        (d) EDULA&
        (e) EDMALA&
        (f) HB-10-1
    \end{tabular}
    \caption{Mode Analysis}
    \label{fig:mode_analysis}
\end{figure}

The results( Figure \ref{fig:mode_analysis}) from our analysis indicated that all samplers produced samples across all digit classes, showing no evidence of mode collapse. Although certain samplers exhibited a preference for specific classes these biases did not reach the level of complete mode omission. Each class was represented in the generated samples, confirming that the samplers achieved an acceptable level of \textit{mode diversity}. By confirming that all classes are covered, we demonstrate that each sampler can adequately approximate the diversity of the MNIST dataset, assuring the samples' representativeness \cite{salimans2016improved, goodfellow2014generative}.

\subsection{BBNN}\label{subsec:app-bbnn}

We train 50 Binary Bayesian Neural Networks in parallel as in Section \ref{sec:exp} and report the Average Training Log-Likelihood for our experiments in Table \ref{tab:tll}. Across all datasets, the EDLP samplers consistently outperform other samplers, demonstrating their ability to maintain or improve log-likelihood values. Importantly, when EDLP does not yield a substantial improvement, it still manages to avoid significantly impacting the training log-likelihood negatively.

\begin{table}[h!]
    \centering
    \renewcommand{\arraystretch}{1.2}
    \caption{Average Training Log-Likelihood}
    \resizebox{\textwidth}{!}{%
    \begin{tabular}{|l|c|c|c|c|c|c|}
        \hline
        Dataset & \textbf{Gibbs} & \textbf{GWG} & \textbf{DULA} & \textbf{DMALA} & \textbf{EDULA} & \textbf{EDMALA} \\
        \hline
        COMPAS & -0.3473 \scriptsize{±0.0337} & -0.3304 \scriptsize{±0.0302} & -0.3385 \scriptsize{±0.0101} & -0.3149 \scriptsize{±0.0145} & -0.3385 \scriptsize{±0.0110} & \textbf{-0.3145} \scriptsize{±0.0149} \\
        \hline
        News & -0.2156 \scriptsize{±0.0003} & -0.2138 \scriptsize{±0.0010} & -0.2101 \scriptsize{±0.0012} & \textbf{-0.2097} \scriptsize{±0.0011} & \textbf{-0.2097} \scriptsize{±0.0012} & -0.2098 \scriptsize{±0.0012} \\
        \hline
        Adult & -0.4310 \scriptsize{±0.0166} & -0.3869 \scriptsize{±0.0325} & -0.3044 \scriptsize{±0.0149} & -0.2988 \scriptsize{±0.0158} & -0.3032 \scriptsize{±0.0141} & \textbf{-0.2987} \scriptsize{±0.0162} \\
        \hline
        Blog & -0.4009 \scriptsize{±0.0072} & -0.3414 \scriptsize{±0.0028} & -0.2732 \scriptsize{±0.0128} & -0.2705 \scriptsize{±0.0129} & \textbf{-0.2699} \scriptsize{±0.0128} & \textbf{-0.2699} \scriptsize{±0.0163} \\
        \hline
    \end{tabular}%
    }
    \label{tab:tll}
\end{table}

The computational burden associated with sampling can be a major bottleneck in scenarios requiring fast training and prediction, such as online systems or real-time applications. Such requirements are seen in {financial modeling} and {stock market prediction}, where models must adapt to real-time data to ensure accuracy \cite{tsantekidis2017using}. Similarly, {industrial IoT systems} rely on real-time predictions to optimize maintenance and reduce downtime, where fast retraining is key \cite{sun2017smart}.

\begin{figure}[H]
\centering
\centering
    \includegraphics[width=0.6\textwidth]{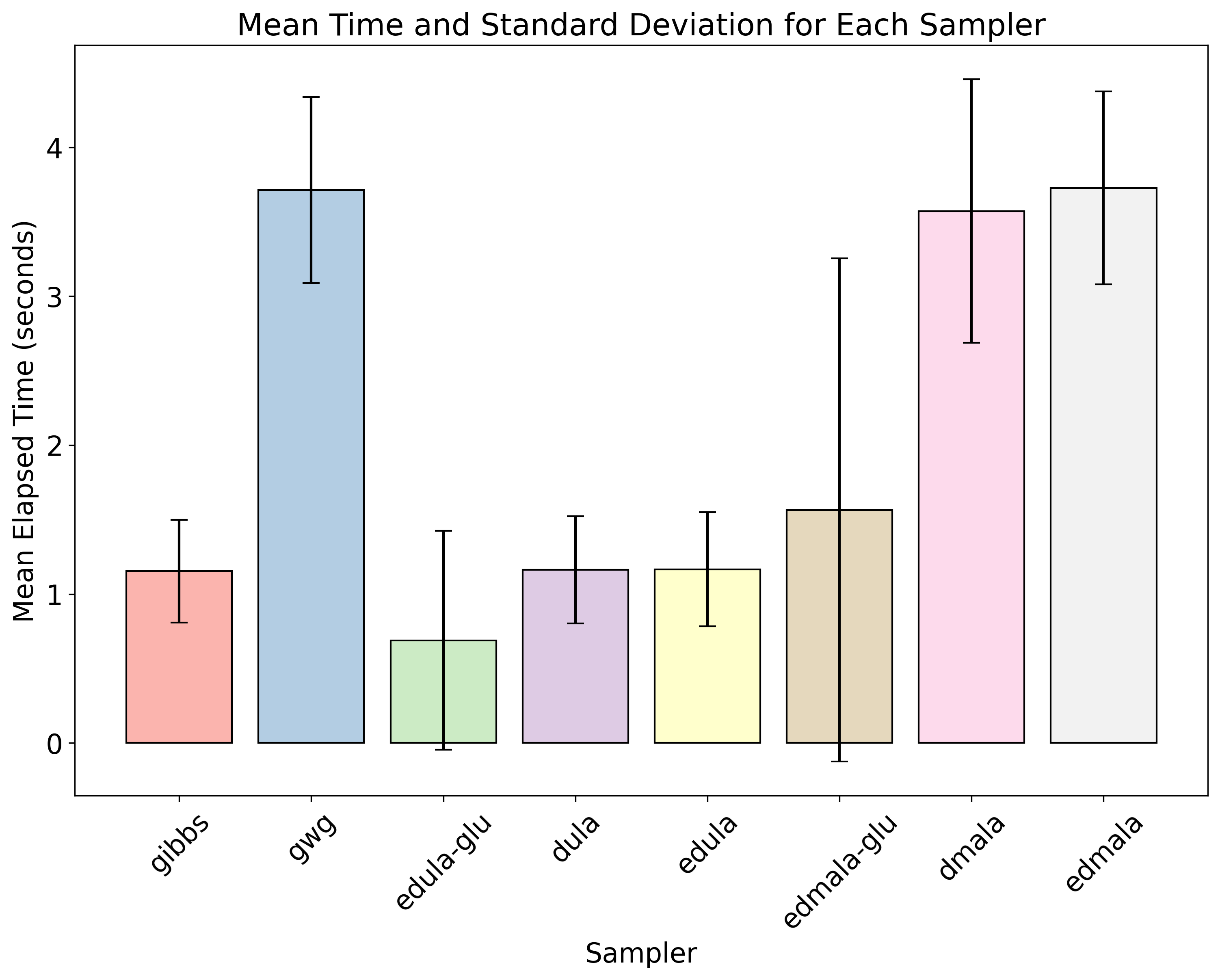}
    \caption{Runtime Analysis on Adult Dataset}
    \label{fig:bnnn_runtime_a}
\end{figure}

In Figure \ref{fig:bnnn_runtime_a}, we present the measured elapsed time per sample for the adult dataset to demonstrate these computational efficiencies, under the same settings as in Section \ref{sec:exp}, extending to include the GLU versions of the EDLP framework(Section \ref{sec:glu}), alongside the results for the standard DLP and EDLP methods.

As illustrated, the EDLP versions exhibit an increase in runtime compared to DLP, due to the modifications discussed in Section \ref{sec:coupling}. While the runtime difference between the DULA and EDULA algorithms (without MH correction) is negligible, the time difference between DMALA and EDMALA is more pronounced. This can be attributed to the more complex joint acceptance probability calculation required by EDMALA. Despite these variations, the overall runtime overhead for EDLP samplers is not substantial and remains practical.

For the EDLP-GLU variants, we maintained the same $\eta$ and $\alpha$ values as their corresponding vanilla DLP samplers. The EDLP-GLU variants naturally achieve an approximate $50\%$ reduction in runtime compared to EDLP. This efficiency stems from the alternating updates between sampling from a modified isotropic Gaussian and conditional DLP, designed to match the conditional distributions more effectively. However, this approach also introduces a higher standard deviation in runtime. The variability is primarily attributed to the contrasting computational costs between the two update types: sampling from the modified Gaussian is relatively lightweight, whereas the conditional DLP update is computationally intensive. As a result, the EDLP-GLU variants exhibit greater fluctuations in runtime compared to other samplers. Furthermore, the negative lower bounds are not physically meaningful and stem from the high variability in runtime measurements.

For details of datasets used, refer to the Appendix of \cite{zhang2022langevin}.

We fix $\alpha$ to 0.1 for DULA, DMALA, EDULA, and EDMALA. For more details on hyperparameters see Table \ref{tab:hyper}.
\begin{table}[ht]
    \centering
    \renewcommand{\arraystretch}{1}  % Adjust row height for better readability
    \caption{Hyper-parameter Settings }
    \begin{tabular}{|l|cc|cc|}
        \hline
        \multicolumn{5}{|c|}{Hyperparameters for EDLP} \\  % Merged header across 5 columns
        \hline
        {Dataset} & \multicolumn{2}{c|}{\textbf{EDULA}} & \multicolumn{2}{c|}{\textbf{EDMALA}} \\
        \hline
        & $\alpha_a$ & $\eta$ & $\alpha_a$ & $\eta$ \\
        \hline
        COMPAS & 0.0100 & 4.0 & 0.0010 & 4.0 \\
        News & 0.0100 & 2.0 & 0.0001 & 0.8 \\
        Adult & 0.0001 & 2.0 & 0.0001 & 4.0 \\
        Blog & 0.0100 & 1.0 & 0.0001 & 1.0 \\
        \hline
    \end{tabular}
    \label{tab:hyper}
\end{table}

All experiments in the paper were run on a single RTX A6000.
\newpage

\end{document}